\documentclass[pdflatex,sn-mathphys-num]{sn-jnl}

\usepackage{graphicx}%
\usepackage{multirow}%
\usepackage{amsmath,amssymb,amsfonts}%
\usepackage{amsthm}%
\usepackage{mathrsfs}%
\usepackage[title]{appendix}%
\usepackage{xcolor}%
\usepackage{textcomp}%
\usepackage{manyfoot}%
\usepackage{booktabs}%
\usepackage{algorithm}%
\usepackage{algorithmicx}%
\usepackage{algpseudocode}%
\usepackage{listings}%
\usepackage{longtable}
\usepackage{xurl}
\usepackage{placeins}

\theoremstyle{thmstyleone}%

\theoremstyle{thmstyletwo}%

\theoremstyle{thmstylethree}%

\begin{document}

\title{MedBOE: Optimization-Based Estimation of Complete Patient Body Models from Whole-Body Medical Imaging}

\author[1]{\fnm{Hoijoon} \sur{Jung}}
\equalcont{These authors contributed equally to this work.}

\author[1]{\fnm{Yuhao} \sur{Wei}}
\equalcont{These authors contributed equally to this work.}

\author[1]{\fnm{Mingxiao} \sur{Tu}}

\author[2]{\fnm{Younhyun} \sur{Jung}}

\author[3]{\fnm{Michael} \sur{Fulham}}

\author*[1]{\fnm{Jinman} \sur{Kim}}
\email{jinman.kim@sydney.edu.au}

\affil*[1]{\orgdiv{School of Computer Science},
\orgname{The University of Sydney},
\orgaddress{\city{Camperdown}, \state{NSW},
\postcode{2050}, \country{Australia}}}

\affil[2]{\orgdiv{School of Computing},
\orgname{Gachon University},
\orgaddress{\city{Seongnam-Si}, \state{Gyeonggi-Do},
\postcode{13120}, \country{Republic of Korea}}}

\affil[3]{\orgdiv{Department of Molecular Imaging},
\orgname{Royal Prince Alfred Hospital},
\orgaddress{\city{Camperdown}, \state{NSW},
\postcode{2050}, \country{Australia}}}

\abstract{True whole-body medical imaging captures a patient from head to toe and provides comprehensive anatomical and/or functional information. This coverage enables reconstruction of the external body surface, yielding patient body models. However, scanner geometry (e.g., bore diameter), patient characteristics, and positioning can leave body regions missing (e.g., lateral arms/torso, hands, forearms, elbows), producing incomplete patient body models. Such incompleteness limits downstream medical analyses, including derivation of body-habitus metrics (BHMs), such as body volume and surface area, for injected radiotracer activity regimens, cardiometabolic risk assessment, and computational anatomical modeling for medical imaging simulation and radiation dosimetry. We present the Medical Body model Optimization-based Estimator (MedBOE), an automated, training-free framework for estimating complete patient body models from whole-body images with missing body regions. Our innovation is to leverage the Skinned Multi-Person Linear (SMPL) template, a parametric body mesh representing a broad range of body shapes and poses, in medical imaging. MedBOE aligns the SMPL template to the incomplete patient body models via attribute-guided optimization and estimates missing body regions through geometrically and volumetrically constrained deformation. We evaluated MedBOE using two public datasets: AutoPET (91 whole-body PET-CT images) and Human Implicit Tissues (HIT; 40 whole-body segmentations from MRI images). MedBOE achieved the best overall volumetric and geometric performance among evaluated methods. BHMs derived from MedBOE were within 2\% error, suggesting low percentage errors for downstream analyses. MedBOE offers an automated, training-free method for estimating complete patient body models from whole-body images with missing body regions.}

\keywords{Patient body surface model, Complete body model estimation, Whole-body medical imaging, Body-habitus metrics, Missing body region}

\maketitle

\section{Introduction}\label{sec1}

Medical imaging is indispensable to current healthcare and is used for diagnosis, treatment planning, intervention, and treatment monitoring \cite{bib1}. Imaging modalities, including computed tomography (CT), magnetic resonance imaging (MRI), and multimodal positron emission tomography (PET)-CT, can provide three-dimensional (3D) anatomical and/or functional information. Recent PET-CT systems with extended axial coverage enable true whole-body imaging from head to toe \cite{badawi2019explorer}.

Whole-body imaging enables comprehensive reconstruction of the patient's external body surface, producing patient body models that can be visualized, manipulated, and analyzed for clinical applications. These applications include surgical navigation, where the body model is aligned with the patient's physical body using surface landmarks to enable on-body visualization of internal structures in conjunction with the model \cite{bib3}. It also enables the derivation of body-habitus metrics (BHMs), including body surface area, body volume, and regional circumferences \cite{bib4}, for optimizing injected radiotracer activity regimens \cite{bib5}, cardiometabolic risk assessment \cite{bib6}, and computational anthropomorphic anatomical models used in medical imaging simulation and radiation dosimetry \cite{bib44}.

\begin{figure}[htbp]
\centering
\includegraphics[width=\linewidth]{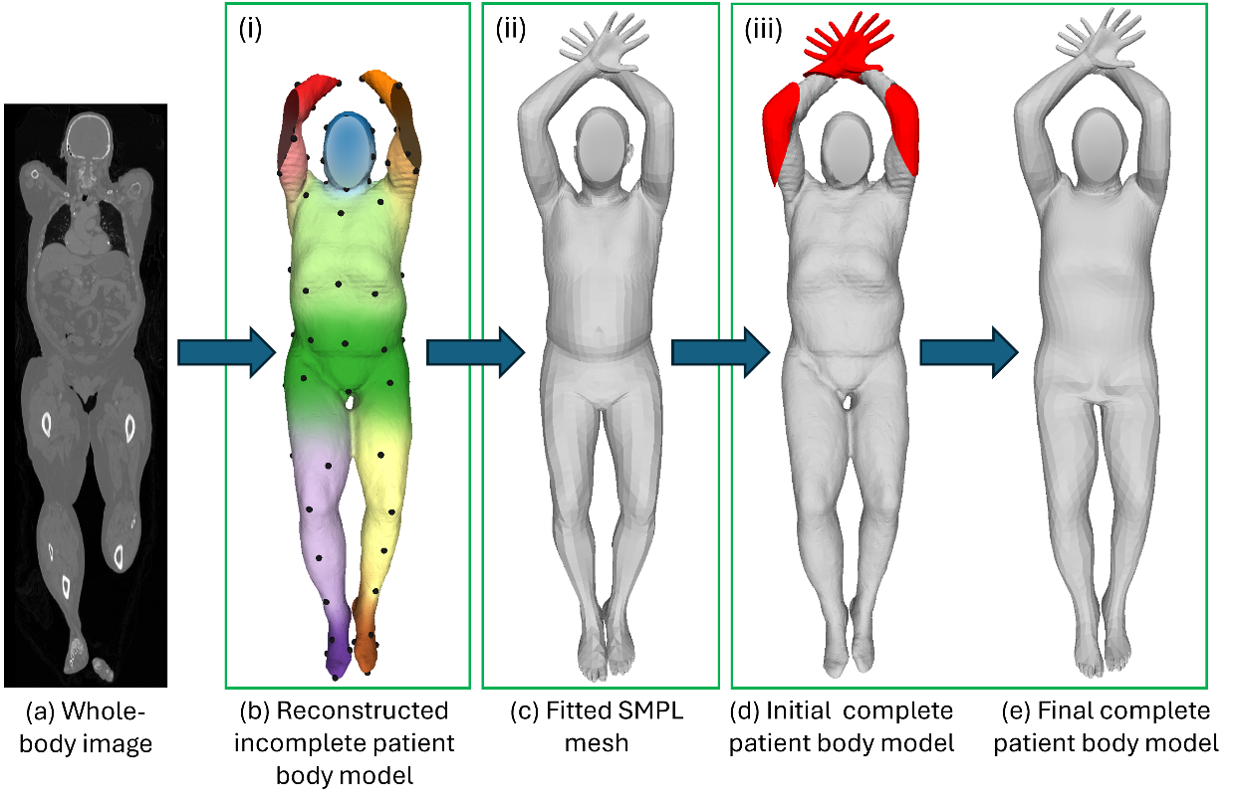}
\caption{Overview of the proposed MedBOE. A whole-body image with missing forearms, elbows, and hands (a) is processed through MedBOE's three modules: (i) \textit{Body model reconstruction}; (ii) \textit{Global parametric fitting}; (iii) \textit{Missing Body Region Estimation and Surface Deformation}, yielding outputs (b)--(e). The patient's face is blurred for anonymization}
\label{fig1}
\end{figure}

However, whole-body imaging (Fig.~\ref{fig1}(a)) often fails to fully capture certain body regions, including lateral arms, parts of the torso, hands, forearms, and lower limbs, primarily due to limited transverse field-of-view (FoV) imposed by the scanner gantry diameter and scanning protocol, insufficient axial coverage, patient body size, and patient positioning~\cite{bib45,bib47}. Such omissions produce incomplete patient body models (Fig.~\ref{fig1}(b)), reducing their utility for downstream applications requiring accurate BHMs \cite{bib5,bib6}.

In the multimedia domain, methods for estimating complete body models from incomplete inputs (e.g., partial meshes or point clouds) have been actively developed to create, for example, ``virtual humans'' for animation and motion capture \cite{bib8,bib9,bib10,bib11,bib12,bib13,bib14,bib15}. For human body reconstruction, these methods are commonly developed and evaluated using non-clinical data depicting upright subjects (e.g., standing or sitting), with incomplete observations arising from occlusion or partial surface capture. In contrast, medical imaging commonly involves supine positioning (lying back) and incomplete coverage, with some regions (e.g., hands and forearms) entirely excluded from the scan. Additionally, supervised completion methods typically rely on large training datasets of paired complete and incomplete models \cite{bib11}, which are not publicly available in medical imaging. Despite these advances in multimedia-based body reconstruction, differences in pose, body coverage, training data, and reconstruction objectives limit their direct applicability to clinical settings. This discrepancy highlights a critical gap between multimedia and medical imaging domains, motivating the need to examine how related challenges have been addressed within medical imaging research and to identify remaining limitations.

In the medical application domain, prior studies address related but distinct problems. Medical-imaging studies that compensate for missing information on a slice-wise or image-volume basis \cite{bib45,bib46,bib47} can recover missing image information for image correction, attenuation correction, and field-of-view extension. However, these truncation-compensation methods focus on recovering missing information in the image, sinogram, or attenuation-map domain, rather than estimating a continuous complete external body model. Other studies have leveraged the Skinned Multi-Person Linear (SMPL) template mesh \cite{bib7}, a parametric complete human body model, in medical imaging to estimate body composition and internal anatomical structures from external body surfaces \cite{bib17,bib18}. These studies typically rely on complete body coverage acquired using specialized imaging protocols or semi-automatic external-surface fitting procedures that are difficult to scale. More importantly, their primary aim was not to estimate complete patient body models from routine whole-body medical images with missing body regions. In our literature review, no prior work has addressed complete body model estimation from whole-body medical images with missing body regions.

The objective of this study was to develop and evaluate the Medical Body model Optimization-based Estimator (MedBOE), a fully automated, training-free framework for estimating complete patient body models from whole-body medical images with missing body regions. MedBOE addresses this gap by estimating a patient-specific complete surface model through a three-module pipeline (Fig.~\ref{fig1}) tailored for medical imaging. The specific contributions are:

\begin{enumerate}
\renewcommand{\labelenumi}{\arabic{enumi})}

\item An automated body model reconstruction module that converts whole-body medical images into image-derived incomplete patient body models and annotates anatomically consistent sparse and dense body-surface landmarks for subsequent fitting (Fig.~\ref{fig1}(b)).

\item A medical-image-tailored global parametric fitting strategy that uses attribute-guided optimization to jointly constrain anatomical alignment, body size, and pose plausibility, producing a patient-specific fitted SMPL mesh from incomplete body surfaces (Fig.~\ref{fig1}(c)).

\item A missing body region estimation and surface deformation strategy that combines missing regions estimated from the fitted SMPL mesh with the observed surface to form a complete target. Fine deformation then fits the SMPL mesh to this target with a volume consistency constraint to generate a continuous complete patient body model (Fig.~\ref{fig1}(d,e)).

\end{enumerate}

We evaluated MedBOE on two public datasets, AutoPET \cite{bib16} and Human Implicit Tissues (HIT) \cite{bib17}, and benchmarked it against the state-of-the-art methods adapted from the multimedia domain, demonstrating MedBOE's high accuracy in complete patient body model estimation and BHM derivation. MedBOE achieved the best overall performance among evaluated methods, and MedBOE-derived BHMs were within 2\% error, suggesting low percentage errors for downstream analyses. The MedBOE source code and optimization weights are publicly available at \url{https://github.com/Joon-Jung/MedBOE}.

\section{Related Work}\label{sec2}

Methods for estimating complete body models from incomplete inputs primarily originate from multimedia research and fall into two categories: 3D shape reconstruction \cite{bib8,bib9,bib10,bib11} and parametric body template mesh fitting \cite{bib12,bib13,bib14,bib15}.

\subsection{3D Shape Reconstruction Methods}\label{subsec:shape_reconstruction}

3D shape reconstruction methods employ deep learning to estimate complete body models from incomplete inputs. Wu, et al. \cite{bib10} introduced a 3D Generative Adversarial Network that learns a probabilistic latent space of 3D shapes using volumetric convolutions to synthesize and interpolate 3D shapes. Yuan, et al. \cite{bib8} proposed Point Completion Network (PCN), an encoder-decoder model using a coarse-to-fine decoding strategy to infer complete dense point clouds. Chibane, et al. \cite{bib11} developed Implicit Feature Network (IF-Net), which uses 3D convolutions to create multi-scale feature tensors and then classify deep features from these tensors to represent and reconstruct 3D surfaces. Yan, et al. \cite{bib9} introduced ShapeFormer, a transformer-based network encoding incomplete shapes into quantized tokens and estimating complete shapes using learned shape priors. Although effective in multimedia applications, these methods face two key challenges when applied to medical imaging: (i) the lack of publicly available large datasets pairing clinically truncated whole-body surfaces with complete reference surfaces, and (ii) the need to handle scans in which entire regions, such as the forearms or hands, may be completely absent.

\subsection{Parametric Body Template Mesh Fitting Methods}\label{subsec:parametric_fitting}

Parametric fitting methods estimate complete body models by aligning a body template mesh (fitting), typically SMPL \cite{bib7}, to incomplete inputs. Optimization-based methods achieve this by minimizing discrepancies in attributes, such as landmark distances and surface alignment, framing the process as an optimization problem. Bhatnagar, et al. \cite{bib15} proposed IP-Net, combining implicit function learning and semantic correspondences to refine SMPL fitting to sparse inputs. Liu, et al. \cite{bib12} incorporated an occlusion-aware voting mechanism and used joint confidence to identify occluded joints and complete their features for SMPL mesh estimation. Feng, et al. \cite{bib13} introduced ArtEq, which directly predicts SMPL shape and pose using part detection and articulated SE(3)-equivariant networks to generalize across unseen poses. Jang, et al. \cite{bib14} fitted an SMPL mesh to sequential partial point clouds using a motion-aware variational autoencoder with kinematic priors to estimate a complete body model. Originally designed for multimedia applications, these methods are generally developed using non-clinical data, and their application to medical imaging must account for differences in patient positioning and scan coverage. Sequence-based methods additionally utilize temporal information that is unavailable in single, static medical images acquired in supine positions, thus limiting their direct applicability.

\subsection{Image Compensation and SMPL-Based Body Model Fitting in Medical Applications}\label{subsec:medical_applications}

Prior medical-imaging studies have addressed two related but distinct problems: compensation for limited-FoV image truncation and SMPL-based body model fitting to medical images or external body surfaces.

Zamyatin et al. \cite{bib45} developed a truncation correction approach for CT reconstruction FoV extension using sinogram information. Fonseca et al. \cite{bib46} evaluated a U-Net-based extended-FoV CT reconstruction method for radiotherapy-related applications. Arabi and Zaidi \cite{bib47} investigated a deep learning-based object completion approach using a dilated CNN with residual connections to compensate for body truncation in PET/MRI attenuation correction. These approaches reconstruct or compensate for missing image and attenuation-map information rather than estimate complete patient body models with continuous surface representations. In contrast, MedBOE estimates a complete continuous patient body surface model for geometric and BHM analysis, which is complementary to image-domain truncation compensation.

A second line of work has used SMPL-based fitting in medical imaging to infer body composition or internal anatomical structures from external body surfaces. Keller, et al. \cite{bib17} fitted SMPL to complete body models reconstructed from extended whole-body MRI images acquired using a specialized protocol for assessing adipose tissue distribution, enabling prediction of internal adipose compartments from complete patient body models. Shetty, et al. \cite{bib18} estimated internal organ positions from external body surfaces by fitting the SMPL mesh to incomplete body models using a multi-stage optimization guided by manually annotated landmarks and surface distances. Although these studies demonstrate the utility of SMPL in medical imaging, their model construction relies on complete body coverage obtained using specialized imaging protocols or semi-manual external-surface fitting procedures that are difficult to scale. Furthermore, their primary aim was not the fully automated estimation of complete external patient body models from routine whole-body medical images with missing body regions.

\section{Method}\label{sec3}

MedBOE estimates complete patient body models from whole-body medical images through a three-module pipeline (Fig.~\ref{fig1}). The body model reconstruction module (Fig.~\ref{fig1}(i)) generates incomplete patient body models from the whole-body images and automatically annotates the body-surface landmarks on the reconstructed models (Fig.~\ref{fig1}(b)). The global parametric fitting module (Fig.~\ref{fig1}(ii)) fits a SMPL template mesh to the incomplete patient body models and produces the fitted SMPL model (Fig.~\ref{fig1}(c)). The missing body region estimation and surface deformation module (Fig.~\ref{fig1}(iii)) estimates the missing body regions (red in Fig.~\ref{fig1}(d)) from the fitted SMPL mesh and stitches them to the incomplete patient body models to construct initial complete body models (Fig.~\ref{fig1}(d)). The fitted SMPL mesh is then further deformed to the initial complete body models, producing the final complete patient body models (Fig.~\ref{fig1}(e)).

\subsection{SMPL Template Mesh}\label{subsec:smpl_template}

The SMPL template mesh \cite{bib7} represents the human body as a triangulated mesh with 6,890 vertices, parameterized by body shape $\beta$, pose $\theta$, and global translation $t$. $\beta$ is modeled using low-dimensional principal component analysis coefficients derived from a statistical body shape space, while $\theta$ is defined through a 24-joint kinematic skeleton. The mesh is deformed using Linear Blend Skinning, with joint regressor $\mathcal{J}(\beta)$, blend weights $\mathcal{W}$, and pose-dependent shape variations $BP(\theta)$. The final mesh $M$ is computed as:

\begin{equation}
M = W\left(
\bar{T} + BS(\beta) + BP(\theta),
\theta, \mathcal{J}(\beta), \mathcal{W}
\right),
\label{eq:smpl_mesh}
\end{equation}

where $\bar{T}$ is the canonical template mesh, $\mathcal{W}$ is the blend weight, $BS(\beta)$ and $BP(\theta)$ are the shape and pose blend shapes, respectively. Refer to Loper et al. \cite{bib7} for further details.

\subsection{Body Model Reconstruction Module}\label{subsec:body_reconstruction}

Whole-body images were segmented using a pre-trained nnU-Net \cite{bib19} to isolate the patient's body. The resulting binary masks were reconstructed into incomplete body models using the marching cubes algorithm \cite{bib20} (Fig.~\ref{fig1}(b)). These models were smoothed using Taubin's algorithm \cite{bib21} and simplified via Schroeder's decimation \cite{bib22}. Patient metadata (gender, weight, height) was extracted from the image headers.

Body landmarks were automatically annotated using the Deep Virtual Marker \cite{bib23}, which assigned each vertex of the incomplete patient body model a 99-dimensional virtual-marker label vector. For each model, sparse landmarks (black dots in Fig.~\ref{fig1}(b)) were obtained by selecting the vertex with the highest response for each of the 99 virtual-marker labels. Dense landmarks were represented by the 99-dimensional virtual-marker label vector assigned to each vertex, visualized as surface colors in Fig.~\ref{fig1}(b). The same procedure was applied to the default male and female SMPL meshes. Only sparse landmarks with confidence greater than 0.5 were included in the sparse landmark distance term.

\subsection{Global Parametric Fitting Module}\label{subsec:global_fitting}

The global parametric fitting module aligns a gender-specific SMPL template mesh to the reconstructed incomplete patient body model (Fig.~\ref{fig1}(b)). The module uses six attributes: sparse landmark distance ($E_{\mathrm{Sparse}}$), surface distance ($E_{\mathrm{Dist}}$), pose prior ($E_{\mathrm{Prior}}$), weight consistency ($E_{\mathrm{Weight}}$), self-intersection penalty ($E_{\mathrm{Contact}}$), and dense landmark distance ($E_{\mathrm{Dense}}$). These attributes collectively enforce global and local alignment, inspired by prior studies \cite{bib18,bib24}. Specifically, $E_{\mathrm{Sparse}}$ and $E_{\mathrm{Dense}}$ constrain anatomical correspondence from virtual-marker labels and improve landmark-level alignment. $E_{\mathrm{Dist}}$ uses a modified symmetric Chamfer distance with correspondence filtering to improve global geometric alignment. $E_{\mathrm{Prior}}$ constrains unrealistic joint configurations and stabilizes pose estimation using VPoser \cite{bib26} and MaxMixturePrior \cite{bib27}. $E_{\mathrm{Weight}}$ constrains SMPL shape parameters using patient weight and improves patient-specific body size matching by leveraging BetaNet \cite{bib24}. $E_{\mathrm{Contact}}$ constrains mesh self-collisions and reduces anatomically implausible overlap, such as hand penetration into the body.

The global parametric fitting module optimizes the weighted sum of these attributes ($E_{\mathrm{Fit}}$) by adjusting the fitting SMPL's $\beta$, $\theta$, and $t$:

\begin{equation}
\begin{aligned}
E_{\mathrm{Fit}} = {}& \lambda_{\mathrm{Sparse}} E_{\mathrm{Sparse}} + \lambda_{\mathrm{Dist}} E_{\mathrm{Dist}} + \lambda_{\mathrm{Prior}} E_{\mathrm{Prior}} \\
&+ \lambda_{\mathrm{Weight}} E_{\mathrm{Weight}} + \lambda_{\mathrm{Contact}} E_{\mathrm{Contact}} + \lambda_{\mathrm{Dense}} E_{\mathrm{Dense}},
\end{aligned}
\label{eq:global_fitting}
\end{equation}

The parameters $\beta$, $\theta$, and $t$ were initialized to zero and jointly optimized using the Adam optimizer \cite{bib25} over 500 iterations, with each attribute scaled by $\lambda_x$.

\subsection{Missing Body Region Estimation and Surface Deformation Module}\label{subsec:surface_deformation}

The missing body region estimation and surface deformation module comprises coarse and fine stages. In the coarse stage, the fitted SMPL mesh (Fig.~\ref{fig1}(c)) is non-rigidly deformed to the incomplete patient body model (Fig.~\ref{fig1}(b)) to estimate missing body regions and construct an initial complete patient body model (Fig.~\ref{fig1}(d)).

The coarse stage uses four attributes: sampled surface distance ($E_{\mathrm{Surf}}$), edge length ($E_{\mathrm{Edge}}$), Laplacian smoothness \cite{bib29} ($E_{\mathrm{Lap}}$), and normal consistency ($E_{\mathrm{Normal}}$). $E_{\mathrm{Surf}}$ compares valid overlapping surface samples between the fitted SMPL mesh and the incomplete patient body model. The attribute minimizes squared surface distance while excluding non-overlapping regions using Euclidean-distance and surface angular-difference thresholds. The torso and lower limbs were always included because they were consistently present in whole-body images. $E_{\mathrm{Edge}}$ constrains changes in vertex edge lengths to preserve local mesh geometry during deformation. $E_{\mathrm{Lap}}$ regularizes abrupt local vertex displacement and enforces local surface smoothness. $E_{\mathrm{Normal}}$ minimizes orientation differences between neighboring surface elements and reduces orientation discontinuity. Together, these attributes coarsely deform the fitted SMPL mesh toward the incomplete patient body model while preserving local surface geometry and continuity.

The coarse stage optimizes the weighted sum of these attributes ($E_{\mathrm{Coarse}}$) over 100 iterations using the Adam optimizer \cite{bib25}:

\begin{equation}
E_{\mathrm{Coarse}} = \lambda_{\mathrm{Surf}} E_{\mathrm{Surf}} + \lambda_{\mathrm{Edge}} E_{\mathrm{Edge}} + \lambda_{\mathrm{Lap}} E_{\mathrm{Lap}} + \lambda_{\mathrm{Normal}} E_{\mathrm{Normal}},
\label{eq:coarse_deformation}
\end{equation}

After the coarse stage, the boundary of the incomplete patient body model was used to trim the deformed SMPL mesh, isolating estimated missing body regions (red in Fig.~\ref{fig1}(d)). These regions were stitched to the incomplete patient body model using Poisson surface reconstruction \cite{bib30}, producing an initial complete patient body model (Fig.~\ref{fig1}(d)).

In the fine stage, the fitted SMPL mesh is further deformed to the initial complete patient body model, formulating the problem as complete-to-complete alignment. This stage smooths visible discontinuities between estimated and existing regions and reduces residual discrepancies introduced during coarse estimation.

The fine stage uses five attributes: Chamfer distance ($E_{\mathrm{CD}}$), Chamfer normal ($E_{\mathrm{CN}}$), volume consistency ($E_{\mathrm{Vol}}$), $E_{\mathrm{Lap}}$, and $E_{\mathrm{Normal}}$. $E_{\mathrm{CD}}$ measures bidirectional nearest-neighbor point distances between the deforming SMPL mesh and the initial complete patient body model, reducing residual pointwise discrepancy. $E_{\mathrm{CN}}$ compares corresponding surface normals between the deforming SMPL mesh and the initial complete patient body model and improves local orientation consistency. $E_{\mathrm{Vol}}$ constrains changes in the SMPL mesh volume to preserve patient-specific body size established during global parametric fitting. $E_{\mathrm{Lap}}$ and $E_{\mathrm{Normal}}$ were reused from the coarse stage.

The fine stage optimizes the weighted sum of these attributes ($E_{\mathrm{Fine}}$) over 150 iterations using the Adam optimizer:

\begin{equation}
E_{\mathrm{Fine}} = \lambda_{\mathrm{CD}} E_{\mathrm{CD}} + \lambda_{\mathrm{CN}} E_{\mathrm{CN}} + \lambda_{\mathrm{Vol}} E_{\mathrm{Vol}} + \lambda_{\mathrm{Lap}} E_{\mathrm{Lap}} + \lambda_{\mathrm{Normal}} E_{\mathrm{Normal}},
\label{eq:fine_deformation}
\end{equation}

Deformation was applied only to regions overlapping with the initial complete body model, determined by pointwise distance. To ensure anatomical completeness, the torso, legs, head, neck, and arms were forcibly included, while deformation of the forearms, hands, and feet was restricted to overlapping regions to avoid propagating discontinuities introduced in the previous step. By jointly optimizing these attributes, the module produces the estimated complete model with a smooth surface, geometric accuracy, and volumetric consistency (Fig.~\ref{fig1}(e)).

Detailed formulations of attributes ($E_x$) used in the global parametric fitting and missing body region estimation and surface deformation modules are provided in Appendix~\ref{sec:appendix_attributes}.

\subsection{Implementation}\label{subsec:implementation}

The body model reconstruction module was implemented using the VTK library \cite{bib31} and the official Deep Virtual Marker package \cite{bib23}. The global parametric fitting and missing body region estimation and surface deformation modules were implemented in PyTorch with smplx and NVIDIA Kaolin \cite{bib26,bib32,bib33}, incorporating official pretrained models of MaxMixturePrior, VPoser, $E_{\mathrm{Contact}}$, and BetaNet \cite{bib24,bib26,bib27,bib28}. Additional libraries, including SimpleITK, OpenCV, pyMeshLab, and pyMeshFix \cite{bib34,bib35,bib36,bib37}, were employed for extracting and integrating missing body regions.

The Adam optimizer was used with learning rates of 0.01 for global parametric fitting and the coarse stage of the missing region estimation and surface deformation, and 0.001 for the fine stage. In $E_{\mathrm{Dist}}$, correspondences were excluded when the Euclidean distance exceeded 0.1 m or 0.01 m, or when the surface-normal angular difference exceeded $45^\circ$ or $30^\circ$, for SMPL-to-incomplete patient body model and incomplete patient body model-to-SMPL correspondences, respectively. In $E_{\mathrm{Dense}}$, correspondences were excluded when the Euclidean distance exceeded 0.1 m or when the surface-normal angular difference exceeded $45^\circ$. Pointwise distance thresholds for detecting overlapping regions were set to 0.01 m for the coarse stage and 0.005 m for the fine stage.

Attribute weights ($\lambda_x$) used in the global parametric fitting module are detailed in Table~\ref{tab:global_weights}. Early iterations prioritized $E_{\mathrm{Sparse}}$ and $E_{\mathrm{Dense}}$, followed by increased emphasis on geometric alignment ($E_{\mathrm{Dist}}$). The weight for $E_{\mathrm{Weight}}$ was reduced to allow flexibility in adapting to the incomplete patient body model geometry. $E_{\mathrm{Prior}}$ remained constant, while $E_{\mathrm{Contact}}$ was activated only during the final 100 iterations to reduce computation.

\begin{table}[htbp]
\centering
\caption{Attribute weights used in the global parametric fitting module during optimization.}\label{tab:global_weights}
\begin{tabular*}{\textwidth}{@{\extracolsep{\fill}}lcccccc@{}}
\toprule
Iteration & $E_{\mathrm{Sparse}}$ & $E_{\mathrm{Dist}}$ & $E_{\mathrm{Prior}}$ & $E_{\mathrm{Weight}}$ & $E_{\mathrm{Contact}}$ & $E_{\mathrm{Dense}}$ \\
\midrule
0 & 100 & 0.5 & 0.5 & 10 & 0 & 10 \\
30 & 80 & 2 & 0.5 & 10 & 0 & 10 \\
100 & 50 & 5 & 0.5 & 10 & 0 & 5 \\
200 & 5 & 10 & 0.5 & 1 & 0 & 2 \\
300 & 2 & 10 & 0.5 & 0.5 & 0 & 2 \\
400 & 2 & 10 & 0.5 & 0 & 2 & 2 \\
\botrule
\end{tabular*}
\end{table}

For the missing region estimation and surface deformation module, weights are fixed at 0.001 ($\lambda_{\mathrm{Surf}}$), 0.01 ($\lambda_{\mathrm{Edge}}$), 0.01 ($\lambda_{\mathrm{Lap}}$; coarse stage), 1 ($\lambda_{\mathrm{Normal}}$; coarse stage), 0.2 ($\lambda_{\mathrm{CD}}$), 0.04 ($\lambda_{\mathrm{CN}}$), 1000 ($\lambda_{\mathrm{Vol}}$), 0.02 ($\lambda_{\mathrm{Lap}}$; fine stage), and 1 ($\lambda_{\mathrm{Normal}}$; fine stage). Geometry-based attributes (e.g., $E_{\mathrm{Dist}}$, $E_{\mathrm{Surf}}$) were constrained below empirically determined upper bounds to maintain optimization stability. Higher weights caused mesh collapse due to overfitting.

\section{Experimental Setting}\label{sec4}

\subsection{Datasets}\label{subsec:datasets}

\subsubsection{AutoPET Dataset}\label{subsubsec:autopet_dataset}

AutoPET \cite{bib16} includes 1,014 PET-CT examinations of 900 patients acquired on a Siemens Biograph mCT PET-CT scanner. AutoPET was selected because it is a publicly available routine clinical PET-CT dataset in which transverse FoV limitations are directly relevant to complete external-body model estimation. After excluding cases lacking weight data ($n = 149$), non-true whole-body scans ($n = 466$), transverse FoV overflow of the body in PET ($n = 307$), and one with motion artifacts, 91 images were retained (35 male, 56 female). The dataset was used in 5-fold cross-validation. Images lacking recorded body weight were excluded because $E_{\mathrm{Weight}}$ uses patient weight.

Ground truth (GT) body models were constructed by segmenting PET images (which have a wider transverse FoV than CT) and applying the marching cubes algorithm \cite{bib20}. Incomplete body models were simulated by trimming GT models using a cylinder with a radius equal to 95\% of the average CT transverse FoV radius, centered on the PET images, to replicate missing body regions typical in routine whole-body imaging. Representative front, left, and right views comparing the virtually trimmed model with the body model reconstructed from CT are provided in Appendix Fig.~6. For arms-up and mixed (up \& down) arm poses, parts of the forearms and hands were absent in the GT due to the diagnostic scan range.

On average, the simulated trimming removed 1.75 $\pm$ 1.62\% (0.26--7.80\%) of body volume and 4.46 $\pm$ 3.84\% (0.84--20.95\%) of body surface area in males, and 1.32 $\pm$ 1.33\% (0.01--6.53\%) of body volume and 3.74 $\pm$ 3.10\% (0.10--15.02\%) of body surface area in females.

\subsubsection{HIT Dataset}\label{subsubsec:hit_dataset}

The HIT dataset \cite{bib17} includes complete body segmentations from 419 MRI images (168 male, 251 female), acquired using a 1.5~T Siemens Magnetom Sonata MR scanner under a specialized protocol for whole-body adipose tissue mapping. Patients were scanned in a prone position (lying face down) with arms extended overhead and compressed within the scanner's transverse FoV, which is smaller than that of most current CT and PET-CT scanners. The dataset includes GT SMPL meshes aligned to the segmentations and annotated with $\beta$ and $\theta$. A predefined split (train/val/test) was used: 136/17/15 for male and 201/25/25 for female subjects.

GT body models were reconstructed from the segmentations using the marching cubes algorithm \cite{bib20}. Incomplete body models were simulated using the same trimming method as in Section~\ref{subsubsec:autopet_dataset} due to the absence of original MRI FoV data. Additionally, regions extending 10~cm above the head were removed to reflect typical scan overage (Fig.~\ref{fig1}). As patient weights were not provided, HIT weights were estimated by multiplying the GT model volumes by gender-specific mean body density \cite{bib38}, following Choutas et al. \cite{bib39}.

This estimation approach was a pragmatic dataset-specific decision rather than the preferred input, because measured patient weight is preferable for the $E_{\mathrm{Weight}}$ term. In AutoPET, a sufficient measured-weight cohort remained after exclusions, so cases without recorded weight were not assigned estimated weights. This avoided mixing measured and estimated weights within the same validation cohort. In HIT, measured weights were not available; without weight estimation, HIT could not be used to evaluate regions such as hands and forearms, which were frequently absent in the AutoPET GT body models.

On average, trimming removed 3.35 $\pm$ 0.72\% (2.09--4.83\%) of body volume and 10.19 $\pm$ 2.15\% (7.06--15.29\%) of body surface area in males, and 2.38 $\pm$ 0.51\% (1.59--3.42\%) of body volume and 7.67 $\pm$ 1.48\% (5.88--11.86\%) of body surface area in females.

\subsection{Metrics}\label{subsec:metrics}

\subsubsection{Accuracy of Estimated Complete Body Models}\label{subsubsec:complete_model_accuracy}

MedBOE's performance was evaluated by comparing estimated complete body models to corresponding GT models. To account for gender-related body shape variation, evaluations were performed separately for male and female models. For the AutoPET dataset, surface and volumetric comparisons were limited to regions present in the PET-derived GT models because parts of the forearms and hands were often excluded during acquisition (Section~\ref{subsubsec:autopet_dataset}). Thus, AutoPET metrics evaluate accuracy over observable GT regions rather than complete accuracy over all body regions. In contrast, the HIT dataset allowed evaluation across the entire body. Additionally, estimated regions were assessed against the trimmed regions.

Five metrics were used, following Jang et al. \cite{bib14}. Relative body surface area error (RBE) and relative volume error (RVE) quantified the mean absolute percentage errors in surface area and volume, respectively. Intersection over Union (IoU) quantified volumetric overlap between estimated and GT models. Chamfer distance (CD) represented the mean bidirectional closest-point distance between surfaces, and Hausdorff distance (HD) captured the maximum pointwise deviation. These metrics assessed differences in body size and volumetric overlap (RBE, RVE, IoU), as well as surface geometry (CD, HD).

\subsubsection{Accuracy of BHMs Derived from Estimated Models}\label{subsubsec:bhm_accuracy}

Five BHMs (i.e., body surface area, body volume, and circumferences of the chest, waist, and hip) were derived from estimated complete body models and compared to those from GT models. These metrics were selected for their clinical relevance in applications, such as injected radiotracer activity regimens \cite{bib5} and cardiometabolic risk assessment \cite{bib6}, where accurate estimation of overall body size and key regional dimensions is critical.

Body surface area was calculated from the total area of triangular facets, while body volume was computed by summing face-based volumetric contributions derived from vertex coordinates. Circumferences were derived using the method of Choutas et al. \cite{bib39}, which computes the closed curves around predefined SMPL body landmarks (right mammary papilla, umbilicus, and pubic symphysis).

Accuracy was expressed as mean percentage and absolute errors. Evaluation was limited to the HIT dataset, as GT circumference measurements were readily available only in this dataset. These measurements require body landmarks that can be automatically extracted, for example, from the GT SMPL meshes. GT values for complete body volume and surface area were also exclusive to HIT.

\subsection{Runtime Delay}\label{subsec:runtime_delay}

Runtime delay was additionally measured to evaluate computational cost. Runtime delay was defined as the elapsed time required to estimate a complete patient body model from an input incomplete patient body model. Therefore, this measurement did not include image segmentation or incomplete patient body model reconstruction from the original medical images. Runtime was measured on a desktop computer with Ubuntu 22.04.5 LTS, an Intel i7-8700K CPU, 64~GB memory and an NVIDIA TITAN V GPU.

\subsection{Baselines}\label{subsec:baselines}

MedBOE was compared with four state-of-the-art methods for complete body model estimation: IF-Net \cite{bib11}, ShapeFormer \cite{bib9}, ArtEq \cite{bib13}, and Point Completion Network (PCN) \cite{bib8}, as introduced in Section~\ref{sec2}. These baselines were selected based on their relevance and availability of official implementations.

IF-Net and ShapeFormer were trained and evaluated on the AutoPET and HIT datasets using incomplete body models automatically generated during training. PCN was trained on both datasets using paired GT and simulated incomplete body models; however, volumetric metrics were excluded for PCN because its point cloud output lacks surface representation.

ArtEq was evaluated only on the HIT dataset due to its reliance on GT SMPL parameters. It was trained using SMPL meshes with GT parameters, with incomplete body models generated by randomly sampling the meshes. Because the official implementation settings failed to estimate complete models, the training epoch count was increased from 15 to 150.

\subsection{Ablation Study}\label{subsec:ablation}

We assessed the contribution of the missing body region estimation and surface deformation module by comparing the full pipeline with a variant that omits this module and directly outputs the fitted SMPL mesh after SMPL fitting. All other settings remained constant.

\subsection{Statistical Analysis}\label{subsec:statistical_analysis}

All results are reported as mean $\pm$ standard deviation, with ranges indicated in parentheses (min--max). One-way ANOVA was used for multi-group comparisons, followed by post-hoc Tukey's HSD tests. Paired comparisons were performed using the paired t-test. Male-female differences were assessed using Welch's independent two-sample t-test \cite{bib48}. The ANOVA-based comparisons were interpreted with consideration of their standard assumptions, including independent observations, approximately normally distributed residuals, and homogeneity of variance. Effect sizes are reported using Cohen's $d$, and uncertainty is reported using 95\% confidence intervals (CIs), where applicable. Statistical significance was set to $p < 0.05$.

\section{Results}\label{sec:results}

\subsection{Estimated Complete Body Models}\label{subsec:results_complete_models}

\begin{figure}[!htbp]
\centering
\includegraphics[width=0.7\linewidth]{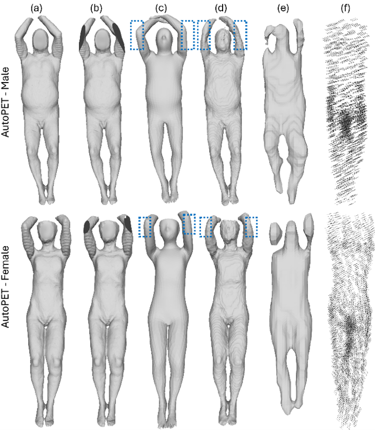}
\caption{Complete body models estimated from the AutoPET dataset. (a) is GT body models, (b) is simulated incomplete models where darker grey indicates the internal surface exposed by the trimming procedure. Results of (c) MedBOE, (d) IF-Net, (e) ShapeFormer, and (f) PCN. Blue dotted boxes highlight differences in estimating the missing body regions (elbows) between MedBOE and IF-Net}
\label{fig2}
\end{figure}

\begin{figure}[!htbp]
\centering
\includegraphics[width=0.7\linewidth]{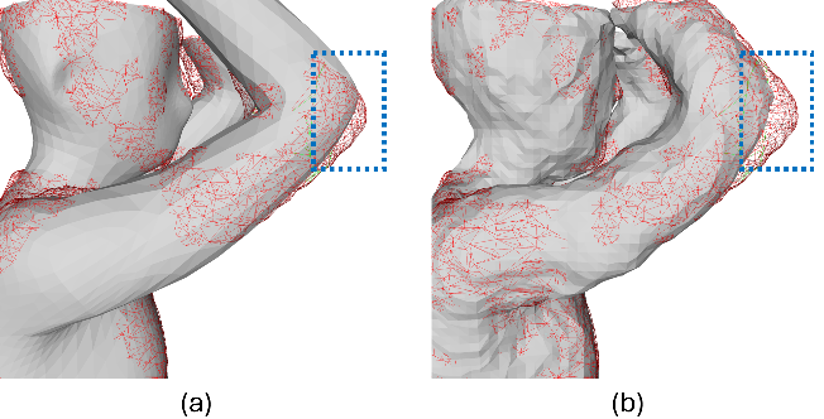}
\caption{Detailed comparison of the missing body regions estimated by (a) MedBOE and (b) IF-Net. Blue dotted boxes highlight the key difference in estimated regions. GT models are shown as red wireframes}
\label{fig3}
\end{figure}

Estimated complete body models from the AutoPET dataset are presented in Fig.~\ref{fig2}, highlighting key differences (blue boxes) between the two best-performing methods, MedBOE (Fig.~\ref{fig2}(c)) and IF-Net (Fig.~\ref{fig2}(d)), detailed further in Fig.~\ref{fig3}. Zoomed views of the regions above the shoulders and around the legs, where differences among method estimates are visible, are provided in Appendix Figs.~\ref{fig7} and~\ref{fig8}. Quantitative results are summarized in Table~\ref{tab:results_comparison}. A representative challenging case is presented in the Appendix Fig.~\ref{fig9}.

For AutoPET males, MedBOE achieved the best volumetric metrics (RBE, RVE, and IoU), followed by IF-Net and ShapeFormer. IF-Net yielded the lowest CD and HD (lower is better), with MedBOE second-best. No significant differences were found between MedBOE and IF-Net across all metrics (RBE: $p = 0.439$; RVE: $p = 0.338$; IoU: $p = 0.854$; CD: $p = 0.719$; HD: $p = 0.972$), but both methods significantly outperformed PCN and ShapeFormer (all $p \leq 7.15 \times 10^{-4}$). For AutoPET females, MedBOE showed superior performance in RVE and was second-best in other metrics. IF-Net led in CD, HD, RBE, and IoU, followed by MedBOE, ShapeFormer, and PCN (excluding RBE and IoU). Similar to males, differences between MedBOE and IF-Net were non-significant across all metrics (RBE: $p = 0.734$; RVE: $p = 0.452$; IoU: $p = 0.502$; CD: $p = 0.395$; HD: $p = 0.499$), whereas MedBOE significantly outperformed ShapeFormer and PCN (all $p \leq 1.22 \times 10^{-7}$). Full p-values, Cohen's $d$, and 95\% CIs for method comparisons are reported in Appendix Table~5.

\begin{sidewaystable}
\centering
\caption{Quantitative evaluation of complete patient body models estimated from the AutoPET and HIT datasets. The best result for each metric is shown in bold, and the second-best result is underlined.}
\label{tab:results_comparison}
\setlength{\tabcolsep}{1.5pt}
\renewcommand{\arraystretch}{2.25}
\footnotesize
\begin{tabular*}{\textheight}{@{\extracolsep{\fill}}llccccclccccc@{}}
\toprule
\multicolumn{7}{@{}c}{\textbf{AutoPET}} & \multicolumn{6}{c@{}}{\textbf{HIT}} \\
\cmidrule(lr){1-7}\cmidrule(lr){8-13}
 & \textbf{Methods} & \textbf{RBE $\downarrow$} & \textbf{RVE $\downarrow$} & \textbf{IoU $\uparrow$} & \textbf{CD $\downarrow$} & \textbf{HD $\downarrow$} & & \textbf{RBE $\downarrow$} & \textbf{RVE $\downarrow$} & \textbf{IoU $\uparrow$} & \textbf{CD $\downarrow$} & \textbf{HD $\downarrow$} \\
\midrule
\multirow{5}{*}{\rotatebox[origin=c]{90}{\textbf{Male (35)}}} & \shortstack[l]{\textbf{MedBOE}} & \shortstack[c]{\textbf{\boldmath $1.47 \pm 1.61$}\\\textbf{(0.12--9.04)}} & \shortstack[c]{\textbf{\boldmath $0.90 \pm 0.80$}\\\textbf{(0.03--4.22)}} & \shortstack[c]{\textbf{\boldmath $0.92 \pm 0.03$}\\\textbf{(0.83--0.94)}} & \shortstack[c]{\underline{$0.02 \pm 0.02$}\\\underline{(0.02--0.08)}} & \shortstack[c]{\underline{$0.1 \pm 0.1$}\\\underline{(0.03--0.39)}} & \multirow{5}{*}{\rotatebox[origin=c]{90}{\textbf{Male (15)}}} & \shortstack[c]{\textbf{\boldmath $1.85 \pm 1.87$ *}\\\textbf{(0.01--7.27)}} & \shortstack[c]{\textbf{\boldmath $0.86 \pm 0.40$ *}\\\textbf{(0.16--1.52)}} & \shortstack[c]{\textbf{\boldmath $0.94 \pm 0.01$}\\\textbf{(0.93--0.95)}} & \shortstack[c]{\textbf{\boldmath $0.04 \pm 0.01$}\\\textbf{(0.02--0.06)}} & \shortstack[c]{\underline{$0.19 \pm 0.09$}\\\underline{(0.08--0.38)}} \\
 & \shortstack[l]{\textbf{IF-Net}} & \shortstack[c]{\underline{$3.29 \pm 6.21$}\\\underline{(0.03--34.34)}} & \shortstack[c]{\underline{$3.82 \pm 8.44$}\\\underline{(0.30--51.15)}} & \shortstack[c]{\underline{$0.91 \pm 0.08$}\\\underline{(0.45--0.94)}} & \shortstack[c]{\textbf{\boldmath $0.01 \pm 0.01$}\\\textbf{(0.01--0.05)}} & \shortstack[c]{\textbf{\boldmath $0.09 \pm 0.03$}\\\textbf{(0.05--0.18)}} &  & \shortstack[c]{$12.90 \pm 2.28$\\(8.71--15.48)} & \shortstack[c]{\underline{$5.24 \pm 1.25$}\\\underline{(2.73--7.32)}} & \shortstack[c]{\underline{$0.91 \pm 0.01$}\\\underline{(0.89--0.93)}} & \shortstack[c]{\underline{$0.04 \pm 0.01$}\\\underline{(0.02--0.06)}} & \shortstack[c]{$0.36 \pm 0.05$\\(0.29--0.46)} \\
 & \shortstack[l]{\textbf{Shape}\\\textbf{Former}} & \shortstack[c]{$14.76 \pm 8.58$\\(1.14--33.12)} & \shortstack[c]{$19.35 \pm 12.33$\\(0.16--41.74)} & \shortstack[c]{$0.58 \pm 0.07$\\(0.42--0.71)} & \shortstack[c]{$0.06 \pm 0.02$\\(0.04--0.10)} & \shortstack[c]{$0.17 \pm 0.08$\\(0.08--0.41)} &  & \shortstack[c]{$12.72 \pm 4.97$\\(1.26--21.49)} & \shortstack[c]{$10.93 \pm 3.99$\\(4.96--18.91)} & \shortstack[c]{$0.70 \pm 0.06$\\(0.59--0.80)} & \shortstack[c]{$0.05 \pm 0.01$\\(0.04--0.08)} & \shortstack[c]{$0.22 \pm 0.08$\\(0.08--0.34)} \\
 & \shortstack[l]{\textbf{PCN}} & - & - & \textbf{-} & \shortstack[c]{$0.16 \pm 0.07$\\(0.05--0.29)} & \shortstack[c]{$0.29 \pm 0.10$\\(0.11--0.53)} &  & - & - & \textbf{-} & \shortstack[c]{$0.05 \pm 0.02$\\(0.03--0.08)} & \shortstack[c]{\textbf{\boldmath $0.16 \pm 0.07$}\\\textbf{(0.07--0.30)}} \\
 & \shortstack[l]{\textbf{ArtEq}} & - & - & \textbf{-} & - & - &  & \shortstack[c]{\underline{$8.23 \pm 5.24$}\\\underline{(0.35--24.88)}} & \shortstack[c]{$8.61 \pm 5.82$\\(1.21--21.63)} & \shortstack[c]{$0.37 \pm 0.05$\\(0.26--0.45)} & \shortstack[c]{$0.15 \pm 0.05$\\(0.08--0.26)} & \shortstack[c]{$0.42 \pm 0.13$\\(0.21--0.63)} \\
\midrule
\multirow{5}{*}{\rotatebox[origin=c]{90}{\textbf{Female (56)}}} & \shortstack[l]{\textbf{MedBOE}} & \shortstack[c]{\underline{$2.06 \pm 1.81$}\\\underline{(0.05--9.62)}} & \shortstack[c]{\textbf{\boldmath $0.93 \pm 0.72$}\\\textbf{(0.03--3.41)}} & \shortstack[c]{\underline{$0.92 \pm 0.01$}\\\underline{(0.85--0.94)}} & \shortstack[c]{\underline{$0.02 \pm 0.02$}\\\underline{(0.01--0.12)}} & \shortstack[c]{\underline{$0.09 \pm 0.10$}\\\underline{(0.02--0.55)}} & \multirow{5}{*}{\rotatebox[origin=c]{90}{\textbf{Female (25)}}} & \shortstack[c]{\textbf{\boldmath $2.15 \pm 1.70$ *}\\\textbf{(0.06--7.09)}} & \shortstack[c]{\textbf{\boldmath $0.85 \pm 0.50$}\\\textbf{(0.13--2.18)}} & \shortstack[c]{\textbf{\boldmath $0.94 \pm 0.01$}\\\textbf{(0.93--0.95)}} & \shortstack[c]{\underline{$0.03 \pm 0.01$}\\\underline{(0.02--0.07)}} & \shortstack[c]{\underline{$0.14 \pm 0.06$}\\\underline{(0.06--0.30)}} \\
 & \shortstack[l]{\textbf{IF-Net}} & \shortstack[c]{\textbf{\boldmath $1.41 \pm 1.45$}\\\textbf{(0.05--9.95)}} & \shortstack[c]{\underline{$2.51 \pm 2.22$}\\\underline{(0.00--13.89)}} & \shortstack[c]{\textbf{\boldmath $0.93 \pm 0.02$}\\\textbf{(0.83--0.95)}} & \shortstack[c]{\textbf{\boldmath $0.01 \pm 0.00$}\\\textbf{(0.01--0.03)}} & \shortstack[c]{\textbf{\boldmath $0.07 \pm 0.03$}\\\textbf{(0.02--0.16)}} &  & \shortstack[c]{$9.12 \pm 1.59$\\(6.78--11.75)} & \shortstack[c]{\underline{$2.96 \pm 1.05$}\\\underline{(1.00--5.30)}} & \shortstack[c]{\underline{$0.92 \pm 0.01$}\\\underline{(0.90--0.94)}} & \shortstack[c]{\textbf{\boldmath $0.03 \pm 0.01$}\\\textbf{(0.02--0.04)}} & \shortstack[c]{$0.30 \pm 0.05$\\(0.21--0.38)} \\
 & \shortstack[l]{\textbf{Shape}\\\textbf{Former}} & \shortstack[c]{$12.89 \pm 7.58$\\(0.95--33.25)} & \shortstack[c]{$16.12 \pm 11.83$\\(0.46--43.26)} & \shortstack[c]{$0.60 \pm 0.07$\\(0.40--0.72)} & \shortstack[c]{$0.06 \pm 0.02$\\(0.04--0.09)} & \shortstack[c]{$0.21 \pm 0.08$\\(0.07--0.39)} &  & \shortstack[c]{$13.70 \pm 5.00$\\(0.40--20.80)} & \shortstack[c]{$10.90 \pm 7.77$\\(0.84--24.47)} & \shortstack[c]{$0.71 \pm 0.06$\\(0.58--0.80)} & \shortstack[c]{$0.04 \pm 0.01$\\(0.03--0.06)} & \shortstack[c]{$0.18 \pm 0.08$\\(0.09--0.36)} \\
 & \shortstack[l]{\textbf{PCN}} & - & - & - & \shortstack[c]{$0.14 \pm 0.07$\\(0.04--0.33)} & \shortstack[c]{$0.25 \pm 0.07$\\(0.11--0.40)} &  & - & - & \textbf{-} & \shortstack[c]{$0.03 \pm 0.01$\\(0.02--0.06)} & \shortstack[c]{\textbf{\boldmath $0.09 \pm 0.03$ *}\\\textbf{(0.06--0.20)}} \\
 & \shortstack[l]{\textbf{ArtEq}} & - & - & - & - & - &  & \shortstack[c]{\underline{$9.08 \pm 6.23$}\\\underline{(0.14--23.33)}} & \shortstack[c]{$15.15 \pm 9.29$\\(1.94--35.92)} & \shortstack[c]{$0.43 \pm 0.06$\\(0.34--0.53)} & \shortstack[c]{$0.19 \pm 0.04$\\(0.11--0.26)} & \shortstack[c]{$0.46 \pm 0.07$\\(0.29--0.55)} \\
\botrule
\end{tabular*}
\footnotetext{Asterisks (*) indicate significance from the second-best results (Tukey's HSD, $p < 0.05$). Model count shown next to each gender.}
\end{sidewaystable}

For male-female comparisons, MedBOE showed no statistically significant differences across all AutoPET metrics (RBE: $p = 0.109$; RVE: $p = 0.849$; IoU: $p = 0.785$; CD: $p = 0.807$; HD: $p = 0.783$). In contrast, statistically significant male-female differences were observed in HD for IF-Net ($p = 0.007$) and ShapeFormer ($p = 0.023$), while PCN showed no statistically significant male-female differences in CD or HD ($p = 0.192$ and $p = 0.088$, respectively). Full p-values, Cohen's $d$, and 95\% CIs are reported in Appendix Table~6.

\begin{figure}[htbp]
\centering
\includegraphics[width=\linewidth]{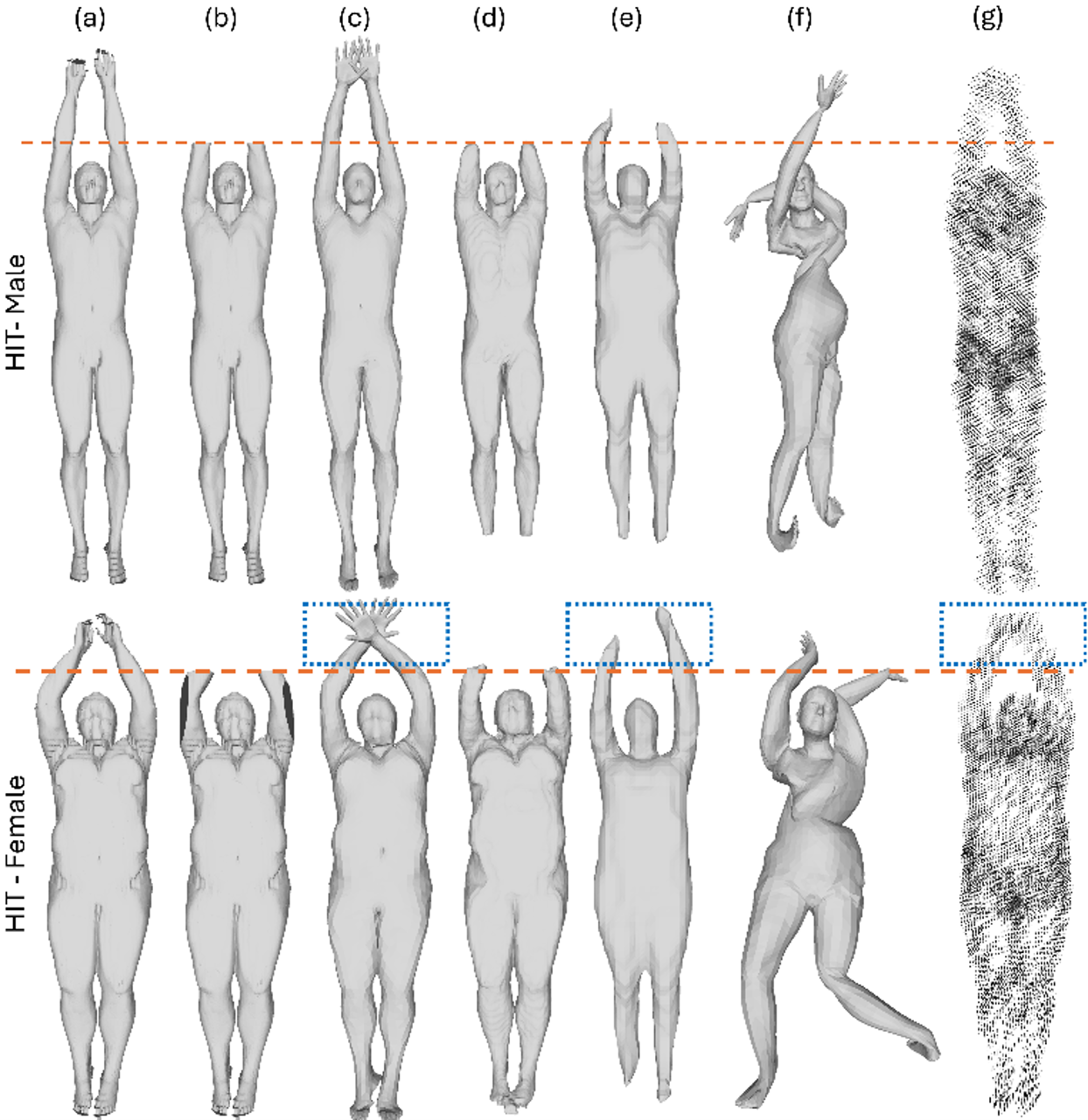}
\caption{Complete body models estimated from the HIT dataset. (a) is GT body models, (b) is simulated incomplete models where darker grey indicates the internal surface exposed by the trimming procedure. Results of (c) MedBOE, (d) IF-Net, (e) ShapeFormer, (f) ArtEq, and (g) PCN. Orange dashed lines mark the 10~cm above-head apex cutoff. Blue dotted boxes highlight differences in forearms and hands region estimation among MedBOE, ShapeFormer, and PCN}
\label{fig4}
\end{figure}

Estimated complete body models from the HIT dataset are presented in Fig.~\ref{fig4}, with metrics reported on Table~\ref{tab:results_comparison}. Zoomed views of the regions above the shoulders and around the legs, where method-specific differences are visible, are provided in Appendix Figs.~\ref{fig10} and~\ref{fig11}. A representative challenging case is presented in the Appendix Fig.~\ref{fig12}.

For HIT males, MedBOE achieved the top results in the volumetric metrics and CD, significantly outperforming IF-Net in RBE and RVE ($p = 1.16 \times 10^{-9}$ and $p = 0.008$, respectively). PCN had the lowest HD, with MedBOE ranking second. No significant differences were found between MedBOE and IF-Net in IoU and CD ($p = 0.251$ and $p = 0.988$, respectively) or between MedBOE and PCN in HD ($p = 0.918$). For HIT females, MedBOE showed the best volumetric results, significantly outperforming ArtEq in RBE ($p = 3.29 \times 10^{-7}$) and marginally surpassing IF-Net in RVE and IoU ($p = 0.610$ and $p = 0.618$, respectively). In geometric metrics, MedBOE ranked second behind IF-Net in CD, with no significant difference between the two methods ($p = 0.933$), and ranked second behind PCN in HD, with a significant difference between the two methods ($p = 0.035$). Full p-values, Cohen's $d$, and 95\% CIs for method comparisons are reported in Appendix Table~7.

For male-female comparisons, MedBOE showed no significant differences across all HIT metrics (RBE: $p = 0.622$; RVE: $p = 0.896$; IoU: $p = 0.123$; CD: $p = 0.112$; HD: $p = 0.074$). In contrast, significant male-female differences were observed across all evaluated metrics for IF-Net (all $p \leq 5.46 \times 10^{-4}$), in CD for ShapeFormer ($p = 0.022$), in CD and HD for PCN ($p = 0.010$ and $p = 0.003$, respectively), and in RVE and IoU for ArtEq ($p = 0.009$ and $p = 0.002$, respectively). Full p-values, Cohen's $d$, and 95\% CIs are reported in Appendix Table~8.

For the dataset-wise comparisons, MedBOE showed no statistically significant differences between AutoPET and HIT in RBE and RVE ($p = 0.543$ and $p = 0.557$, respectively), whereas statistically significant differences were observed in IoU, CD, and HD ($p = 1.65 \times 10^{-14}$, $p = 5.30 \times 10^{-5}$, and $p = 3.45 \times 10^{-5}$, respectively). IF-Net showed statistically significant differences in RBE, CD, and HD ($p = 1.57 \times 10^{-26}$, $p = 3.61 \times 10^{-14}$, and $p = 1.20 \times 10^{-31}$, respectively), while RVE and IoU showed no statistically significant differences ($p = 0.204$ and $p = 0.848$, respectively). ShapeFormer showed statistically significant differences in RVE, IoU, and CD ($p = 1.31 \times 10^{-4}$, $p = 1.18 \times 10^{-14}$, and $p = 1.39 \times 10^{-9}$, respectively), but not in RBE or HD ($p = 0.811$ and $p = 0.858$, respectively). PCN showed statistically significant differences in CD and HD ($p = 9.22 \times 10^{-26}$ and $p = 4.59 \times 10^{-20}$, respectively). When the absolute mean differences between AutoPET and HIT were compared, MedBOE showed smaller differences than IF-Net in RBE, RVE, CD, and HD, and smaller differences than ShapeFormer in RBE, RVE, IoU, and CD. Full p-values, Cohen's $d$, and 95\% CIs are reported in Appendix Table~9.

\subsection{Body Habitus Metrics}\label{subsec:results_bhm}

\begin{figure}[!htbp]
\centering
\includegraphics[width=0.7\linewidth]{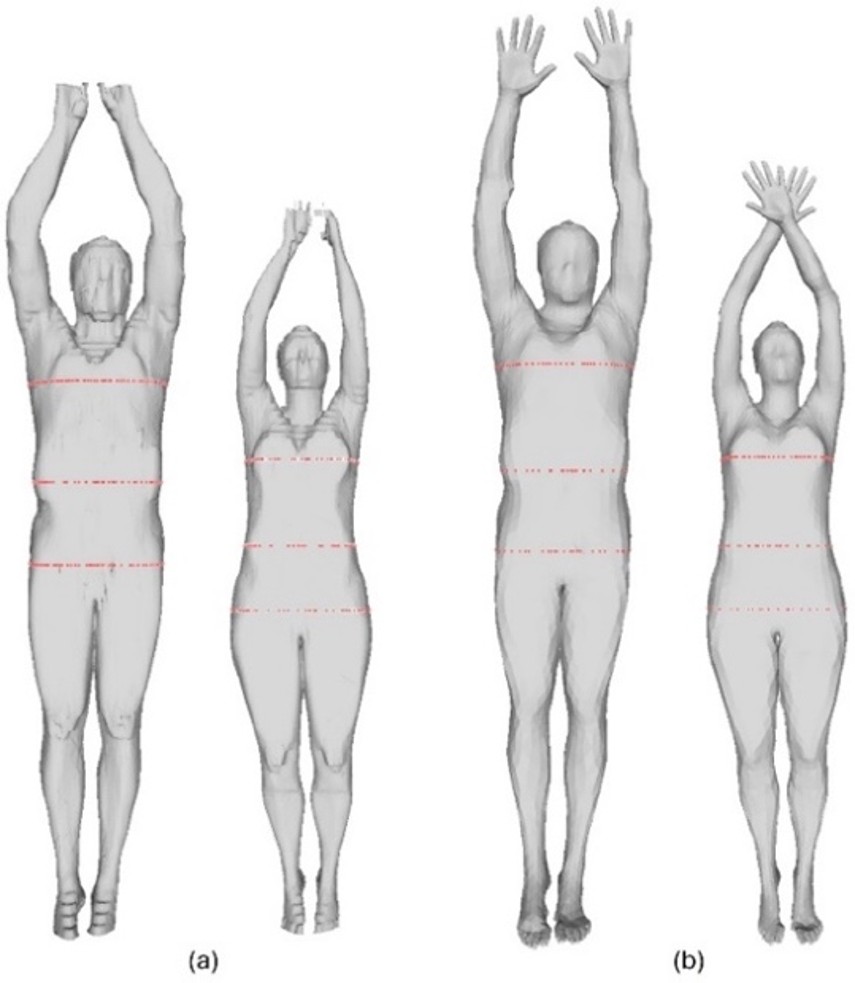}
\caption{Circumferential measurement locations (red dotted lines) for chest, waist, and hip, on (a) the GT and (b) the MedBOE-estimated complete body models}
\label{fig5}
\end{figure}

Quantitative evaluation of BHMs derived from MedBOE on the HIT dataset is summarized in Table~\ref{tab:results_bhm}. Mean absolute percentage errors for MedBOE were below 1\% for body volume and hip circumference, approximately 1.6\% for waist circumference, and around 2\% for body surface area and chest circumference. The mean absolute errors were $0.001 \pm 0.000$ $\mathrm{m}^{3}$ for body volume, $0.042 \pm 0.031$ $\mathrm{m}^{2}$ for body surface area, $0.021 \pm 0.017$ m for chest circumference, $0.016 \pm 0.012$ m for waist circumference, and $0.008 \pm 0.005$ m for hip circumference. Circumferential measurement locations on GT and MedBOE-estimated complete body models are shown in Fig.~\ref{fig5}.

\begin{table}[t]
\centering
\caption{BHM errors derived from the estimated complete body models of the HIT dataset.}
\label{tab:results_bhm}
\setlength{\tabcolsep}{3pt}
\renewcommand{\arraystretch}{1.15}
\begin{tabular*}{\textwidth}{@{\extracolsep{\fill}}lccccc@{}}
\toprule
\textbf{Metrics} & \textbf{Volume $\downarrow$} & \textbf{Surface area $\downarrow$} & \textbf{Chest cir. $\downarrow$} & \textbf{Waist cir. $\downarrow$} & \textbf{Hip cir. $\downarrow$} \\
\midrule
\textbf{APE} & \shortstack[c]{$0.85 \pm 0.46$\\(0.13--2.18)} & \shortstack[c]{$2.04 \pm 1.75$\\(0.01--7.27)} & \shortstack[c]{$2.05 \pm 1.61$\\(0.08--8.02)} & \shortstack[c]{$1.64 \pm 1.22$\\(0.12--5.67)} & \shortstack[c]{$0.77 \pm 0.42$\\(0.08--1.75)} \\[3pt]
\textbf{MAE} & \shortstack[c]{$0.001 \pm 0.000$\\(0.000--0.002)} & \shortstack[c]{$0.042 \pm 0.031$\\(0.004--0.138)} & \shortstack[c]{$0.021 \pm 0.017$\\(0.001--0.096)} & \shortstack[c]{$0.016 \pm 0.012$\\(0.001--0.056)} & \shortstack[c]{$0.008 \pm 0.005$\\(0.001--0.022)} \\[3pt]
\botrule
\end{tabular*}
\footnotetext{APE: absolute percentage error, MAE: mean absolute error, cir.: circumference.}
\end{table}

\subsection{Runtime Delay}\label{subsec:results_runtime}

Runtime-delay results were measured from an input incomplete patient body model to an estimated complete patient body model, excluding image segmentation and incomplete patient body model reconstruction. MedBOE required $1254.07 \pm 184.47$ seconds (s) per complete patient body model (813.56--1808.82~s). The baseline methods required shorter runtime delays: ArtEq required $0.32 \pm 0.07$ s (0.25--0.64~s), PCN required $0.95 \pm 0.17$ s (0.87--2.60~s), IF-Net required $6.74 \pm 0.07$ s (6.68--7.05~s), and ShapeFormer required $45.03 \pm 16.14$ s (18.76--103.67~s). MedBOE had a longer runtime delay than ArtEq, PCN, IF-Net, and ShapeFormer (Tukey's HSD, all $p < 1.00 \times 10^{-12}$).

\subsection{Ablation Study}\label{subsec:results_ablation}

The ablation results are presented in Table~\ref{tab:results_ablation}. Incorporating the missing region estimation and surface deformation module significantly improved performance across all metrics (paired t-test, $p < 0.05$), except for HD in the HIT dataset, where no significant difference was observed. Paired t-test p-values, paired effect sizes, and 95\% confidence intervals for the ablation study are reported in Appendix Table~10.

\begin{table}[t]
\centering
\caption{Performance of MedBOE with and without the missing body region estimation and surface deformation module. The better results are shown in bold.}
\label{tab:results_ablation}
\setlength{\tabcolsep}{2pt}
\renewcommand{\arraystretch}{2.4}
\begin{tabular*}{\textwidth}{@{\extracolsep{\fill}}llccccc@{}}
\toprule
 & \textbf{Dataset}
 & \textbf{RBE $\downarrow$}
 & \textbf{RVE $\downarrow$}
 & \textbf{IoU $\uparrow$}
 & \textbf{CD $\downarrow$}
 & \textbf{HD $\downarrow$} \\
\midrule
\multirow{4}{*}{\rotatebox[origin=c]{90}{\textbf{With the module}}}
 & \shortstack[c]{\textbf{AutoPET}\\\textbf{Male}}
 & \shortstack[c]{\textbf{\boldmath $1.47 \pm 1.61$ *}\\\textbf{(0.12--9.04)}}
 & \shortstack[c]{\textbf{\boldmath $0.90 \pm 0.80$ *}\\\textbf{(0.03--4.22)}}
 & \shortstack[c]{\textbf{\boldmath $0.92 \pm 0.03$ *}\\\textbf{(0.83--0.94)}}
 & \shortstack[c]{\textbf{\boldmath $0.02 \pm 0.02$ *}\\\textbf{(0.02--0.08)}}
 & \shortstack[c]{\textbf{\boldmath $0.10 \pm 0.10$ *}\\\textbf{(0.03--0.39)}} \\
 & \shortstack[c]{\textbf{AutoPET}\\\textbf{Female}}
 & \shortstack[c]{\textbf{\boldmath $2.06 \pm 1.81$ *}\\\textbf{(0.05--9.62)}}
 & \shortstack[c]{\textbf{\boldmath $0.93 \pm 0.72$ *}\\\textbf{(0.03--3.41)}}
 & \shortstack[c]{\textbf{\boldmath $0.92 \pm 0.01$ *}\\\textbf{(0.85--0.94)}}
 & \shortstack[c]{\textbf{\boldmath $0.02 \pm 0.02$ *}\\\textbf{(0.01--0.12)}}
 & \shortstack[c]{\textbf{\boldmath $0.09 \pm 0.10$ *}\\\textbf{(0.02--0.55)}} \\
 & \shortstack[c]{\textbf{HIT}\\\textbf{Male}}
 & \shortstack[c]{\textbf{\boldmath $1.85 \pm 1.87$ *}\\\textbf{(0.01--7.27)}}
 & \shortstack[c]{\textbf{\boldmath $0.86 \pm 0.40$ *}\\\textbf{(0.16--1.52)}}
 & \shortstack[c]{\textbf{\boldmath $0.94 \pm 0.01$ *}\\\textbf{(0.93--0.95)}}
 & \shortstack[c]{\textbf{\boldmath $0.04 \pm 0.01$ *}\\\textbf{(0.02--0.06)}}
 & \shortstack[c]{$0.19 \pm 0.09$\\(0.08--0.38)} \\
 & \shortstack[c]{\textbf{HIT}\\\textbf{Female}}
 & \shortstack[c]{\textbf{\boldmath $2.15 \pm 1.70$ *}\\\textbf{(0.06--7.09)}}
 & \shortstack[c]{\textbf{\boldmath $0.85 \pm 0.50$ *}\\\textbf{(0.13--2.18)}}
 & \shortstack[c]{\textbf{\boldmath $0.94 \pm 0.01$ *}\\\textbf{(0.93--0.95)}}
 & \shortstack[c]{\textbf{\boldmath $0.03 \pm 0.01$ *}\\\textbf{(0.02--0.07)}}
 & \shortstack[c]{$0.14 \pm 0.06$\\(0.06--0.30)} \\
\midrule
\multirow{4}{*}{\rotatebox[origin=c]{90}{\textbf{Without the module}}}
 & \shortstack[c]{\textbf{AutoPET}\\\textbf{Male}}
 & \shortstack[c]{$3.83 \pm 2.39$\\(0.65--13.87)}
 & \shortstack[c]{$5.93 \pm 2.45$\\(1.68--11.19)}
 & \shortstack[c]{$0.78 \pm 0.03$\\(0.70--0.83)}
 & \shortstack[c]{$0.03 \pm 0.02$\\(0.02--0.09)}
 & \shortstack[c]{$0.10 \pm 0.09$\\(0.04--0.39)} \\
 & \shortstack[c]{\textbf{AutoPET}\\\textbf{Female}}
 & \shortstack[c]{$4.23 \pm 2.93$\\(0.07--13.81)}
 & \shortstack[c]{$7.32 \pm 2.78$\\(0.02--15.51)}
 & \shortstack[c]{$0.77 \pm 0.03$\\(0.68--0.82)}
 & \shortstack[c]{$0.03 \pm 0.02$\\(0.02--0.14)}
 & \shortstack[c]{$0.10 \pm 0.09$\\(0.04--0.55)} \\
 & \shortstack[c]{\textbf{HIT}\\\textbf{Male}}
 & \shortstack[c]{$6.97 \pm 2.82$\\(2.94--13.10)}
 & \shortstack[c]{$11.36 \pm 3.17$\\(4.41--16.85)}
 & \shortstack[c]{$0.76 \pm 0.02$\\(0.72--0.79)}
 & \shortstack[c]{$0.05 \pm 0.01$\\(0.03--0.07)}
 & \shortstack[c]{$0.19 \pm 0.09$\\(0.08--0.38)} \\
 & \shortstack[c]{\textbf{HIT}\\\textbf{Female}}
 & \shortstack[c]{$6.42 \pm 2.91$\\(1.16--12.17)}
 & \shortstack[c]{$11.85 \pm 2.04$\\(5.70--15.63)}
 & \shortstack[c]{$0.75 \pm 0.03$\\(0.66--0.80)}
 & \shortstack[c]{$0.04 \pm 0.02$\\(0.01--0.09)}
 & \shortstack[c]{$0.14 \pm 0.06$\\(0.06--0.29)} \\
\botrule
\end{tabular*}
\footnotetext{An asterisk (*) indicates significance from the results without the module (paired t-test, $p < 0.05$).}
\end{table}

\section{Discussion}\label{sec:discussion}

Our main findings were that MedBOE: (i) achieved the highest volumetric accuracy (RBE, RVE, IoU) by accurately estimating missing regions; (ii) ranked a close second in geometric accuracy (CD and HD); (iii) maintained consistent performance across genders and datasets compared to other methods; (iv) accurately derived BHMs with low percentage errors; and (v) had a longer runtime delay than the baseline methods, but did not require dataset- or gender-specific training. Further, most baseline methods, except PCN, had notable limitations in estimating missing regions.

MedBOE consistently outperformed baseline methods in volumetric metrics across genders and datasets, except for RBE and IoU in the AutoPET female models. It achieved RBE $\leq 2.15\%$, RVE $\leq 1.01\%$, and IoU $\geq 0.92$ (Table~\ref{tab:results_comparison}), reflecting a high degree of spatial alignment between the estimated complete and GT body models. We note that volumetric performance can be more crucial than geometric performance for the estimation, as it directly impacts the accuracy of BHMs used in medical applications \cite{bib5,bib6}. Importantly, RBE and RVE values were lower than the proportions of trimmed body regions during the incomplete body model simulation in the AutoPET and HIT datasets, indicating effective estimation of missing regions.

The improved volumetric performance was primarily driven by MedBOE's accurate estimation of missing regions, which significantly outperformed the second-best methods in the volumetric metrics (Tukey's HSD, $p < 0.05$), with the exception of RBE for both HIT male and female models and IoU for HIT female models, where MedBOE still outperformed but only marginally (Appendix~\ref{sec:appendix_missing_regions}). This performance gain was attributed to the missing region estimation and surface deformation modules, which leverage the SMPL's integrated body shape knowledge while addressing the geometric discrepancies between the multimedia-derived SMPL meshes and incomplete body models from medical imaging. This module also preserved patient-specific geometric and volumetric details captured in the incomplete models. The ablation results (Table~\ref{tab:results_ablation}) support these findings, showing that the module significantly improved performance (paired t-tests, $p < 0.05$).

In geometric metrics, MedBOE was comparable, ranking second to IF-Net on AutoPET and to PCN on HIT, with only marginal differences (Tukey's HSD, $p > 0.05$). These outcomes likely reflect the small extent of missing body regions in AutoPET (1.485\% of GT body model) and the limited joint pose cues in HIT. Although MedBOE estimated these small regions significantly more accurately than IF-Net (Appendix~\ref{sec:appendix_missing_regions}; Tukey's HSD, $p < 0.05$), this gain was not fully reflected in the global geometric metrics (Table~\ref{tab:results_comparison}). MedBOE preserved anatomical continuity, whereas IF-Net often bridged gaps with near-planar patches that minimize nearest-point distances (Fig.~\ref{fig2} and~\ref{fig3}). Chamfer and Hausdorff distances quantify nearest-point distances to the GT surface (Chamfer averages bidirectional distances; Hausdorff reports the maximum) and do not enforce anatomical correspondence. Consequently, surfaces that simply span cut edges can lie closer to the nearest points on the GT mesh yet remain anatomically implausible. For example, in Fig.~\ref{fig3}, the distance from the inferior elbow surface in IF-Net's estimate to the GT model is shorter than for MedBOE's anatomically continuous estimate, even though the latter better reflects elbow morphology. This effect is most pronounced when missing regions are small, as in AutoPET.

On the HIT dataset, MedBOE's geometric performance was influenced by ambiguity in joint poses estimation, particularly at the elbows and wrists, due to entirely missing regions lacking anatomical references (Fig.~\ref{fig4}(b)). Unlike AutoPET, which retained partial arm and forearm geometry aiding elbow inference, HIT provided no such cues. As a result, MedBOE's joint pose estimates showed slight deviations from the GT models (Fig.~\ref{fig4}(c)), while PCN and ShapeFormer produced closer estimates (Fig.~\ref{fig4}(g) and (e)). This is likely because those models learned joint configurations directly from the HIT dataset through dataset-specific training, whereas MedBOE relied on $E_{\mathrm{Prior}}$ pre-trained on larger multimedia datasets with more diverse human poses. The mismatch between the HIT-specific poses and those in the pre-training datasets likely contributed to these deviations. Although retaining (e.g., fine tuning) on HIT could improve accuracy, MedBOE prioritized generalizability across datasets. Exclusive training on HIT could result in greater deviations on AutoPET, which features different patient poses. By leveraging pre-trained models, MedBOE maintained consistent performance across both datasets and is expected to generalize well to new datasets.

MedBOE demonstrated robust and consistent performance across genders and datasets. MedBOE showed no statistically significant differences ($p > 0.05$) across all metrics in both AutoPET and HIT. In contrast, the baseline methods showed statistically significant male-female differences ($p < 0.05$) in HD on AutoPET and in several metrics on HIT, including all evaluated metrics for IF-Net, while Baseline methods, except ArtEq, generally performing worse in estimating complete male patient models. The MedBOE's stability is likely attributed to its independence from training data. The gender-based variation in the baseline methods is presumed to stem partly from imbalanced gender representation in the training datasets (male-to-female ratios: 28:44 in AutoPET and 136:201 in HIT). In the dataset-wise comparisons, MedBOE showed no statistically significant differences in RBE and RVE, while differences were observed in IoU, CD, and HD. However, the absolute mean differences between AutoPET and HIT were smaller for MedBOE than for the baseline methods in most evaluated metrics, suggesting lower dataset-dependent variation. MedBOE also maintained relatively consistent performance even on HIT, despite its significantly larger missing regions than AutoPET (1.485 vs. 2.740\%; unpaired t-test, $p < 0.05$). IF-Net's performance declined on HIT, likely due to difficulties in recovering these larger missing body regions. ShapeFormer and PCN improved on HIT, which is likely due to the larger training sets in HIT (male: 28 vs. 136; female: 44 vs. 201).

The baseline methods demonstrated several limitations in estimating missing body regions and complete patient body models. PCN, despite generating well-aligned dense point clouds, lacked surfaces information required for volumetric evaluation. IF-Net underperformed when estimating large missing body regions and could not estimate them adequately. ShapeFormer failed to capture detailed geometric features and frequently produced outputs with unresolved missing body regions. ArtEq, although designed to generalize to unseen poses, was unable to accurately estimate complete patient body models. In contrast, MedBOE addressed these shortcomings by leveraging SMPL-integrated body shape knowledge and attribute-based optimization in the global parametric fitting and missing region estimation and surface deformation modules.

MedBOE demonstrated high accuracy in deriving BHMs and importantly, its absolute errors fall within clinically acceptable ranges. For instance, differences of up to 0.5~$\mathrm{m}^{2}$ in body surface area are considered acceptable among commonly used clinical estimation formulas \cite{bib40}, and errors, such as $0.001 \pm 0.000$~$\mathrm{m}^{3}$ in volume and $0.016 \pm 0.012$~m in waist circumference, are within tolerances reported in anthropometric studies \cite{bib41,bib42}.

Runtime-delay analysis indicates a trade-off between training-free case-wise optimization and computational efficiency. Despite requiring no dataset- or gender-specific training, MedBOE was significantly slower than all baseline methods (Tukey's HSD, all $p < 1.00 \times 10^{-12}$). This is expected because MedBOE performs iterative global parametric fitting and surface deformation for each input model, while the learning-based baselines perform inference using trained networks. In particular, runtime increased when $E_{\mathrm{Contact}}$ was enabled during global parametric fitting, due to its heavy computational cost for determining self-intersections at each vertex. The runtime-delay experiment was performed on an NVIDIA TITAN V GPU because PCN was not compatible with newer GPUs in our experimental environment. This hardware constraint should be considered when interpreting the measured runtime. For example, the NVIDIA GeForce RTX 5090 provides approximately seven-fold higher peak FP32 performance than TITAN V~\cite{nvidia2025blackwell,nvidiatitanv}.

\subsection{Limitations and Future Work}\label{subsec:limitations_future_work}

MedBOE has several limitations. First, MedBOE relies on pre-trained models to fit the SMPL mesh to incomplete patient body models; however, these pre-trained models were originally trained on multimedia datasets, differed from medical imaging datasets like AutoPET and HIT. This discrepancy may have introduced challenges in achieving optimal fitting accuracy on incomplete patient body models derived from whole-body medical images. Future research will explore domain-specific fitting models trained on medical imaging datasets, for example, by curating a whole-body imaging dataset that includes complete patient bodies with full external-body coverage in a supine pose, a diverse range of patient poses, and annotated body landmarks from the new extended FoV scanners (uEXPLORER and Biograph Vision Quadra).

Second, MedBOE optimizes global SMPL parameters, and the degrees of freedom for aligning the SMPL mesh are therefore limited by the SMPL shape and pose parameterization. This can introduce minor morphological variations in centered regions, such as the abdomen or thorax, observed in the incomplete patient body models while aligning the SMPL mesh. In addition, the final fine-deformation stage is intentionally constrained to stabilize the surface and prevent unrealistic shape changes, which means that it may not fully deform the aligned SMPL mesh in all local regions. Despite these constraints, MedBOE achieved higher overall accuracy than the baseline methods. Future work will investigate local deformation constraints that better preserve the observed body surface.

Third, the incomplete patient body models were simulated by virtually trimming GT body models. This design allowed controlled quantitative comparison while avoiding potential misalignment between PET-derived GT models and CT-derived incomplete models. However, it cannot fully reproduce all sources of real clinical incompleteness, including case-specific FoV geometry, PET-CT misregistration, and modality-specific artifacts. In addition, AutoPET evaluation was limited to body regions present in the PET-derived GT models because parts of the forearms and hands were often outside the acquisition range; therefore, AutoPET metrics do not reflect complete accuracy over all body regions. Appendix Fig.~6 compares the simulated trimmed model with the CT-derived body model, supporting the relevance of the simulation setting. Future work will investigate the performance of MedBOE using paired routine and wide-transverse-FoV CT images, where routine CT provides naturally incomplete patient body models and wide-transverse-FoV CT provides reference complete patient body models for quantitative evaluation.

Although total-body PET-CT scanners, such as uEXPLORER and Biograph Vision Quadra, provide extended axial coverage, this corresponds to the patient's head-to-toe direction and does not necessarily address missing body regions caused by limited transverse CT FoV. Depending on patient body size and positioning, hands, forearms, or elbows can remain outside the reconstructed CT body contour. Wide-transverse-FoV CT protocols can capture a more complete external body contour, but they are not required in routine clinical imaging. During this study, we could not identify a publicly available full-body, wide-transverse-FoV CT dataset suitable for this evaluation. Therefore, AutoPET was used because it reflects common clinical PET-CT acquisition, and HIT was used to enable evaluation with near-complete patient body coverage. Future studies will evaluate MedBOE using paired routine and wide-transverse-FoV CT acquisitions when available.

Additionally, MedBOE used Deep Virtual Marker \cite{bib23} to annotate sparse and dense body landmarks during body model reconstruction (Section~\ref{subsec:body_reconstruction}). However, its performance declined when body parts were in contact (e.g., arms in contact with the torso; Appendix Fig.~\ref{fig9}), leading to less reliable landmark annotations and complicated SMPL fitting. To address this, future work will include investigating deep learning-based models (e.g., PaMIR \cite{bib43}) to improve fitting robustness in complex poses.

\section{Conclusion}\label{sec:conclusion}

This study introduced MedBOE, a method for estimating complete patient body models from whole-body medical images with missing body regions. By integrating the SMPL template mesh and attribute-based optimization strategies tailored for medical imaging, MedBOE achieved the best overall performance across volumetric and geometric metrics, superior or comparable to the state-of-the-art methods. MedBOE-derived BHMs exhibited errors within clinically accepted tolerances, supporting their reliability for downstream analyses. Because the primary evaluation used incomplete patient body models simulated by virtual trimming, further validation using clinical acquisitions with real FoV-related incompleteness is needed to clarify generalization to clinical scenarios. These results demonstrate MedBOE's robustness and generalizability across genders and datasets, as well as its potential applicability for medical imaging applications that require complete patient body models.

\backmatter

\clearpage

\begin{appendices}

\renewcommand{\thefigure}{\arabic{figure}}
\setcounter{figure}{5}
\renewcommand{\thetable}{\arabic{table}}
\setcounter{table}{4}

\section{Attribute Formulations and Optimization Settings}\label{sec:appendix_attributes}

Detailed formulations of the attributes used in the global parametric fitting and missing body region estimation and surface deformation modules are provided below.

\subsection{Global Parametric Fitting Attributes}\label{subsec:appendix_global_attributes}

The global parametric fitting module optimizes the weighted sum of sparse landmark distance ($E_{\mathrm{Sparse}}$), surface distance ($E_{\mathrm{Dist}}$), pose prior ($E_{\mathrm{Prior}}$), weight consistency ($E_{\mathrm{Weight}}$), self-intersection penalty ($E_{\mathrm{Contact}}$), and dense landmark distance ($E_{\mathrm{Dense}}$), as defined in Eq.~\eqref{eq:global_fitting} of the manuscript.

$E_{\mathrm{Sparse}}$ penalizes the discrepancy between sparse landmarks $k_i$ on the incomplete patient body model and their correspondences $K_i$ on the SMPL mesh, weighted by confidence scores $w_i$ of $k_i$ from Deep Virtual Marker. Only landmarks with $w_i > 0.5$ were included:
\begin{equation}
E_{\mathrm{Sparse}} = \sum_{i=1}^{n} w_i^2 \cdot \lVert k_i - K_i \rVert_2^2.
\label{eq:appendix_sparse}
\end{equation}

$E_{\mathrm{Dist}}$ is a modified symmetric Chamfer distance between SMPL vertices $V$ and nearest vertices $v$ on the incomplete patient body model:
\begin{equation}
E_{\mathrm{Dist}} = 0.5 \times \left(\sum_{j=1}^{|V|} p_j \cdot \lVert V_j - v_j \rVert_2^2 + \sum_{k=1}^{|v|} p_k \cdot \lVert v_k - V_k \rVert_2^2\right),
\label{eq:appendix_dist}
\end{equation}
where $p$ masks unreliable correspondences and is set to zero when: (i) the Euclidean distance exceeds a predefined threshold, (ii) the angular difference between surface normals exceeds a threshold, or (iii) $v_x$ belongs to the head while $V_x$ belongs to an arm. The third condition prevents misalignment of SMPL arm vertices to the head region of the incomplete patient body model when the arm segments are absent.

$E_{\mathrm{Prior}}$ regularizes SMPL pose parameter $\theta$ using VPoser~\cite{bib26} and MaxMixturePrior~\cite{bib27}. VPoser encoded $\theta$ into a latent representation and reconstructed it to $\hat{\theta}$; the mean squared error (MSE) between $\theta$ and $\hat{\theta}$ was calculated. MaxMixturePrior calculated the negative log-likelihood of $\theta$ using a pre-trained Gaussian Mixture Model (GMM). $E_{\mathrm{Prior}}$ was computed:
\begin{equation}
E_{\mathrm{Prior}} = 0.5 \cdot \mathrm{MSE}(\theta,\hat{\theta}) + 100 \cdot \mathrm{GMM}_{\mathrm{MaxMixturePrior}}(\theta).
\label{eq:appendix_prior}
\end{equation}

$E_{\mathrm{Weight}}$ constrains SMPL shape parameter $\beta$ by matching the patient's estimated body weight, $w_{\mathrm{est}}$, to the patient's reference body weight, $w_{\mathrm{actual}}$. The $w_{\mathrm{est}}$ was predicted by BetaNet~\cite{bib24}, a two-layer fully connected network that predicts weight from $\beta$ and gender. $E_{\mathrm{Weight}}$ was computed:
\begin{equation}
E_{\mathrm{Weight}} = \left|w_{\mathrm{est}} - w_{\mathrm{actual}}\right|^2.
\label{eq:appendix_weight}
\end{equation}

$E_{\mathrm{Contact}}$ applies a repulsion loss~\cite{bib28} to SMPL's intersecting vertices $(v_c,v_c')$ and normals $(n_c,n_c')$:
\begin{equation}
E_{\mathrm{Contact}} = \sum_c \left(\lVert n_c - n_c' \rVert_2^2 + \lVert v_c - v_c' \rVert_2^2\right).
\label{eq:appendix_contact}
\end{equation}
Vertices were considered intersecting if Euclidean distance was $<2$~cm while their geodesic distance exceeded 30~cm~\cite{bib28}.

$E_{\mathrm{Dense}}$ penalizes discrepancy between incomplete model vertex $p_k$ and SMPL mesh vertex $P_k$ with closest 99-dimensional feature vector of $p_k$:
\begin{equation}
E_{\mathrm{Dense}} = \sum_{k=1}^{N} r_k \cdot \lVert p_k - P_k \rVert_2^2,
\label{eq:appendix_dense}
\end{equation}
where $r_k$ excludes poor spatial or angular agreements according to $E_{\mathrm{Dist}}$ criteria (i) and (ii).

\subsection{Missing Body Region Estimation and Surface Deformation Attributes}\label{subsec:appendix_deformation_attributes}

The coarse stage deforms the fitted SMPL mesh to the incomplete patient body model using sampled surface distance ($E_{\mathrm{Surf}}$), edge length ($E_{\mathrm{Edge}}$), Laplacian smoothness~\cite{bib29} ($E_{\mathrm{Lap}}$), and normal consistency ($E_{\mathrm{Normal}}$), as defined in Eq.~\eqref{eq:coarse_deformation} of the manuscript. A uniformly sampled set of 6,000 surface points from the fitted SMPL mesh ($P_{\mathrm{SMPL}}$) and the incomplete patient body model ($P_{\mathrm{Incomp}}$) was used, with $V_i$ and $V_i'$ denoting vertex positions before and after deformation, respectively.

$E_{\mathrm{Surf}}$ minimizes the squared distance between the sampled point sets $x$ and $y$, retaining only pairs within a Euclidean distance of 0.3~m and with a surface-normal angular difference below $90^\circ$:
\begin{equation}
E_{\mathrm{Surf}} = \frac{1}{|x|}\sum_{i=1}^{|x|} \min_k \lVert x_i - y_k \rVert_2^2 + \frac{1}{|y|}\sum_{k=1}^{|y|} \min_i \lVert y_k - x_i \rVert_2^2.
\label{eq:appendix_surf}
\end{equation}

$E_{\mathrm{Edge}}$ preserves local geometry by penalizing changes in edge lengths:
\begin{equation}
E_{\mathrm{Edge}} = \sum_{(i,j)\in\varepsilon} \left(\lVert V_i' - V_j' \rVert - \lVert V_i - V_j \rVert\right)^2,
\label{eq:appendix_edge}
\end{equation}
where $\varepsilon\subset V\times V$ denotes the set of edges.

$E_{\mathrm{Lap}}$~\cite{bib29} enforces local surface smoothness by encouraging each vertex to remain near the centroid of its one-ring neighbors $N(i)$:
\begin{equation}
E_{\mathrm{Lap}} = \sum_{i=1}^{|V|} \left\lVert V_i' - \frac{1}{|N(i)|}\sum_{j\in N(i)} V_j' \right\rVert^2.
\label{eq:appendix_lap}
\end{equation}

$E_{\mathrm{Normal}}$ promotes geometric continuity by penalizing angular deviations between adjacent face normals. Let $F$ be the set of faces in the fitted SMPL mesh, and $N(f_i)\subset F$ denote the set of faces adjacent to face $f_i$:
\begin{equation}
E_{\mathrm{Normal}} = \sum_{i=1}^{|F|} \left(\sum_{f'\in N(f_i)} \left(1 - n_{f_i}\cdot n_{f'}\right)\right).
\label{eq:appendix_normal}
\end{equation}

The fine stage deforms the fitted SMPL mesh to the initial complete patient body model using Chamfer distance ($E_{\mathrm{CD}}$), Chamfer normal ($E_{\mathrm{CN}}$), volume consistency ($E_{\mathrm{Vol}}$), $E_{\mathrm{Lap}}$, and $E_{\mathrm{Normal}}$, as defined in Eq.~\eqref{eq:fine_deformation} of the manuscript.

$E_{\mathrm{CD}}$ minimizes geometric disparity between the fitted SMPL mesh ($P_{\mathrm{SMPL}}$) and the initial complete patient body model ($P_{\mathrm{Pre}}$) by computing the average squared bidirectional distance. Here, $x\in P_{\mathrm{SMPL}}$ and $y\in P_{\mathrm{Pre}}$ are vertices:
\begin{equation}
E_{\mathrm{CD}} = \frac{1}{|P_{\mathrm{SMPL}}|}\sum_{i=1}^{|P_{\mathrm{SMPL}}|} \min_k \lVert x_i - y_k \rVert^2 + \frac{1}{|P_{\mathrm{Pre}}|}\sum_{k=1}^{|P_{\mathrm{Pre}}|} \min_i \lVert y_k - x_i \rVert^2.
\label{eq:appendix_cd}
\end{equation}

$E_{\mathrm{CN}}$ promotes surface orientation consistency by penalizing angular differences between surface normals. Here, $n_x$ and $n_y$ are unit normal vectors at $x$ and $y$:
\begin{equation}
E_{\mathrm{CN}} = \frac{1}{|P_{\mathrm{SMPL}}|}\sum_{i=1}^{|P_{\mathrm{SMPL}}|} \min_k \left\lVert 1 - n_{x_i}\cdot n_{y_k}\right\rVert^2 + \frac{1}{|P_{\mathrm{Pre}}|}\sum_{k=1}^{|P_{\mathrm{Pre}}|} \min_i \left\lVert 1 - n_{y_k}\cdot n_{x_i}\right\rVert^2.
\label{eq:appendix_cn}
\end{equation}

$E_{\mathrm{Vol}}$ preserves the fitted SMPL mesh volume ($\mathrm{Vol}_{\mathrm{SMPL}}$), computed via signed tetrahedral integration, by minimizing difference to deformed SMPL mesh volume ($\mathrm{Vol}_{\mathrm{SMPL}}'$):
\begin{equation}
E_{\mathrm{Vol}} = \left|\mathrm{Vol}_{\mathrm{SMPL}} - \mathrm{Vol}_{\mathrm{SMPL}}'\right|.
\label{eq:appendix_vol}
\end{equation}

\section{Incomplete Patient Body Models from Simulated Trimming and CT Images}\label{sec:appendix_trimming}

\begin{figure}[htbp]
\centering
\includegraphics[width=0.9\linewidth]{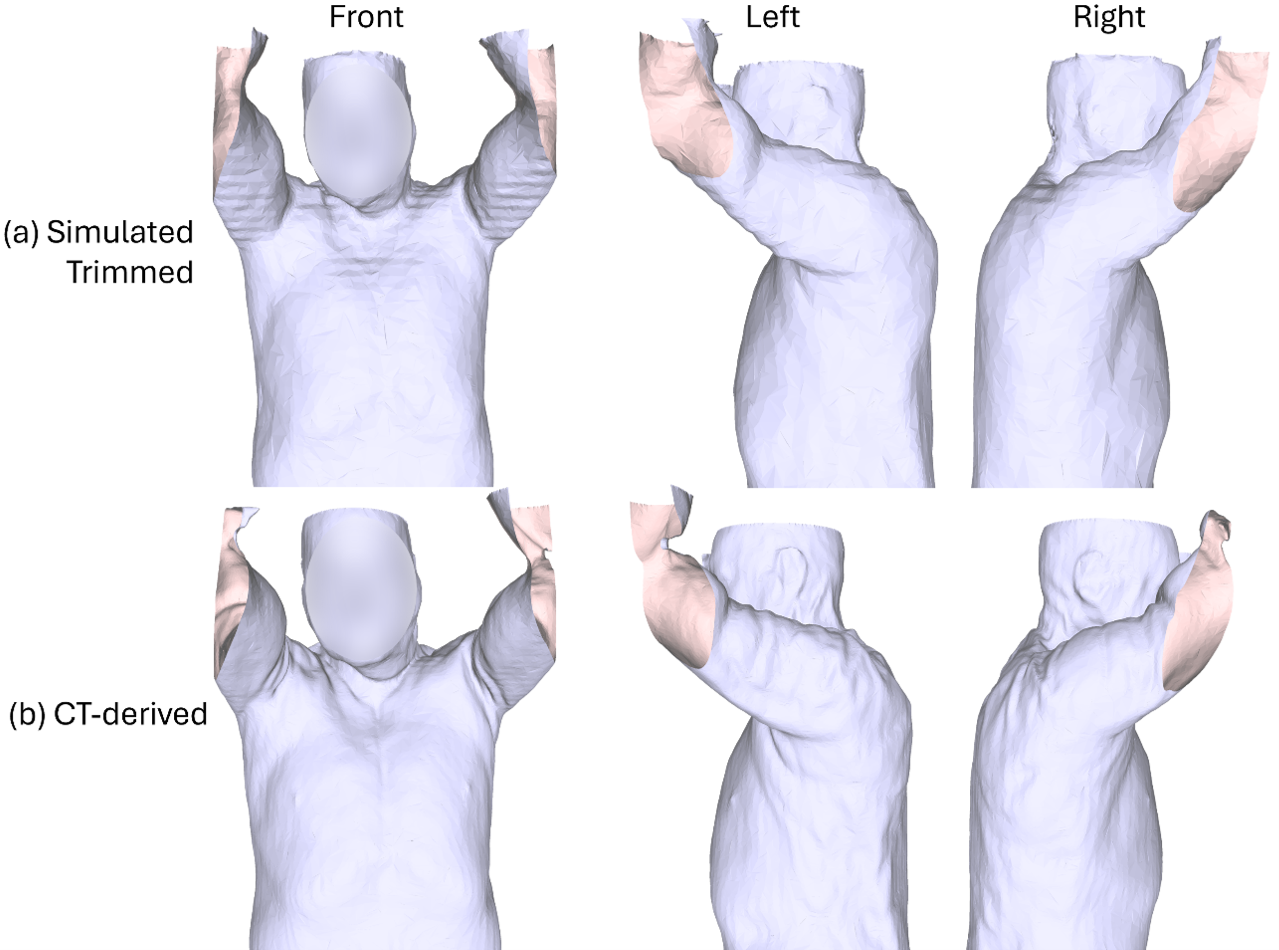}
\caption{Representative front, left, and right views (left-to-right) comparing (a) the simulated trimmed incomplete patient body model and (b) the corresponding CT-derived incomplete patient body model. External surfaces are rendered in blue, and internal cut surfaces exposed by truncation are rendered in red}
\label{fig6}
\end{figure}

Fig.~\ref{fig6} compares front, left, and right views of a simulated trimmed incomplete patient body model (Fig.~\ref{fig6}(a)) and the corresponding CT-derived incomplete patient body model (Fig.~\ref{fig6}(b)). The simulated trimmed model was generated from the PET-derived GT body model using the simulated transverse FoV constraint described in Section~\ref{subsubsec:autopet_dataset}. External surfaces are rendered in blue, and internal surfaces exposed by truncation are rendered in red. The comparison supports the relevance of simulated trimming for approximating FoV-related incompleteness in routine PET-CT acquisitions.

\section{Results of Estimated Complete Body Models}\label{sec:appendix_complete_results}

\subsection{Zoomed Comparison of AutoPET Estimated Complete Patient Body Models}\label{subsec:appendix_autopet_zoom}

\begin{figure}[!htbp]
\centering
\includegraphics[width=\linewidth]{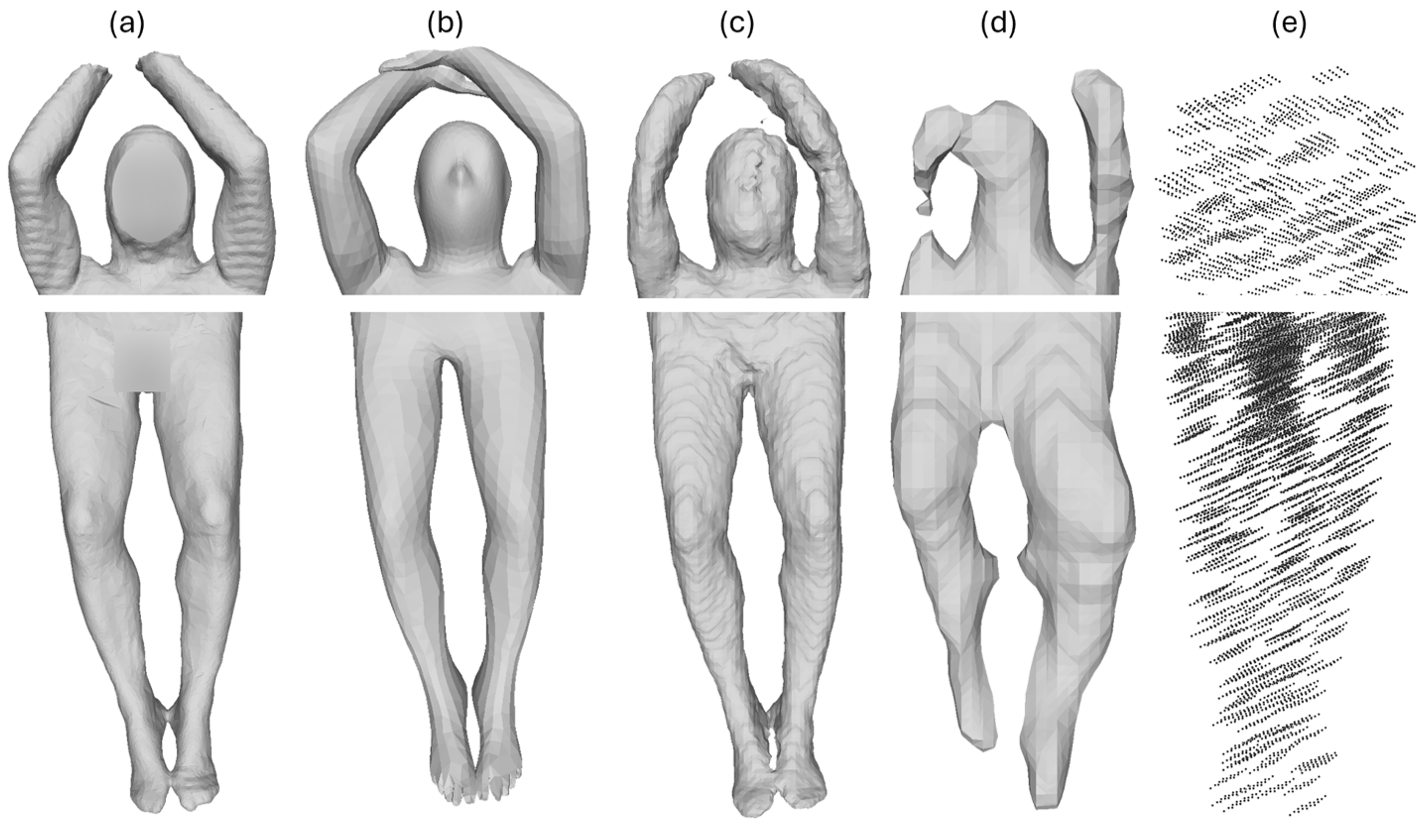}
\caption{Zoomed views of the AutoPET male complete body model estimates. The top row shows the region above the shoulders, and the bottom row shows the leg region. (a) GT body model, (b) MedBOE, (c) IF-Net, (d) ShapeFormer, and (e) PCN}
\label{fig7}
\end{figure}

\begin{figure}[htbp]
\centering
\includegraphics[width=0.95\linewidth]{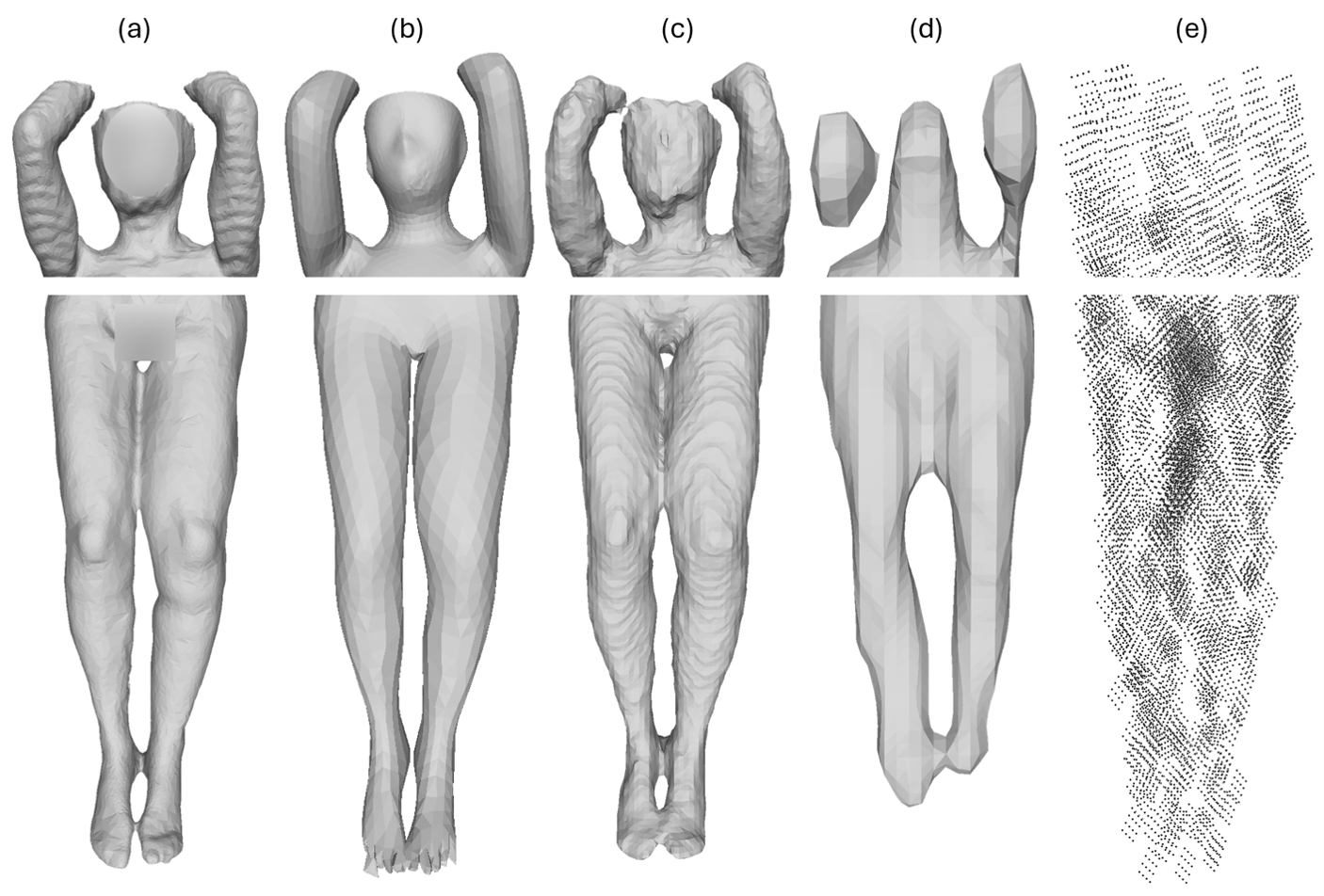}
\caption{Zoomed views of the AutoPET female complete body model estimates. The top row shows the region above the shoulders, and the bottom row shows the leg region. (a) GT body model, (b) MedBOE, (c) IF-Net, (d) ShapeFormer, and (e) PCN}
\label{fig8}
\end{figure}

Fig.~\ref{fig7} and~\ref{fig8} provide zoomed views of the regions above the shoulders (top) and around the legs (bottom) from the AutoPET male and female examples, respectively. These regions were selected because visible differences among method estimates were observed in the full-body views. The zoomed views provide additional visual detail for comparing surface continuity and local body-shape estimation across methods, particularly around missing body regions. PCN is shown as point-cloud output.

\subsection{Representative Challenging Case from AutoPET}\label{subsec:appendix_autopet_challenging}

\begin{figure}[htbp]
\centering
\includegraphics[width=\linewidth]{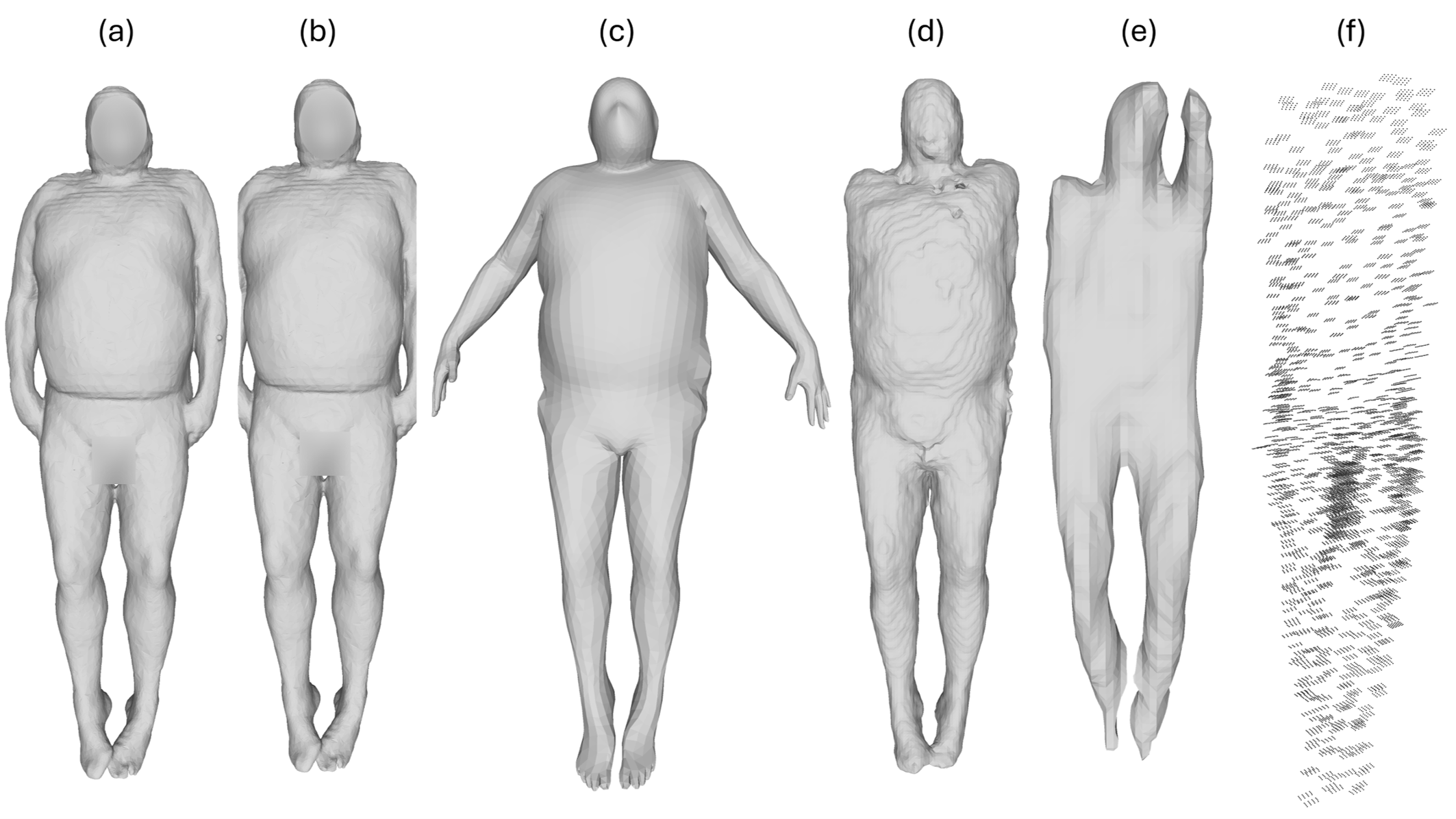}
\caption{Representative challenging case from the AutoPET dataset. (a) GT body model, (b) incomplete patient body model, (c) MedBOE, (d) IF-Net, (e) ShapeFormer, and (f) PCN}
\label{fig9}
\end{figure}

Fig.~\ref{fig9} shows a representative challenging case from the AutoPET dataset. The patient had a large body volume and was scanned with the arms positioned alongside the body, causing parts of the lateral arms to be attached to the torso in the incomplete patient body model. This made the arm regions difficult to distinguish from the trunk for all methods. MedBOE estimated complete arm regions, although the estimated arm pose was inaccurate. IF-Net mainly closed the missing surface regions without reconstructing anatomically complete arms, while ShapeFormer and PCN appeared to estimate arms-up configurations and did not fully recover the missing arm regions.

\subsection{Statistical Analysis of AutoPET Estimated Complete Patient Body Models}\label{subsec:appendix_autopet_statistics}

Table~\ref{tab:appendix5} and~\ref{tab:appendix6} provide detailed statistical results for the AutoPET complete patient body model estimates. Pairwise method-comparison statistics (Table~\ref{tab:appendix5}) are reported from Tukey HSD tests, including $p$-values, Cohen's $d$, and 95\% confidence intervals (CIs).

\begingroup
\makeatletter
\def\LT@makecaption#1#2#3{%
  \LT@mcol\LT@cols c{\hbox to\z@{\hss%
    \parbox[t]{\textwidth}{\@tablecaption{#1{#2}}{#3}\vskip\belowcaptionskip}%
  \hss}}}
\makeatother
\footnotesize
\setlength{\tabcolsep}{3pt}
\renewcommand{\arraystretch}{1.15}
\begin{longtable}{@{\extracolsep{\fill}}lllcccc@{}}
\caption{Pairwise method-comparison statistics from Tukey HSD tests for the AutoPET dataset.}\label{tab:appendix5} \\
\toprule
\textbf{Gender} & \textbf{Metric} & \textbf{Comparison} & \textbf{\shortstack{Mean\\difference}} & \textbf{$p$-value} & \textbf{Cohen's $d$} & \textbf{95\% CI} \\
\midrule
\endfirsthead
\caption[]{Pairwise method-comparison statistics from Tukey HSD tests for the AutoPET dataset. (continued)} \\
\toprule
\textbf{Gender} & \textbf{Metric} & \textbf{Comparison} & \textbf{\shortstack{Mean\\difference}} & \textbf{$p$-value} & \textbf{Cohen's $d$} & \textbf{95\% CI} \\
\midrule
\endhead
\midrule
\multicolumn{7}{r}{\footnotesize Continued on next page} \\
\endfoot
\botrule
\endlastfoot
Male & RBE & MedBOE - IF-Net & $-1.817$ & $0.439$ & $-0.400$ & $[-5.334, 1.699]$ \\*
 &  & ShapeFormer - IF-Net & $11.474$ & $2.02 \times 10^{-11}$ & $1.532$ & $[7.957, 14.991]$ \\*
 &  & ShapeFormer - MedBOE & $13.291$ & $<1.00 \times 10^{-12}$ & $2.154$ & $[9.774, 16.808]$ \\
 & RVE & MedBOE - IF-Net & $-2.920$ & $0.338$ & $-0.487$ & $[-7.833, 1.993]$ \\*
 &  & ShapeFormer - IF-Net & $15.539$ & $6.51 \times 10^{-11}$ & $1.470$ & $[10.626, 20.452]$ \\*
 &  & ShapeFormer - MedBOE & $18.459$ & $<1.00 \times 10^{-12}$ & $2.112$ & $[13.546, 23.372]$ \\
 & IoU & MedBOE - IF-Net & $0.008$ & $0.854$ & $0.138$ & $[-0.029, 0.046]$ \\*
 &  & ShapeFormer - IF-Net & $-0.331$ & $<1.00 \times 10^{-12}$ & $-4.213$ & $[-0.368, -0.293]$ \\*
 &  & ShapeFormer - MedBOE & $-0.339$ & $<1.00 \times 10^{-12}$ & $-6.093$ & $[-0.377, -0.302]$ \\
 & CD & MedBOE - IF-Net & $0.009$ & $0.719$ & $0.782$ & $[-0.014, 0.033]$ \\*
 &  & PCN - IF-Net & $0.144$ & $<1.00 \times 10^{-12}$ & $2.842$ & $[0.120, 0.167]$ \\*
 &  & ShapeFormer - IF-Net & $0.045$ & $9.98 \times 10^{-6}$ & $3.666$ & $[0.022, 0.068]$ \\*
 &  & PCN - MedBOE & $0.134$ & $<1.00 \times 10^{-12}$ & $2.603$ & $[0.111, 0.157]$ \\*
 &  & ShapeFormer - MedBOE & $0.035$ & $7.05 \times 10^{-4}$ & $2.220$ & $[0.012, 0.059]$ \\*
 &  & ShapeFormer - PCN & $-0.099$ & $<1.00 \times 10^{-12}$ & $-1.914$ & $[-0.122, -0.075]$ \\
 & HD & MedBOE - IF-Net & $0.009$ & $0.972$ & $0.118$ & $[-0.042, 0.059]$ \\*
 &  & PCN - IF-Net & $0.201$ & $<1.00 \times 10^{-12}$ & $2.667$ & $[0.151, 0.252]$ \\*
 &  & ShapeFormer - IF-Net & $0.086$ & $1.33 \times 10^{-4}$ & $1.502$ & $[0.035, 0.136]$ \\*
 &  & PCN - MedBOE & $0.193$ & $<1.00 \times 10^{-12}$ & $1.922$ & $[0.142, 0.244]$ \\*
 &  & ShapeFormer - MedBOE & $0.077$ & $7.15 \times 10^{-4}$ & $0.883$ & $[0.026, 0.128]$ \\*
 &  & ShapeFormer - PCN & $-0.116$ & $1.36 \times 10^{-7}$ & $-1.286$ & $[-0.167, -0.065]$ \\
\midrule
Female & RBE & MedBOE - IF-Net & $0.649$ & $0.734$ & $0.396$ & $[-1.396, 2.694]$ \\*
 &  & ShapeFormer - IF-Net & $11.474$ & $<1.00 \times 10^{-12}$ & $2.103$ & $[9.429, 13.519]$ \\*
 &  & ShapeFormer - MedBOE & $10.825$ & $<1.00 \times 10^{-12}$ & $1.965$ & $[8.780, 12.870]$ \\
 & RVE & MedBOE - IF-Net & $-1.585$ & $0.452$ & $-0.959$ & $[-4.696, 1.527]$ \\*
 &  & ShapeFormer - IF-Net & $13.609$ & $<1.00 \times 10^{-12}$ & $1.599$ & $[10.497, 16.721]$ \\*
 &  & ShapeFormer - MedBOE & $15.194$ & $<1.00 \times 10^{-12}$ & $1.813$ & $[12.082, 18.305]$ \\
 & IoU & MedBOE - IF-Net & $-0.009$ & $0.502$ & $-0.488$ & $[-0.027, 0.010]$ \\*
 &  & ShapeFormer - IF-Net & $-0.325$ & $<1.00 \times 10^{-12}$ & $-6.601$ & $[-0.343, -0.307]$ \\*
 &  & ShapeFormer - MedBOE & $-0.316$ & $<1.00 \times 10^{-12}$ & $-6.574$ & $[-0.335, -0.298]$ \\
 & CD & MedBOE - IF-Net & $0.011$ & $0.395$ & $0.832$ & $[-0.007, 0.028]$ \\*
 &  & PCN - IF-Net & $0.126$ & $<1.00 \times 10^{-12}$ & $2.617$ & $[0.108, 0.143]$ \\*
 &  & ShapeFormer - IF-Net & $0.050$ & $1.75 \times 10^{-11}$ & $4.637$ & $[0.033, 0.068]$ \\*
 &  & PCN - MedBOE & $0.115$ & $<1.00 \times 10^{-12}$ & $2.319$ & $[0.098, 0.133]$ \\*
 &  & ShapeFormer - MedBOE & $0.040$ & $1.22 \times 10^{-7}$ & $2.402$ & $[0.022, 0.057]$ \\*
 &  & ShapeFormer - PCN & $-0.076$ & $<1.00 \times 10^{-12}$ & $-1.537$ & $[-0.093, -0.058]$ \\
 & HD & MedBOE - IF-Net & $0.019$ & $0.499$ & $0.277$ & $[-0.016, 0.055]$ \\*
 &  & PCN - IF-Net & $0.184$ & $<1.00 \times 10^{-12}$ & $3.617$ & $[0.148, 0.220]$ \\*
 &  & ShapeFormer - IF-Net & $0.141$ & $<1.00 \times 10^{-12}$ & $2.286$ & $[0.106, 0.177]$ \\*
 &  & PCN - MedBOE & $0.165$ & $<1.00 \times 10^{-12}$ & $1.993$ & $[0.129, 0.200]$ \\*
 &  & ShapeFormer - MedBOE & $0.122$ & $<1.00 \times 10^{-12}$ & $1.359$ & $[0.086, 0.158]$ \\*
 &  & ShapeFormer - PCN & $-0.043$ & $0.012$ & $-0.560$ & $[-0.078, -0.007]$ \\
\end{longtable}
\endgroup

Male-female comparison statistics (Table~\ref{tab:appendix6}) are reported from Welch's independent two-sample $t$-tests, with mean differences, Cohen's $d$, and 95\% CIs reported as female minus male.

\begin{table}[htbp]
\centering
\caption{Male-female comparison statistics from Welch's independent two-sample $t$-tests for the AutoPET dataset.}
\label{tab:appendix6}
\footnotesize
\setlength{\tabcolsep}{3pt}
\renewcommand{\arraystretch}{1.15}
\begin{tabular*}{\textwidth}{@{\extracolsep{\fill}}llcccc@{}}
\toprule
\textbf{Method} & \textbf{Metric} & \textbf{\shortstack{Mean\\difference}} & \textbf{$p$-value} & \textbf{Cohen's $d$} & \textbf{95\% CI} \\
\midrule
MedBOE & RBE & $0.590$ & $0.109$ & $0.340$ & $[-0.134, 1.315]$ \\
 & RVE & $0.032$ & $0.849$ & $0.042$ & $[-0.300, 0.363]$ \\
 & IoU & $0.001$ & $0.785$ & $0.067$ & $[-0.008, 0.011]$ \\
 & CD & $-0.001$ & $0.807$ & $-0.051$ & $[-0.008, 0.006]$ \\
 & HD & $-0.006$ & $0.783$ & $-0.060$ & $[-0.047, 0.036]$ \\
\midrule
IF-Net & RBE & $-1.876$ & $0.088$ & $-0.468$ & $[-4.041, 0.290]$ \\
 & RVE & $-1.304$ & $0.377$ & $-0.237$ & $[-4.256, 1.649]$ \\
 & IoU & $0.018$ & $0.202$ & $0.345$ & $[-0.010, 0.047]$ \\
 & CD & $-0.002$ & $0.073$ & $-0.452$ & $[-0.004, 0.000]$ \\
 & HD & $-0.017$ & $0.007$ & $-0.622$ & $[-0.028, -0.005]$ \\
\midrule
ShapeFormer & RBE & $-1.876$ & $0.293$ & $-0.235$ & $[-5.407, 1.655]$ \\
 & RVE & $-3.234$ & $0.221$ & $-0.269$ & $[-8.452, 1.984]$ \\
 & IoU & $0.024$ & $0.122$ & $0.346$ & $[-0.007, 0.055]$ \\
 & CD & $0.003$ & $0.335$ & $0.213$ & $[-0.003, 0.010]$ \\
 & HD & $0.039$ & $0.023$ & $0.487$ & $[0.006, 0.073]$ \\
\midrule
PCN & CD & $-0.020$ & $0.192$ & $-0.287$ & $[-0.050, 0.010]$ \\
 & HD & $-0.034$ & $0.088$ & $-0.411$ & $[-0.073, 0.005]$ \\
\botrule
\end{tabular*}
\end{table}

\subsection{Zoomed Comparison of HIT Estimated Complete Patient Body Models}\label{subsec:appendix_hit_zoom}

\begin{figure}[htbp]
\centering
\includegraphics[width=\linewidth]{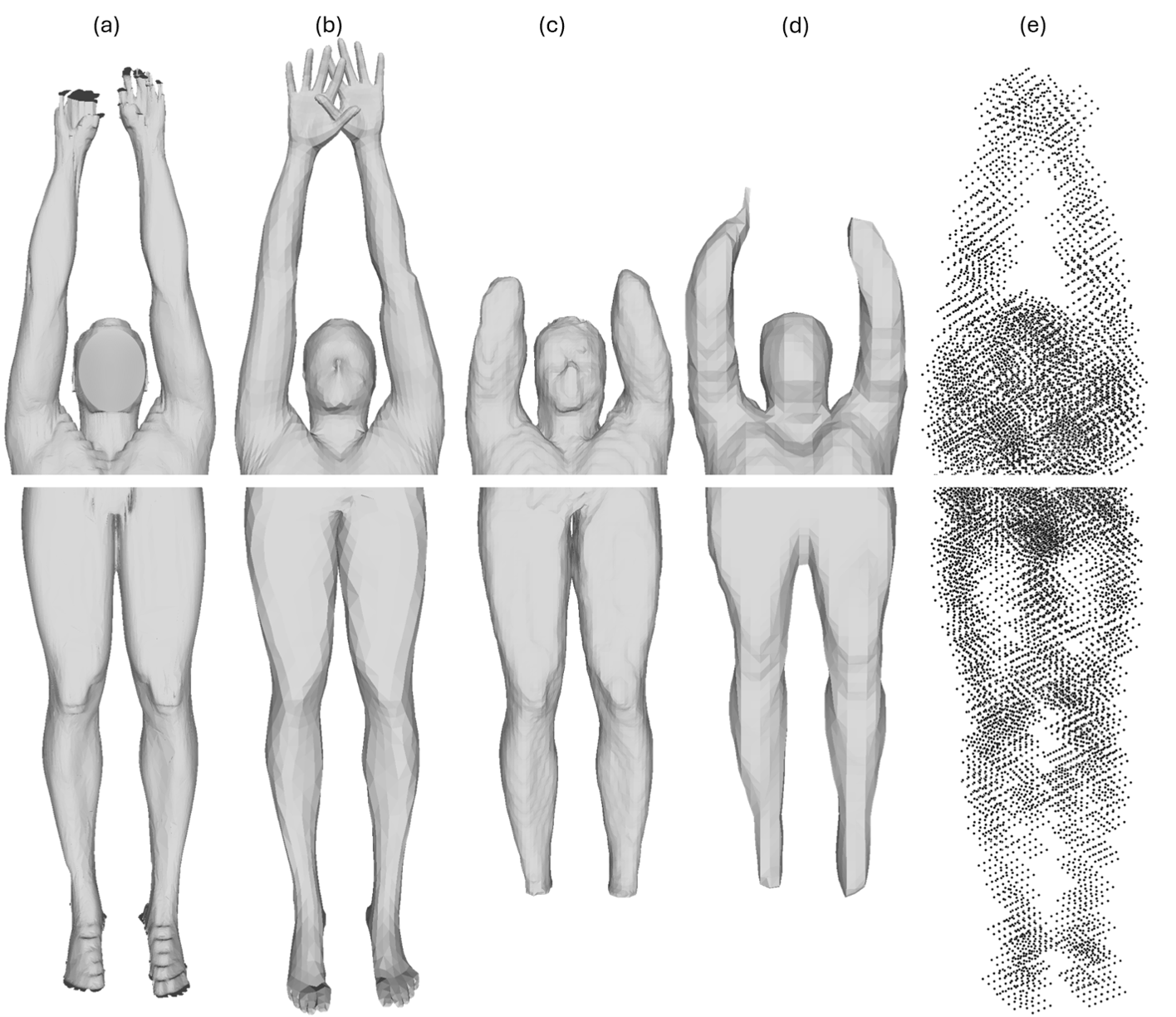}
\caption{Zoomed views of the HIT male complete patient body model estimates. The top row shows the hand and forearm regions, and the bottom row shows the leg region. (a) GT body model, (b) MedBOE, (c) IF-Net, (d) ShapeFormer, and (e) PCN}
\label{fig10}
\end{figure}

\begin{figure}[htbp]
\centering
\includegraphics[width=\linewidth]{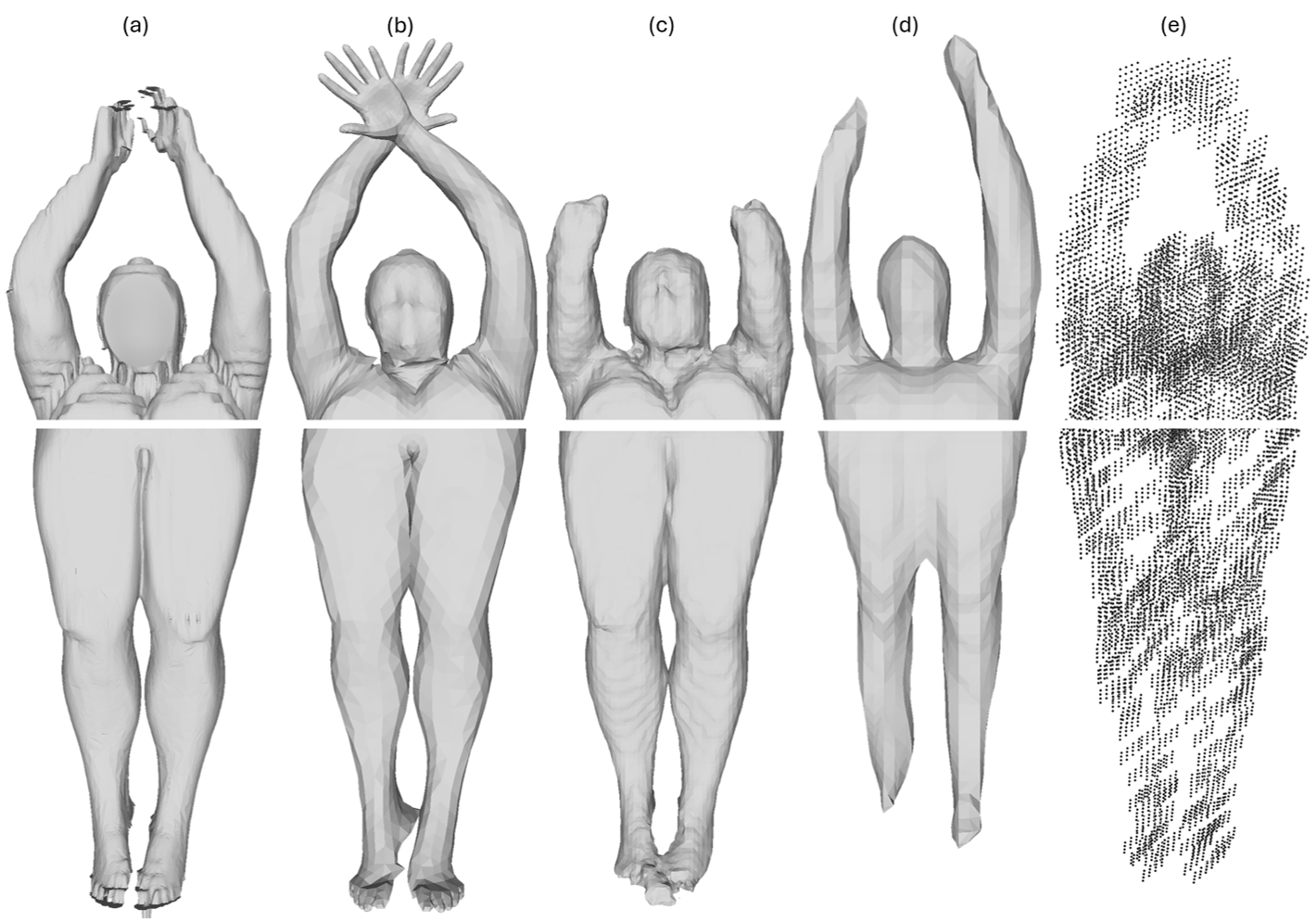}
\caption{Zoomed views of the HIT female complete patient body model estimates. The top row shows the hand and forearm regions, and the bottom row shows the leg region. (a) GT body model, (b) MedBOE, (c) IF-Net, (d) ShapeFormer, and (e) PCN. PCN is shown as point-cloud output}
\label{fig11}
\end{figure}

Fig.~\ref{fig10} and~\ref{fig11} provide zoomed views of the hand, forearm, and leg regions from the HIT male and female examples, respectively. These regions were included because HIT provides near-complete patient body coverage, enabling visual comparison of estimated missing body regions after simulated trimming. The zoomed views show local differences among MedBOE and the baseline methods in regions where anatomical cues were limited. PCN is shown as point-cloud output.

\subsection{Representative Challenging Case from HIT}\label{subsec:appendix_hit_challenging}

\begin{figure}[htbp]
\centering
\includegraphics[width=0.9\linewidth]{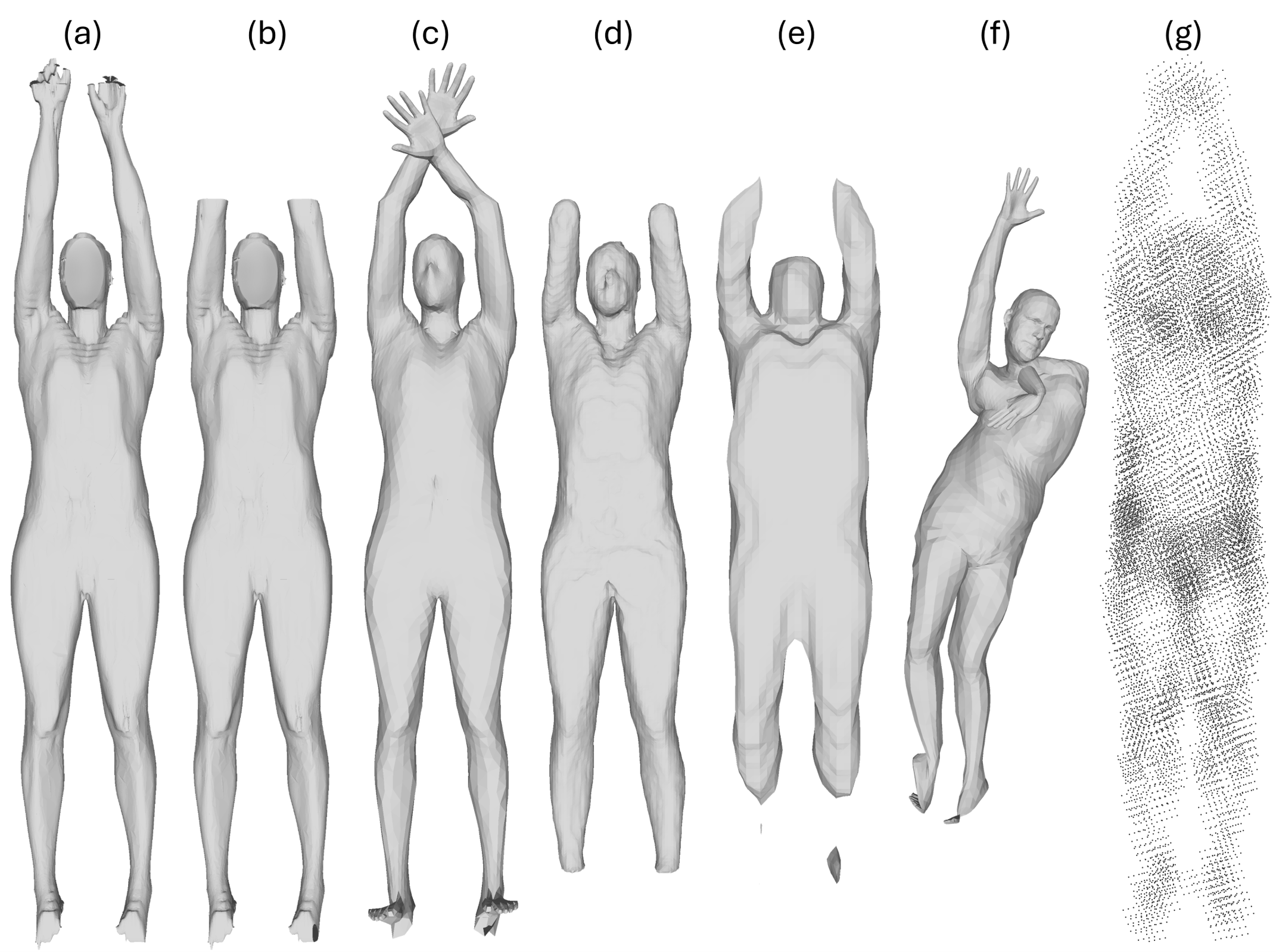}
\caption{Representative challenging case from the HIT dataset. (a) GT body model, (b) incomplete patient body model, (c) MedBOE, (d) IF-Net, (e) ShapeFormer, (f) ArtEq, and (g) PCN}
\label{fig12}
\end{figure}

Fig.~\ref{fig12} shows a representative challenging case from the HIT dataset. In this case, the hand and forearm regions were missing after simulated trimming, which limited the local anatomical cues available for pose and surface estimation. MedBOE estimated complete upper limbs, but also produced local artifacts. Around the elbow regions, artifacts appeared where the missing-region estimates joined the incomplete patient body model, reflecting uncertainty in forearm pose estimation. The incomplete foot shape also led to heel deformation artifacts. IF-Net and ShapeFormer both produced incomplete upper-limb estimates, although ShapeFormer predicted more of the missing limb regions than IF-Net; both methods also produced lower-limb estimates that were more incomplete than the input incomplete body model. ArtEq produced collapsed body-model estimates, while PCN generated point-cloud predictions that were visually closer to the GT in some regions but did not provide a continuous surface estimate.

\subsection{Statistical Analysis of HIT Estimated Complete Patient Body Models}\label{subsec:appendix_hit_statistics}

Table~\ref{tab:appendix7} and~\ref{tab:appendix8} provide detailed statistical results for the HIT complete patient body model estimates. Pairwise method-comparison statistics (Table~\ref{tab:appendix7}) are reported from Tukey HSD tests, including $p$-values, Cohen's $d$, and 95\% CIs.

\begin{sidewaystable}[p]
\centering
\caption{Pairwise method-comparison statistics from Tukey HSD tests for the HIT dataset.}
\label{tab:appendix7}
\footnotesize
\setlength{\tabcolsep}{1.5pt}
\renewcommand{\arraystretch}{0.95}
\begin{tabular}{@{}llrrrr@{\hspace{8pt}}rrrr@{}}
\toprule
 & & \multicolumn{4}{c}{\textbf{Male}} & \multicolumn{4}{c}{\textbf{Female}} \\
\cmidrule(lr){3-6}\cmidrule(lr){7-10}
\textbf{Metric} & \textbf{Comparison} & \textbf{\shortstack{Mean\\difference}} & \textbf{$p$-value} & \textbf{Cohen's $d$} & \textbf{95\% CI} & \textbf{\shortstack{Mean\\difference}} & \textbf{$p$-value} & \textbf{Cohen's $d$} & \textbf{95\% CI} \\
\midrule
RBE & IF-Net - ArtEq & $4.672$ & $0.009$ & $1.156$ & $[0.902, 8.441]$ & $0.036$ & $1.000$ & $0.008$ & $[-3.039, 3.110]$ \\
 & MedBOE - ArtEq & $-6.374$ & $2.16 \times 10^{-4}$ & $-1.620$ & $[-10.143, -2.604]$ & $-6.935$ & $3.29 \times 10^{-7}$ & $-1.520$ & $[-10.010, -3.860]$ \\
 & ShapeFormer - ArtEq & $4.494$ & $0.013$ & $0.880$ & $[0.724, 8.264]$ & $4.619$ & $9.16 \times 10^{-4}$ & $0.818$ & $[1.544, 7.694]$ \\
 & MedBOE - IF-Net & $-11.045$ & $1.16 \times 10^{-9}$ & $-5.296$ & $[-14.815, -7.275]$ & $-6.971$ & $2.87 \times 10^{-7}$ & $-4.235$ & $[-10.045, -3.896]$ \\
 & ShapeFormer - IF-Net & $-0.177$ & $0.999$ & $-0.046$ & $[-3.947, 3.592]$ & $4.584$ & $0.001$ & $1.236$ & $[1.509, 7.659]$ \\
 & ShapeFormer - MedBOE & $10.868$ & $1.85 \times 10^{-9}$ & $2.897$ & $[7.098, 14.637]$ & $11.554$ & $<1.00 \times 10^{-12}$ & $3.096$ & $[8.480, 14.629]$ \\
RVE & IF-Net - ArtEq & $-3.364$ & $0.061$ & $-0.800$ & $[-6.834, 0.105]$ & $-12.191$ & $1.39 \times 10^{-9}$ & $-1.844$ & $[-16.689, -7.692]$ \\
 & MedBOE - ArtEq & $-7.741$ & $1.27 \times 10^{-6}$ & $-1.878$ & $[-11.210, -4.272]$ & $-14.307$ & $3.67 \times 10^{-12}$ & $-2.175$ & $[-18.805, -9.808]$ \\
 & ShapeFormer - ArtEq & $2.326$ & $0.296$ & $0.466$ & $[-1.143, 5.795]$ & $-4.256$ & $0.071$ & $-0.497$ & $[-8.754, 0.243]$ \\
 & MedBOE - IF-Net & $-4.377$ & $0.008$ & $-4.716$ & $[-7.846, -0.907]$ & $-2.116$ & $0.610$ & $-2.575$ & $[-6.615, 2.383]$ \\
 & ShapeFormer - IF-Net & $5.690$ & $3.40 \times 10^{-4}$ & $1.923$ & $[2.221, 9.160]$ & $7.935$ & $7.20 \times 10^{-5}$ & $1.431$ & $[3.436, 12.434]$ \\
 & ShapeFormer - MedBOE & $10.067$ & $1.53 \times 10^{-9}$ & $3.547$ & $[6.598, 13.536]$ & $10.051$ & $4.20 \times 10^{-7}$ & $1.825$ & $[5.552, 14.550]$ \\
IoU & IF-Net - ArtEq & $0.543$ & $<1.00 \times 10^{-12}$ & $15.128$ & $[0.504, 0.581]$ & $0.499$ & $<1.00 \times 10^{-12}$ & $10.892$ & $[0.466, 0.532]$ \\
 & MedBOE - ArtEq & $0.570$ & $<1.00 \times 10^{-12}$ & $16.068$ & $[0.531, 0.608]$ & $0.514$ & $<1.00 \times 10^{-12}$ & $11.265$ & $[0.481, 0.547]$ \\
 & ShapeFormer - ArtEq & $0.329$ & $<1.00 \times 10^{-12}$ & $5.909$ & $[0.291, 0.367]$ & $0.283$ & $<1.00 \times 10^{-12}$ & $4.578$ & $[0.250, 0.316]$ \\
 & MedBOE - IF-Net & $0.027$ & $0.251$ & $3.410$ & $[-0.011, 0.066]$ & $0.015$ & $0.618$ & $1.682$ & $[-0.017, 0.048]$ \\
 & ShapeFormer - IF-Net & $-0.214$ & $<1.00 \times 10^{-12}$ & $-4.893$ & $[-0.252, -0.175]$ & $-0.216$ & $<1.00 \times 10^{-12}$ & $-5.063$ & $[-0.249, -0.183]$ \\
 & ShapeFormer - MedBOE & $-0.241$ & $<1.00 \times 10^{-12}$ & $-5.558$ & $[-0.279, -0.202]$ & $-0.231$ & $<1.00 \times 10^{-12}$ & $-5.442$ & $[-0.264, -0.198]$ \\
CD & IF-Net - ArtEq & $-0.113$ & $<1.00 \times 10^{-12}$ & $-3.103$ & $[-0.138, -0.087]$ & $-0.159$ & $<1.00 \times 10^{-12}$ & $-5.314$ & $[-0.175, -0.143]$ \\
 & MedBOE - ArtEq & $-0.117$ & $<1.00 \times 10^{-12}$ & $-3.245$ & $[-0.143, -0.091]$ & $-0.154$ & $<1.00 \times 10^{-12}$ & $-5.035$ & $[-0.170, -0.138]$ \\
 & PCN - ArtEq & $-0.107$ & $<1.00 \times 10^{-12}$ & $-2.858$ & $[-0.132, -0.081]$ & $-0.152$ & $<1.00 \times 10^{-12}$ & $-5.024$ & $[-0.168, -0.136]$ \\
 & ShapeFormer - ArtEq & $-0.103$ & $<1.00 \times 10^{-12}$ & $-2.828$ & $[-0.129, -0.078]$ & $-0.144$ & $<1.00 \times 10^{-12}$ & $-4.776$ & $[-0.160, -0.128]$ \\
 & MedBOE - IF-Net & $-0.005$ & $0.988$ & $-0.435$ & $[-0.030, 0.021]$ & $0.005$ & $0.933$ & $0.489$ & $[-0.011, 0.021]$ \\
 & PCN - IF-Net & $0.006$ & $0.968$ & $0.414$ & $[-0.020, 0.032]$ & $0.007$ & $0.734$ & $0.911$ & $[-0.009, 0.023]$ \\
 & ShapeFormer - IF-Net & $0.009$ & $0.854$ & $0.772$ & $[-0.017, 0.035]$ & $0.015$ & $0.085$ & $1.924$ & $[-0.001, 0.031]$ \\
 & PCN - MedBOE & $0.010$ & $0.789$ & $0.762$ & $[-0.015, 0.036]$ & $0.003$ & $0.992$ & $0.245$ & $[-0.013, 0.018]$ \\
 & ShapeFormer - MedBOE & $0.014$ & $0.570$ & $1.215$ & $[-0.012, 0.039]$ & $0.010$ & $0.399$ & $0.997$ & $[-0.006, 0.026]$ \\
 & ShapeFormer - PCN & $0.003$ & $0.996$ & $0.225$ & $[-0.022, 0.029]$ & $0.008$ & $0.675$ & $0.868$ & $[-0.008, 0.024]$ \\
HD & IF-Net - ArtEq & $-0.062$ & $0.309$ & $-0.628$ & $[-0.153, 0.028]$ & $-0.156$ & $<1.00 \times 10^{-12}$ & $-2.624$ & $[-0.203, -0.109]$ \\
 & MedBOE - ArtEq & $-0.234$ & $4.26 \times 10^{-9}$ & $-2.092$ & $[-0.324, -0.144]$ & $-0.318$ & $<1.00 \times 10^{-12}$ & $-4.783$ & $[-0.365, -0.271]$ \\
 & PCN - ArtEq & $-0.261$ & $1.24 \times 10^{-10}$ & $-2.434$ & $[-0.351, -0.171]$ & $-0.367$ & $<1.00 \times 10^{-12}$ & $-6.743$ & $[-0.414, -0.320]$ \\
 & ShapeFormer - ArtEq & $-0.204$ & $2.02 \times 10^{-7}$ & $-1.858$ & $[-0.295, -0.114]$ & $-0.280$ & $<1.00 \times 10^{-12}$ & $-3.707$ & $[-0.327, -0.233]$ \\
 & MedBOE - IF-Net & $-0.172$ & $1.15 \times 10^{-5}$ & $-2.488$ & $[-0.262, -0.081]$ & $-0.162$ & $<1.00 \times 10^{-12}$ & $-2.997$ & $[-0.209, -0.114]$ \\
 & PCN - IF-Net & $-0.199$ & $4.07 \times 10^{-7}$ & $-3.251$ & $[-0.289, -0.108]$ & $-0.211$ & $<1.00 \times 10^{-12}$ & $-5.524$ & $[-0.258, -0.164]$ \\
 & ShapeFormer - IF-Net & $-0.142$ & $3.63 \times 10^{-4}$ & $-2.160$ & $[-0.232, -0.051]$ & $-0.124$ & $3.70 \times 10^{-10}$ & $-1.913$ & $[-0.171, -0.077]$ \\
 & PCN - MedBOE & $-0.027$ & $0.918$ & $-0.338$ & $[-0.117, 0.063]$ & $-0.049$ & $0.035$ & $-1.022$ & $[-0.097, -0.002]$ \\
 & ShapeFormer - MedBOE & $0.030$ & $0.886$ & $0.358$ & $[-0.060, 0.120]$ & $0.038$ & $0.184$ & $0.527$ & $[-0.010, 0.085]$ \\
 & ShapeFormer - PCN & $0.057$ & $0.403$ & $0.737$ & $[-0.033, 0.147]$ & $0.087$ & $1.22 \times 10^{-5}$ & $1.443$ & $[0.040, 0.134]$ \\
\botrule
\end{tabular}
\end{sidewaystable}

Male-female comparison statistics (Table~\ref{tab:appendix8}) are reported from Welch's independent two-sample $t$-tests, with mean differences, Cohen's $d$, and 95\% CIs reported as female minus male.

\begin{table}[!htbp]
\centering
\caption{Male-female comparison statistics from Welch's independent two-sample $t$-tests for the HIT dataset.}
\label{tab:appendix8}
\footnotesize
\setlength{\tabcolsep}{3pt}
\renewcommand{\arraystretch}{1.15}
\begin{tabular*}{\textwidth}{@{\extracolsep{\fill}}llcccc@{}}
\toprule
\textbf{Method} & \textbf{Metric} & \textbf{\shortstack{Mean\\difference}} & \textbf{$p$-value} & \textbf{Cohen's $d$} & \textbf{95\% CI} \\
\midrule
MedBOE & RBE & $0.294$ & $0.622$ & $0.167$ & $[-0.915, 1.504]$ \\
 & RVE & $-0.019$ & $0.896$ & $-0.041$ & $[-0.310, 0.272]$ \\
 & IoU & $0.004$ & $0.123$ & $0.476$ & $[-0.001, 0.008]$ \\
 & CD & $-0.006$ & $0.112$ & $-0.510$ & $[-0.012, 0.001]$ \\
 & HD & $-0.048$ & $0.074$ & $-0.666$ & $[-0.100, 0.005]$ \\
\midrule
IF-Net & RBE & $-3.780$ & $1.08 \times 10^{-5}$ & $-2.013$ & $[-5.169, -2.391]$ \\
 & RVE & $-2.279$ & $3.18 \times 10^{-6}$ & $-2.020$ & $[-3.071, -1.487]$ \\
 & IoU & $0.016$ & $3.08 \times 10^{-5}$ & $1.590$ & $[0.009, 0.022]$ \\
 & CD & $-0.015$ & $1.53 \times 10^{-4}$ & $-1.732$ & $[-0.021, -0.008]$ \\
 & HD & $-0.058$ & $5.46 \times 10^{-4}$ & $-1.272$ & $[-0.088, -0.027]$ \\
\midrule
ShapeFormer & RBE & $0.981$ & $0.551$ & $0.197$ & $[-2.340, 4.302]$ \\
 & RVE & $-0.035$ & $0.985$ & $-0.005$ & $[-3.813, 3.743]$ \\
 & IoU & $0.013$ & $0.510$ & $0.219$ & $[-0.027, 0.054]$ \\
 & CD & $-0.009$ & $0.022$ & $-0.881$ & $[-0.017, -0.001]$ \\
 & HD & $-0.040$ & $0.139$ & $-0.498$ & $[-0.094, 0.014]$ \\
\midrule
PCN & CD & $-0.013$ & $0.010$ & $-1.087$ & $[-0.023, -0.004]$ \\
 & HD & $-0.070$ & $0.003$ & $-1.390$ & $[-0.112, -0.028]$ \\
\midrule
ArtEq & RBE & $0.856$ & $0.645$ & $0.146$ & $[-2.882, 4.594]$ \\
 & RVE & $6.547$ & $0.009$ & $0.800$ & $[1.710, 11.383]$ \\
 & IoU & $0.059$ & $0.002$ & $1.000$ & $[0.022, 0.096]$ \\
 & CD & $0.032$ & $0.051$ & $0.701$ & $[-0.000, 0.063]$ \\
 & HD & $0.036$ & $0.345$ & $0.366$ & $[-0.042, 0.114]$ \\
\botrule
\end{tabular*}
\end{table}

\subsection{Statistical Analysis of Dataset-Wise Comparisons}\label{subsec:appendix_dataset_statistics}

Table~\ref{tab:appendix9} provides dataset-wise comparison statistics between AutoPET and HIT for each method and metric. The table reports $p$-values, Cohen's $d$, and 95\% CIs. Mean differences are reported as HIT minus AutoPET.

\begin{table}[!htbp]
\centering
\caption{Dataset-wise comparison statistics between AutoPET and HIT.}
\label{tab:appendix9}
\footnotesize
\setlength{\tabcolsep}{3pt}
\renewcommand{\arraystretch}{1.15}
\begin{tabular*}{\textwidth}{@{\extracolsep{\fill}}llcccc@{}}
\toprule
\textbf{Method} & \textbf{Metric} & \textbf{\shortstack{Mean\\difference}} & \textbf{$p$-value} & \textbf{Cohen's $d$} & \textbf{95\% CI} \\
\midrule
MedBOE & RBE & $0.202$ & $0.543$ & $0.116$ & $[-0.458, 0.862]$ \\
 & RVE & $-0.063$ & $0.557$ & $-0.093$ & $[-0.274, 0.148]$ \\
 & IoU & $0.021$ & $<1.00 \times 10^{-12}$ & $1.218$ & $[0.016, 0.025]$ \\
 & CD & $0.011$ & $5.30 \times 10^{-5}$ & $0.680$ & $[0.006, 0.015]$ \\
 & HD & $0.067$ & $3.45 \times 10^{-5}$ & $0.748$ & $[0.037, 0.098]$ \\
\midrule
IF-Net & RBE & $8.401$ & $<1.00 \times 10^{-12}$ & $2.266$ & $[7.220, 9.583]$ \\
 & RVE & $0.803$ & $0.204$ & $0.171$ & $[-0.443, 2.048]$ \\
 & IoU & $-0.001$ & $0.848$ & $-0.025$ & $[-0.013, 0.011]$ \\
 & CD & $0.020$ & $<1.00 \times 10^{-12}$ & $2.689$ & $[0.016, 0.023]$ \\
 & HD & $0.248$ & $<1.00 \times 10^{-12}$ & $6.661$ & $[0.230, 0.266]$ \\
\midrule
ShapeFormer & RBE & $-0.274$ & $0.811$ & $-0.038$ & $[-2.543, 1.994]$ \\
 & RVE & $-6.455$ & $1.31 \times 10^{-4}$ & $-0.603$ & $[-9.690, -3.221]$ \\
 & IoU & $0.111$ & $<1.00 \times 10^{-12}$ & $1.653$ & $[0.087, 0.135]$ \\
 & CD & $-0.016$ & $1.39 \times 10^{-9}$ & $-1.120$ & $[-0.021, -0.011]$ \\
 & HD & $-0.003$ & $0.858$ & $-0.034$ & $[-0.034, 0.028]$ \\
\midrule
PCN & CD & $-0.106$ & $<1.00 \times 10^{-12}$ & $-1.819$ & $[-0.122, -0.091]$ \\
 & HD & $-0.149$ & $<1.00 \times 10^{-12}$ & $-1.923$ & $[-0.175, -0.123]$ \\
\botrule
\end{tabular*}
\end{table}

\subsection{Statistical Analysis of Ablation Study}\label{subsec:appendix_ablation_statistics}

Table~\ref{tab:appendix10} provides ablation study statistics between with and without the missing region estimation and surface deformation module for each dataset and gender. The table reports $p$-values, Cohen's $d$, and 95\% CIs. Mean differences are reported as with the module minus without the module.

\begin{table}[!htbp]
\centering
\caption{Ablation study statistics between with and without the missing region estimation and surface deformation module.}
\label{tab:appendix10}
\footnotesize
\setlength{\tabcolsep}{3pt}
\renewcommand{\arraystretch}{1.15}
\begin{tabular*}{\textwidth}{@{\extracolsep{\fill}}lllcccc@{}}
\toprule
\textbf{Dataset} & \textbf{Gender} & \textbf{Metric} & \textbf{\shortstack{Mean\\difference}} & \textbf{$p$-value} & \textbf{Cohen's $d$} & \textbf{95\% CI} \\
\midrule
AutoPET & Male & RBE & $-2.359$ & $9.85 \times 10^{-9}$ & $-1.271$ & $[-2.996, -1.721]$ \\
 &  & RVE & $-5.036$ & $<1.00 \times 10^{-12}$ & $-2.112$ & $[-5.855, -4.217]$ \\
 &  & IoU & $0.133$ & $<1.00 \times 10^{-12}$ & $6.648$ & $[0.126, 0.140]$ \\
 &  & CD & $-0.009$ & $<1.00 \times 10^{-12}$ & $-3.779$ & $[-0.010, -0.009]$ \\
 &  & HD & $-0.009$ & $5.31 \times 10^{-6}$ & $-0.912$ & $[-0.012, -0.005]$ \\
 & Female & RBE & $-2.168$ & $4.97 \times 10^{-9}$ & $-0.926$ & $[-2.795, -1.541]$ \\
 &  & RVE & $-6.397$ & $<1.00 \times 10^{-12}$ & $-2.032$ & $[-7.240, -5.554]$ \\
 &  & IoU & $0.144$ & $<1.00 \times 10^{-12}$ & $6.725$ & $[0.139, 0.150]$ \\
 &  & CD & $-0.010$ & $<1.00 \times 10^{-12}$ & $-2.740$ & $[-0.011, -0.009]$ \\
 &  & HD & $-0.009$ & $3.38 \times 10^{-9}$ & $-0.939$ & $[-0.012, -0.006]$ \\
\midrule
HIT & Male & RBE & $-5.121$ & $2.99 \times 10^{-8}$ & $-2.828$ & $[-6.123, -4.118]$ \\
 &  & RVE & $-10.500$ & $3.68 \times 10^{-9}$ & $-3.331$ & $[-12.246, -8.754]$ \\
 &  & IoU & $0.178$ & $<1.00 \times 10^{-12}$ & $9.257$ & $[0.167, 0.188]$ \\
 &  & CD & $-0.013$ & $1.63 \times 10^{-11}$ & $-5.007$ & $[-0.014, -0.011]$ \\
 &  & HD & $0.000$ & $0.932$ & $0.022$ & $[-0.005, 0.005]$ \\
 & Female & RBE & $-4.278$ & $2.40 \times 10^{-11}$ & $-2.325$ & $[-5.037, -3.519]$ \\
 &  & RVE & $-11.004$ & $<1.00 \times 10^{-12}$ & $-5.121$ & $[-11.891, -10.117]$ \\
 &  & IoU & $0.191$ & $<1.00 \times 10^{-12}$ & $5.920$ & $[0.177, 0.204]$ \\
 &  & CD & $-0.012$ & $<1.00 \times 10^{-12}$ & $-3.019$ & $[-0.014, -0.011]$ \\
 &  & HD & $-0.000$ & $0.574$ & $-0.114$ & $[-0.000, 0.000]$ \\
\botrule
\end{tabular*}
\end{table}

\section{Results of Missing Region Estimation}\label{sec:appendix_missing_regions}

Missing body region estimation results on simulated incomplete patient body models from the AutoPET dataset are reported in Table~\ref{tab:appendix11}. Due to failure in generating valid outputs, four IF-Net cases (two male, two female) and 12 ShapeFormer cases (six male, six female) were excluded; all other results were retained. MedBOE achieved the best performance across all metrics for both male and female models. ShapeFormer ranked second in volumetric metrics, while IF-Net ranked second in geometric metrics.

\begin{table}[!htbp]
\centering
\caption{Results of missing region estimation on the AutoPET dataset.}
\label{tab:appendix11}
\footnotesize
\setlength{\tabcolsep}{2pt}
\renewcommand{\arraystretch}{2.0}
\begin{tabular*}{\textwidth}{@{\extracolsep{\fill}}llccccc@{}}
\toprule
 & \textbf{Methods} & \textbf{RBE $\downarrow$} & \textbf{RVE $\downarrow$} & \textbf{IoU $\uparrow$} & \textbf{CD $\downarrow$} & \textbf{HD $\downarrow$} \\
\midrule
\multirow{4}{*}{\rotatebox[origin=c]{90}{\textbf{Male (35)}}} & \textbf{MedBOE} & \shortstack[c]{\textbf{\boldmath $30.28 \pm 42.69$ *}\\\textbf{(0.32--225.96)}} & \shortstack[c]{\textbf{\boldmath $46.48 \pm 78.26$ *}\\\textbf{(3.06--386.43)}} & \shortstack[c]{\textbf{\boldmath $0.44 \pm 0.16$ *}\\\textbf{(0.09--0.70)}} & \shortstack[c]{\textbf{\boldmath $0.06 \pm 0.06$}\\\textbf{(0.01--0.24)}} & \shortstack[c]{\textbf{\boldmath $0.17 \pm 0.19$}\\\textbf{(0.02--0.76)}} \\
 & \textbf{IF-Net}\textsuperscript{\textasciicircum} & \shortstack[c]{$91.38 \pm 7.46$\\(68.81--99.98)} & \shortstack[c]{$97.67 \pm 3.52$\\(82.99--100.00)} & \shortstack[c]{$0.02 \pm 0.03$\\(0.00--0.17)} & \shortstack[c]{\underline{$0.16 \pm 0.15$}\\\underline{(0.04--0.69)}} & \shortstack[c]{\underline{$0.33 \pm 0.24$}\\\underline{(0.06--0.90)}} \\
 & \shortstack[l]{\textbf{Shape}\\\textbf{Former}}\textsuperscript{\textasciicircum} & \shortstack[c]{\underline{$77.74 \pm 71.95$}\\\underline{(10.34--412.23)}} & \shortstack[c]{\underline{$94.45 \pm 116.24$}\\\underline{(5.26--676.19)}} & \shortstack[c]{\underline{$0.05 \pm 0.07$}\\\underline{(0.00--0.23)}} & \shortstack[c]{$0.28 \pm 0.22$\\(0.06--0.89)} & \shortstack[c]{$0.57 \pm 0.34$\\(0.09--1.33)} \\
 & \textbf{PCN} & - & - & - & \shortstack[c]{$0.62 \pm 0.32$\\(0.25--1.54)} & \shortstack[c]{$1.18 \pm 0.43$\\(0.45--1.96)} \\
\midrule
\multirow{4}{*}{\rotatebox[origin=c]{90}{\textbf{Female (56)}}} & \textbf{MedBOE} & \shortstack[c]{\textbf{\boldmath $39.29 \pm 57.76$ *}\\\textbf{(0.28--322.29)}} & \shortstack[c]{\textbf{\boldmath $56.36 \pm 72.07$ *}\\\textbf{(0.57--363.75)}} & \shortstack[c]{\textbf{\boldmath $0.39 \pm 0.16$ *}\\\textbf{(0.00--0.66)}} & \shortstack[c]{\textbf{\boldmath $0.06 \pm 0.07$}\\\textbf{(0.01--0.35)}} & \shortstack[c]{\textbf{\boldmath $0.17 \pm 0.21$ *}\\\textbf{(0.02--0.84)}} \\
 & \textbf{IF-Net}\textsuperscript{\textasciicircum} & \shortstack[c]{$95.06 \pm 10.92$\\(21.09--99.93)} & \shortstack[c]{$95.06 \pm 10.92$\\(21.09--99.93)} & \shortstack[c]{$0.04 \pm 0.05$\\(0.00--0.25)} & \shortstack[c]{\underline{$0.12 \pm 0.12$}\\\underline{(0.01--0.85)}} & \shortstack[c]{\underline{$0.31 \pm 0.26$}\\\underline{(0.02--0.92)}} \\
 & \shortstack[l]{\textbf{Shape}\\\textbf{Former}}\textsuperscript{\textasciicircum} & \shortstack[c]{\underline{$77.42 \pm 41.82$}\\\underline{(3.18--264.06)}} & \shortstack[c]{\underline{$92.56 \pm 61.10$}\\\underline{(1.15--432.19)}} & \shortstack[c]{\underline{$0.05 \pm 0.07$}\\\underline{(0.00--0.34)}} & \shortstack[c]{$0.49 \pm 0.48$\\(0.09--2.45)} & \shortstack[c]{$0.57 \pm 0.24$\\(0.22--1.37)} \\
 & \textbf{PCN} & - & - & - & \shortstack[c]{$0.46 \pm 0.28$\\(0.06--1.29)} & \shortstack[c]{$0.90 \pm 0.38$\\(0.38--1.74)} \\
\botrule
\end{tabular*}
\footnotetext{An Asterisk (*) indicates statistical differences from the second-best results. A circumflex (\textasciicircum) indicates methods with estimation failure cases excluded from the results.}
\end{table}

Results on the HIT dataset are shown in Table~\ref{tab:appendix12}. Severe self-overlapping artifacts in ArtEq outputs, such as elbows embedded in the head and collapsed chest surfaces, prevented the generation of watertight surfaces required for computing RVE and IoU, leading to exclusion of those metrics. MedBOE achieved the best performance across all metrics for both genders, except CD in female models, where PCN performed best. ShapeFormer ranked second in volumetric metrics, and IF-Net in HD.

\begin{table}[!htbp]
\centering
\caption{Results of body region estimation on the HIT dataset.}
\label{tab:appendix12}
\footnotesize
\setlength{\tabcolsep}{2pt}
\renewcommand{\arraystretch}{2.0}
\begin{tabular*}{\textwidth}{@{\extracolsep{\fill}}llccccc@{}}
\toprule
 & \textbf{Methods} & \textbf{RBE $\downarrow$} & \textbf{RVE $\downarrow$} & \textbf{IoU $\uparrow$} & \textbf{CD $\downarrow$} & \textbf{HD $\downarrow$} \\
\midrule
\multirow{5}{*}{\rotatebox[origin=c]{90}{\textbf{Male (15)}}} & \textbf{MedBOE} & \shortstack[c]{\textbf{\boldmath $19.50 \pm 17.96$}\\\textbf{(1.82--68.95)}} & \shortstack[c]{\textbf{\boldmath $15.50 \pm 15.97$ *}\\\textbf{(1.73--51.69)}} & \shortstack[c]{\textbf{\boldmath $0.29 \pm 0.12$ *}\\\textbf{(0.10--0.51)}} & \shortstack[c]{\textbf{\boldmath $0.12 \pm 0.07$}\\\textbf{(0.03--0.28)}} & \shortstack[c]{\textbf{\boldmath $0.53 \pm 0.59$}\\\textbf{(0.08--1.84)}} \\
 & \textbf{IF-Net} & \shortstack[c]{$94.03 \pm 2.40$\\(88.87--97.98)} & \shortstack[c]{$97.78 \pm 1.39$\\(93.85--99.25)} & \shortstack[c]{$0.02 \pm 0.01$\\(0.01--0.05)} & \shortstack[c]{$0.20 \pm 0.06$\\(0.12--0.32)} & \shortstack[c]{\underline{$0.63 \pm 0.56$}\\\underline{(0.29--2.03)}} \\
 & \shortstack[l]{\textbf{Shape}\\\textbf{Former}} & \shortstack[c]{\underline{$56.05 \pm 20.33$}\\\underline{(10.61--82.13)}} & \shortstack[c]{\underline{$61.50 \pm 23.69$}\\\underline{(6.27--88.70)}} & \shortstack[c]{\underline{$0.16 \pm 0.09$}\\\underline{(0.05--0.32)}} & \shortstack[c]{\underline{$0.14 \pm 0.06$}\\\underline{(0.05--0.30)}} & \shortstack[c]{$0.70 \pm 0.55$\\(0.17--1.75)} \\
 & \textbf{PCN} & - & - & - & \shortstack[c]{$0.16 \pm 0.13$\\(0.03--0.44)} & \shortstack[c]{$0.82 \pm 0.37$\\(0.30--1.60)} \\
 & \textbf{ArtEq} & \shortstack[c]{$148.41 \pm 89.68$\\(11.85--293.73)} & - & - & \shortstack[c]{$0.60 \pm 0.25$\\(0.29--1.22)} & \shortstack[c]{$1.07 \pm 0.48$\\(0.46--1.82)} \\
\midrule
\multirow{5}{*}{\rotatebox[origin=c]{90}{\textbf{Female (25)}}} & \textbf{MedBOE} & \shortstack[c]{\textbf{\boldmath $26.50 \pm 17.05$}\\\textbf{(0.17--55.70)}} & \shortstack[c]{\textbf{\boldmath $24.81 \pm 19.84$ *}\\\textbf{(1.44--80.44)}} & \shortstack[c]{\textbf{\boldmath $0.26 \pm 0.09$}\\\textbf{(0.12--0.44)}} & \shortstack[c]{\underline{$0.10 \pm 0.05$}\\\underline{(0.04--0.24)}} & \shortstack[c]{\textbf{\boldmath $0.28 \pm 0.37$}\\\textbf{(0.06--1.71)}} \\
 & \textbf{IF-Net} & \shortstack[c]{$94.94 \pm 4.05$\\(86.44--99.71)} & \shortstack[c]{$97.95 \pm 2.02$\\(93.61--100.00)} & \shortstack[c]{$0.02 \pm 0.02$\\(0.00--0.06)} & \shortstack[c]{$0.19 \pm 0.04$\\(0.11--0.29)} & \shortstack[c]{\underline{$0.34 \pm 0.11$}\\\underline{(0.21--0.79)}} \\
 & \shortstack[l]{\textbf{Shape}\\\textbf{Former}} & \shortstack[c]{\underline{$55.94 \pm 28.76$}\\\underline{(0.50--97.20)}} & \shortstack[c]{\underline{$57.59 \pm 29.92$}\\\underline{(6.11--98.72)}} & \shortstack[c]{\underline{$0.20 \pm 0.15$}\\\underline{(0.01--0.49)}} & \shortstack[c]{$0.13 \pm 0.10$\\(0.03--0.43)} & \shortstack[c]{$0.47 \pm 0.45$\\(0.05--1.76)} \\
 & \textbf{PCN} & - & - & - & \shortstack[c]{\textbf{\boldmath $0.09 \pm 0.08$}\\\textbf{(0.03--0.30)}} & \shortstack[c]{$0.49 \pm 0.42$\\(0.06--1.37)} \\
 & \textbf{ArtEq} & \shortstack[c]{$231.44 \pm 130.53$\\(14.97--477.11)} & - & - & \shortstack[c]{$0.76 \pm 0.28$\\(0.34--1.56)} & \shortstack[c]{$1.24 \pm 0.35$\\(0.53--1.73)} \\
\botrule
\end{tabular*}
\footnotetext{An Asterisk (*) indicates statistical differences from the second-best results.}
\end{table}

\end{appendices}

\clearpage


\bibliography{sn-bibliography}


\begin{thebibliography}{50}
\ifx \bisbn   \undefined \def \bisbn  #1{ISBN #1}\fi
\ifx \binits  \undefined \def \binits#1{#1}\fi
\ifx \bauthor  \undefined \def \bauthor#1{#1}\fi
\ifx \batitle  \undefined \def \batitle#1{#1}\fi
\ifx \bjtitle  \undefined \def \bjtitle#1{#1}\fi
\ifx \bvolume  \undefined \def \bvolume#1{\textbf{#1}}\fi
\ifx \byear  \undefined \def \byear#1{#1}\fi
\ifx \bissue  \undefined \def \bissue#1{#1}\fi
\ifx \bfpage  \undefined \def \bfpage#1{#1}\fi
\ifx \blpage  \undefined \def \blpage #1{#1}\fi
\ifx \burl  \undefined \def \burl#1{\textsf{#1}}\fi
\ifx \doiurl  \undefined \def \doiurl#1{\url{https://doi.org/#1}}\fi
\ifx \betal  \undefined \def \betal{\textit{et al.}}\fi
\ifx \binstitute  \undefined \def \binstitute#1{#1}\fi
\ifx \binstitutionaled  \undefined \def \binstitutionaled#1{#1}\fi
\ifx \bctitle  \undefined \def \bctitle#1{#1}\fi
\ifx \beditor  \undefined \def \beditor#1{#1}\fi
\ifx \bpublisher  \undefined \def \bpublisher#1{#1}\fi
\ifx \bbtitle  \undefined \def \bbtitle#1{#1}\fi
\ifx \bedition  \undefined \def \bedition#1{#1}\fi
\ifx \bseriesno  \undefined \def \bseriesno#1{#1}\fi
\ifx \blocation  \undefined \def \blocation#1{#1}\fi
\ifx \bsertitle  \undefined \def \bsertitle#1{#1}\fi
\ifx \bsnm \undefined \def \bsnm#1{#1}\fi
\ifx \bsuffix \undefined \def \bsuffix#1{#1}\fi
\ifx \bparticle \undefined \def \bparticle#1{#1}\fi
\ifx \barticle \undefined \def \barticle#1{#1}\fi
\bibcommenthead
\ifx \bconfdate \undefined \def \bconfdate #1{#1}\fi
\ifx \botherref \undefined \def \botherref #1{#1}\fi
\ifx \url \undefined \def \url#1{\textsf{#1}}\fi
\ifx \bchapter \undefined \def \bchapter#1{#1}\fi
\ifx \bbook \undefined \def \bbook#1{#1}\fi
\ifx \bcomment \undefined \def \bcomment#1{#1}\fi
\ifx \oauthor \undefined \def \oauthor#1{#1}\fi
\ifx \citeauthoryear \undefined \def \citeauthoryear#1{#1}\fi
\ifx \endbibitem  \undefined \def \endbibitem {}\fi
\ifx \bconflocation  \undefined \def \bconflocation#1{#1}\fi
\ifx \arxivurl  \undefined \def \arxivurl#1{\textsf{#1}}\fi
\csname PreBibitemsHook\endcsname

\bibitem[\protect\citeauthoryear{Doi}{2007}]{bib1}
\begin{barticle}
\bauthor{\bsnm{Doi}, \binits{K.}}:
\batitle{Computer-aided diagnosis in medical imaging: Historical review, current status and future potential}.
\bjtitle{Computerized Medical Imaging and Graphics}
\bvolume{31}(\bissue{4--5}),
\bfpage{198}--\blpage{211}
(\byear{2007})
\doiurl{10.1016/j.compmedimag.2007.02.002}
\end{barticle}
\endbibitem

\bibitem[\protect\citeauthoryear{Badawi et~al.}{2019}]{badawi2019explorer}
\begin{barticle}
\bauthor{\bsnm{Badawi}, \binits{R.D.}},
\bauthor{\bsnm{Shi}, \binits{H.}},
\bauthor{\bsnm{Hu}, \binits{P.}},
\bauthor{\bsnm{Chen}, \binits{S.}},
\bauthor{\bsnm{Xu}, \binits{T.}},
\bauthor{\bsnm{Price}, \binits{P.M.}},
\bauthor{\bsnm{Ding}, \binits{Y.}},
\bauthor{\bsnm{Spencer}, \binits{B.A.}},
\bauthor{\bsnm{Nardo}, \binits{L.}},
\bauthor{\bsnm{Liu}, \binits{W.}},
\bauthor{\bsnm{Bao}, \binits{J.}},
\bauthor{\bsnm{Jones}, \binits{T.}},
\bauthor{\bsnm{Li}, \binits{H.}},
\bauthor{\bsnm{Cherry}, \binits{S.R.}}:
\batitle{First human imaging studies with the {EXPLORER} total-body {PET} scanner}.
\bjtitle{Journal of Nuclear Medicine}
\bvolume{60}(\bissue{3}),
\bfpage{299}--\blpage{303}
(\byear{2019})
\doiurl{10.2967/jnumed.119.226498}
\end{barticle}
\endbibitem

\bibitem[\protect\citeauthoryear{Gsaxner et~al.}{2023}]{bib3}
\begin{barticle}
\bauthor{\bsnm{Gsaxner}, \binits{C.}},
\bauthor{\bsnm{Li}, \binits{J.}},
\bauthor{\bsnm{Pepe}, \binits{A.}},
\bauthor{\bsnm{Jin}, \binits{Y.}},
\bauthor{\bsnm{Kleesiek}, \binits{J.}},
\bauthor{\bsnm{Schmalstieg}, \binits{D.}},
\bauthor{\bsnm{Egger}, \binits{J.}}:
\batitle{The {HoloLens} in medicine: A systematic review and taxonomy}.
\bjtitle{Medical Image Analysis}
\bvolume{85},
\bfpage{102757}
(\byear{2023})
\doiurl{10.1016/j.media.2023.102757}
\end{barticle}
\endbibitem

\bibitem[\protect\citeauthoryear{Kaashki et~al.}{2023}]{bib4}
\begin{barticle}
\bauthor{\bsnm{Kaashki}, \binits{N.N.}},
\bauthor{\bsnm{Hu}, \binits{P.}},
\bauthor{\bsnm{Munteanu}, \binits{A.}}:
\batitle{{Anet}: A deep neural network for automatic {3D} anthropometric measurement extraction}.
\bjtitle{IEEE Transactions on Multimedia}
\bvolume{25},
\bfpage{831}--\blpage{844}
(\byear{2023})
\doiurl{10.1109/TMM.2021.3132487}
\end{barticle}
\endbibitem

\bibitem[\protect\citeauthoryear{Xiao et~al.}{2021}]{bib5}
\begin{barticle}
\bauthor{\bsnm{Xiao}, \binits{J.}},
\bauthor{\bsnm{Yu}, \binits{H.}},
\bauthor{\bsnm{Sui}, \binits{X.}},
\bauthor{\bsnm{Hu}, \binits{Y.}},
\bauthor{\bsnm{Cao}, \binits{Y.}},
\bauthor{\bsnm{Liu}, \binits{G.}},
\bauthor{\bsnm{Zhang}, \binits{Y.}},
\bauthor{\bsnm{Hu}, \binits{P.}},
\bauthor{\bsnm{Wang}, \binits{Y.}},
\bauthor{\bsnm{Li}, \binits{C.}},
\bauthor{\bsnm{Xu}, \binits{B.}},
\bauthor{\bsnm{Shi}, \binits{H.}}:
\batitle{Can the {BMI}-based dose regimen be used to reduce injection activity and to obtain a constant image quality in oncological patients by {18F-FDG} total-body {PET/CT} imaging?}
\bjtitle{European Journal of Nuclear Medicine and Molecular Imaging}
\bvolume{49}(\bissue{1}),
\bfpage{269}--\blpage{278}
(\byear{2021})
\doiurl{10.1007/s00259-021-05462-5}
\end{barticle}
\endbibitem

\bibitem[\protect\citeauthoryear{Janiszewski et~al.}{2007}]{bib6}
\begin{barticle}
\bauthor{\bsnm{Janiszewski}, \binits{P.M.}},
\bauthor{\bsnm{Janssen}, \binits{I.}},
\bauthor{\bsnm{Ross}, \binits{R.}}:
\batitle{Does waist circumference predict diabetes and cardiovascular disease beyond commonly evaluated cardiometabolic risk factors?}
\bjtitle{Diabetes Care}
\bvolume{30}(\bissue{12}),
\bfpage{3105}--\blpage{3109}
(\byear{2007})
\doiurl{10.2337/dc07-0945}
\end{barticle}
\endbibitem

\bibitem[\protect\citeauthoryear{Zaidi and Tsui}{2009}]{bib44}
\begin{barticle}
\bauthor{\bsnm{Zaidi}, \binits{H.}},
\bauthor{\bsnm{Tsui}, \binits{B.M.W.}}:
\batitle{Review of computational anthropomorphic anatomical and physiological models}.
\bjtitle{Proceedings of the IEEE}
\bvolume{97}(\bissue{12}),
\bfpage{1938}--\blpage{1953}
(\byear{2009})
\doiurl{10.1109/JPROC.2009.2032852}
\end{barticle}
\endbibitem

\bibitem[\protect\citeauthoryear{Zamyatin and Nakanishi}{2007}]{bib45}
\begin{barticle}
\bauthor{\bsnm{Zamyatin}, \binits{A.A.}},
\bauthor{\bsnm{Nakanishi}, \binits{S.}}:
\batitle{Extension of the reconstruction field of view and truncation correction using sinogram decomposition}.
\bjtitle{Medical Physics}
\bvolume{34}(\bissue{5}),
\bfpage{1593}--\blpage{1604}
(\byear{2007})
\doiurl{10.1118/1.2721656}
\end{barticle}
\endbibitem

\bibitem[\protect\citeauthoryear{Arabi and Zaidi}{2020}]{bib47}
\begin{barticle}
\bauthor{\bsnm{Arabi}, \binits{H.}},
\bauthor{\bsnm{Zaidi}, \binits{H.}}:
\batitle{Truncation compensation and metallic dental implant artefact reduction in {PET/MRI} attenuation correction using deep learning-based object completion}.
\bjtitle{Physics in Medicine \& Biology}
\bvolume{65}(\bissue{19}),
\bfpage{195002}
(\byear{2020})
\doiurl{10.1088/1361-6560/abb02c}
\end{barticle}
\endbibitem

\bibitem[\protect\citeauthoryear{Yuan et~al.}{2018}]{bib8}
\begin{bchapter}
\bauthor{\bsnm{Yuan}, \binits{W.}},
\bauthor{\bsnm{Khot}, \binits{T.}},
\bauthor{\bsnm{Held}, \binits{D.}},
\bauthor{\bsnm{Mertz}, \binits{C.}},
\bauthor{\bsnm{Hebert}, \binits{M.}}:
\bctitle{{PCN}: Point completion network}.
In: \bbtitle{2018 International Conference on 3D Vision (3DV)},
pp. \bfpage{728}--\blpage{737}
(\byear{2018}).
\doiurl{10.1109/3DV.2018.00088}
\end{bchapter}
\endbibitem

\bibitem[\protect\citeauthoryear{Yan et~al.}{2022}]{bib9}
\begin{bchapter}
\bauthor{\bsnm{Yan}, \binits{X.}},
\bauthor{\bsnm{Lin}, \binits{L.}},
\bauthor{\bsnm{Mitra}, \binits{N.J.}},
\bauthor{\bsnm{Lischinski}, \binits{D.}},
\bauthor{\bsnm{Cohen-Or}, \binits{D.}},
\bauthor{\bsnm{Huang}, \binits{H.}}:
\bctitle{{ShapeFormer}: Transformer-based shape completion via sparse representation}.
In: \bbtitle{Proceedings of the IEEE/CVF Conference on Computer Vision and Pattern Recognition (CVPR)},
pp. \bfpage{6239}--\blpage{6249}
(\byear{2022}).
\doiurl{10.1109/CVPR52688.2022.00614}
\end{bchapter}
\endbibitem

\bibitem[\protect\citeauthoryear{Wu et~al.}{2016}]{bib10}
\begin{bchapter}
\bauthor{\bsnm{Wu}, \binits{J.}},
\bauthor{\bsnm{Zhang}, \binits{C.}},
\bauthor{\bsnm{Xue}, \binits{T.}},
\bauthor{\bsnm{Freeman}, \binits{B.}},
\bauthor{\bsnm{Tenenbaum}, \binits{J.}}:
\bctitle{Learning a probabilistic latent space of object shapes via {3D} generative-adversarial modeling}.
In: \bbtitle{Advances in Neural Information Processing Systems},
vol. \bseriesno{29}
(\byear{2016}).
\burl{https://proceedings.neurips.cc/paper/2016/hash/44f683a84163b3523afe57c2e008bc8c-Abstract.html}
\end{bchapter}
\endbibitem

\bibitem[\protect\citeauthoryear{Chibane et~al.}{2020}]{bib11}
\begin{bchapter}
\bauthor{\bsnm{Chibane}, \binits{J.}},
\bauthor{\bsnm{Alldieck}, \binits{T.}},
\bauthor{\bsnm{Pons-Moll}, \binits{G.}}:
\bctitle{Implicit functions in feature space for {3D} shape reconstruction and completion}.
In: \bbtitle{Proceedings of the IEEE/CVF Conference on Computer Vision and Pattern Recognition (CVPR)},
pp. \bfpage{6970}--\blpage{6981}
(\byear{2020}).
\doiurl{10.1109/CVPR42600.2020.00700}
\end{bchapter}
\endbibitem

\bibitem[\protect\citeauthoryear{Liu et~al.}{2021}]{bib12}
\begin{bchapter}
\bauthor{\bsnm{Liu}, \binits{G.}},
\bauthor{\bsnm{Rong}, \binits{Y.}},
\bauthor{\bsnm{Sheng}, \binits{L.}}:
\bctitle{{VoteHMR}: Occlusion-aware voting network for robust {3D} human mesh recovery from partial point clouds}.
In: \bbtitle{Proceedings of the 29th ACM International Conference on Multimedia},
pp. \bfpage{955}--\blpage{964}
(\byear{2021}).
\doiurl{10.1145/3474085.3475309}
\end{bchapter}
\endbibitem

\bibitem[\protect\citeauthoryear{Feng et~al.}{2023}]{bib13}
\begin{bchapter}
\bauthor{\bsnm{Feng}, \binits{H.}},
\bauthor{\bsnm{Kulits}, \binits{P.}},
\bauthor{\bsnm{Liu}, \binits{S.}},
\bauthor{\bsnm{Black}, \binits{M.J.}},
\bauthor{\bsnm{Abrevaya}, \binits{V.F.}}:
\bctitle{Generalizing neural human fitting to unseen poses with articulated {SE(3)} equivariance}.
In: \bbtitle{Proceedings of the IEEE/CVF International Conference on Computer Vision (ICCV)},
pp. \bfpage{7977}--\blpage{7988}
(\byear{2023}).
\doiurl{10.1109/ICCV51070.2023.00733}
\end{bchapter}
\endbibitem

\bibitem[\protect\citeauthoryear{Jang et~al.}{2023}]{bib14}
\begin{bchapter}
\bauthor{\bsnm{Jang}, \binits{H.}},
\bauthor{\bsnm{Kim}, \binits{M.}},
\bauthor{\bsnm{Bae}, \binits{J.}},
\bauthor{\bsnm{Kim}, \binits{Y.M.}}:
\bctitle{Dynamic mesh recovery from partial point cloud sequence}.
In: \bbtitle{Proceedings of the IEEE/CVF International Conference on Computer Vision (ICCV)},
pp. \bfpage{15074}--\blpage{15084}
(\byear{2023}).
\doiurl{10.1109/ICCV51070.2023.01384}
\end{bchapter}
\endbibitem

\bibitem[\protect\citeauthoryear{Bhatnagar et~al.}{2020}]{bib15}
\begin{bchapter}
\bauthor{\bsnm{Bhatnagar}, \binits{B.L.}},
\bauthor{\bsnm{Sminchisescu}, \binits{C.}},
\bauthor{\bsnm{Theobalt}, \binits{C.}},
\bauthor{\bsnm{Pons-Moll}, \binits{G.}}:
\bctitle{Combining implicit function learning and parametric models for {3D} human reconstruction}.
In: \beditor{\bsnm{Vedaldi}, \binits{A.}},
\beditor{\bsnm{Bischof}, \binits{H.}},
\beditor{\bsnm{Brox}, \binits{T.}},
\beditor{\bsnm{Frahm}, \binits{J.-M.}} (eds.)
\bbtitle{Computer Vision -- ECCV 2020}.
\bsertitle{Lecture Notes in Computer Science},
vol. \bseriesno{12347},
pp. \bfpage{311}--\blpage{329}
(\byear{2020}).
\doiurl{10.1007/978-3-030-58536-5_19}
\end{bchapter}
\endbibitem

\bibitem[\protect\citeauthoryear{Fonseca et~al.}{2021}]{bib46}
\begin{barticle}
\bauthor{\bsnm{Fonseca}, \binits{G.P.}},
\bauthor{\bsnm{Baer-Beck}, \binits{M.}},
\bauthor{\bsnm{Fournie}, \binits{E.}},
\bauthor{\bsnm{Hofmann}, \binits{C.}},
\bauthor{\bsnm{Rinaldi}, \binits{I.}},
\bauthor{\bsnm{Ollers}, \binits{M.C.}},
\bauthor{\bsnm{Elmpt}, \binits{W.J.C.}},
\bauthor{\bsnm{Verhaegen}, \binits{F.}}:
\batitle{Evaluation of novel {AI}-based extended field-of-view {CT} reconstructions}.
\bjtitle{Medical Physics}
\bvolume{48}(\bissue{7}),
\bfpage{3583}--\blpage{3594}
(\byear{2021})
\doiurl{10.1002/mp.14937}
\end{barticle}
\endbibitem

\bibitem[\protect\citeauthoryear{Loper et~al.}{2015}]{bib7}
\begin{barticle}
\bauthor{\bsnm{Loper}, \binits{M.}},
\bauthor{\bsnm{Mahmood}, \binits{N.}},
\bauthor{\bsnm{Romero}, \binits{J.}},
\bauthor{\bsnm{Pons-Moll}, \binits{G.}},
\bauthor{\bsnm{Black}, \binits{M.J.}}:
\batitle{{SMPL}: A skinned multi-person linear model}.
\bjtitle{ACM Transactions on Graphics}
\bvolume{34}(\bissue{6}),
\bfpage{1}--\blpage{16}
(\byear{2015})
\doiurl{10.1145/2816795.2818013} .
\bcomment{Article 248}
\end{barticle}
\endbibitem

\bibitem[\protect\citeauthoryear{Keller et~al.}{2024}]{bib17}
\begin{bchapter}
\bauthor{\bsnm{Keller}, \binits{M.}},
\bauthor{\bsnm{Arora}, \binits{V.}},
\bauthor{\bsnm{Dakri}, \binits{A.}},
\bauthor{\bsnm{Chandhok}, \binits{S.}},
\bauthor{\bsnm{Machann}, \binits{J.}},
\bauthor{\bsnm{Fritsche}, \binits{A.}},
\bauthor{\bsnm{Black}, \binits{M.J.}},
\bauthor{\bsnm{Pujades}, \binits{S.}}:
\bctitle{{HIT}: Estimating internal human implicit tissues from the body surface}.
In: \bbtitle{Proceedings of the IEEE/CVF Conference on Computer Vision and Pattern Recognition (CVPR)},
pp. \bfpage{3480}--\blpage{3490}
(\byear{2024}).
\doiurl{10.1109/CVPR52733.2024.00334}
\end{bchapter}
\endbibitem

\bibitem[\protect\citeauthoryear{Shetty et~al.}{2023}]{bib18}
\begin{barticle}
\bauthor{\bsnm{Shetty}, \binits{K.}},
\bauthor{\bsnm{Birkhold}, \binits{A.}},
\bauthor{\bsnm{Jaganathan}, \binits{S.}},
\bauthor{\bsnm{Strobel}, \binits{N.}},
\bauthor{\bsnm{Egger}, \binits{B.}},
\bauthor{\bsnm{Kowarschik}, \binits{M.}},
\bauthor{\bsnm{Maier}, \binits{A.}}:
\batitle{{BOSS}: Bones, organs and skin shape model}.
\bjtitle{Computers in Biology and Medicine}
\bvolume{165},
\bfpage{107383}
(\byear{2023})
\doiurl{10.1016/j.compbiomed.2023.107383}
\end{barticle}
\endbibitem

\bibitem[\protect\citeauthoryear{Gatidis et~al.}{2022}]{bib16}
\begin{barticle}
\bauthor{\bsnm{Gatidis}, \binits{S.}},
\bauthor{\bsnm{Hepp}, \binits{T.}},
\bauthor{\bsnm{Fr{\"u}h}, \binits{M.}},
\bauthor{\bsnm{La~Foug{\`e}re}, \binits{C.}},
\bauthor{\bsnm{Nikolaou}, \binits{K.}},
\bauthor{\bsnm{Pfannenberg}, \binits{C.}},
\bauthor{\bsnm{Sch{\"o}lkopf}, \binits{B.}},
\bauthor{\bsnm{K{\"u}stner}, \binits{T.}},
\bauthor{\bsnm{Cyran}, \binits{C.}},
\bauthor{\bsnm{Rubin}, \binits{D.}}:
\batitle{A whole-body {FDG-PET/CT} dataset with manually annotated tumor lesions}.
\bjtitle{Scientific Data}
\bvolume{9}(\bissue{1}),
\bfpage{601}
(\byear{2022})
\doiurl{10.1038/s41597-022-01718-3}
\end{barticle}
\endbibitem

\bibitem[\protect\citeauthoryear{Isensee et~al.}{2021}]{bib19}
\begin{barticle}
\bauthor{\bsnm{Isensee}, \binits{F.}},
\bauthor{\bsnm{Jaeger}, \binits{P.F.}},
\bauthor{\bsnm{Kohl}, \binits{S.A.A.}},
\bauthor{\bsnm{Petersen}, \binits{J.}},
\bauthor{\bsnm{Maier-Hein}, \binits{K.H.}}:
\batitle{{nnU-Net}: a self-configuring method for deep learning-based biomedical image segmentation}.
\bjtitle{Nature Methods}
\bvolume{18}(\bissue{2}),
\bfpage{203}--\blpage{211}
(\byear{2021})
\doiurl{10.1038/s41592-020-01008-z}
\end{barticle}
\endbibitem

\bibitem[\protect\citeauthoryear{Lorensen and Cline}{1987}]{bib20}
\begin{barticle}
\bauthor{\bsnm{Lorensen}, \binits{W.E.}},
\bauthor{\bsnm{Cline}, \binits{H.E.}}:
\batitle{Marching cubes: A high resolution {3D} surface construction algorithm}.
\bjtitle{ACM SIGGRAPH Computer Graphics}
\bvolume{21}(\bissue{4}),
\bfpage{163}--\blpage{169}
(\byear{1987})
\doiurl{10.1145/37402.37422}
\end{barticle}
\endbibitem

\bibitem[\protect\citeauthoryear{Taubin}{1995}]{bib21}
\begin{bchapter}
\bauthor{\bsnm{Taubin}, \binits{G.}}:
\bctitle{Curve and surface smoothing without shrinkage}.
In: \bbtitle{Proceedings of the Fifth International Conference on Computer Vision (ICCV)},
pp. \bfpage{852}--\blpage{857}
(\byear{1995}).
\doiurl{10.1109/ICCV.1995.466848}
\end{bchapter}
\endbibitem

\bibitem[\protect\citeauthoryear{Schroeder et~al.}{1992}]{bib22}
\begin{bchapter}
\bauthor{\bsnm{Schroeder}, \binits{W.J.}},
\bauthor{\bsnm{Zarge}, \binits{J.A.}},
\bauthor{\bsnm{Lorensen}, \binits{W.E.}}:
\bctitle{Decimation of triangle meshes}.
In: \bbtitle{Proceedings of the 19th Annual Conference on Computer Graphics and Interactive Techniques},
pp. \bfpage{65}--\blpage{70}
(\byear{1992}).
\doiurl{10.1145/133994.134010}
\end{bchapter}
\endbibitem

\bibitem[\protect\citeauthoryear{Kim et~al.}{2021}]{bib23}
\begin{bchapter}
\bauthor{\bsnm{Kim}, \binits{H.}},
\bauthor{\bsnm{Kim}, \binits{J.}},
\bauthor{\bsnm{Kam}, \binits{J.}},
\bauthor{\bsnm{Park}, \binits{J.}},
\bauthor{\bsnm{Lee}, \binits{S.}}:
\bctitle{Deep virtual markers for articulated {3D} shapes}.
In: \bbtitle{Proceedings of the IEEE/CVF International Conference on Computer Vision (ICCV)},
pp. \bfpage{11615}--\blpage{11625}
(\byear{2021}).
\doiurl{10.1109/ICCV48922.2021.01141}
\end{bchapter}
\endbibitem

\bibitem[\protect\citeauthoryear{Clever et~al.}{2023}]{bib24}
\begin{barticle}
\bauthor{\bsnm{Clever}, \binits{H.M.}},
\bauthor{\bsnm{Grady}, \binits{P.L.}},
\bauthor{\bsnm{Turk}, \binits{G.}},
\bauthor{\bsnm{Kemp}, \binits{C.C.}}:
\batitle{{BodyPressure} - inferring body pose and contact pressure from a depth image}.
\bjtitle{IEEE Transactions on Pattern Analysis and Machine Intelligence}
\bvolume{45}(\bissue{1}),
\bfpage{137}--\blpage{153}
(\byear{2023})
\doiurl{10.1109/TPAMI.2022.3158902}
\end{barticle}
\endbibitem

\bibitem[\protect\citeauthoryear{Pavlakos et~al.}{2019}]{bib26}
\begin{bchapter}
\bauthor{\bsnm{Pavlakos}, \binits{G.}},
\bauthor{\bsnm{Choutas}, \binits{V.}},
\bauthor{\bsnm{Ghorbani}, \binits{N.}},
\bauthor{\bsnm{Bolkart}, \binits{T.}},
\bauthor{\bsnm{Osman}, \binits{A.A.A.}},
\bauthor{\bsnm{Tzionas}, \binits{D.}},
\bauthor{\bsnm{Black}, \binits{M.J.}}:
\bctitle{Expressive body capture: {3D} hands, face, and body from a single image}.
In: \bbtitle{Proceedings of the IEEE/CVF Conference on Computer Vision and Pattern Recognition (CVPR)},
pp. \bfpage{10975}--\blpage{10985}
(\byear{2019}).
\doiurl{10.1109/CVPR.2019.01123}
\end{bchapter}
\endbibitem

\bibitem[\protect\citeauthoryear{Bogo et~al.}{2016}]{bib27}
\begin{bchapter}
\bauthor{\bsnm{Bogo}, \binits{F.}},
\bauthor{\bsnm{Kanazawa}, \binits{A.}},
\bauthor{\bsnm{Lassner}, \binits{C.}},
\bauthor{\bsnm{Gehler}, \binits{P.}},
\bauthor{\bsnm{Romero}, \binits{J.}},
\bauthor{\bsnm{Black}, \binits{M.J.}}:
\bctitle{Keep it {SMPL}: Automatic estimation of {3D} human pose and shape from a single image}.
In: \bbtitle{Computer Vision -- ECCV 2016}.
\bsertitle{Lecture Notes in Computer Science},
vol. \bseriesno{9909},
pp. \bfpage{561}--\blpage{578}
(\byear{2016}).
\doiurl{10.1007/978-3-319-46454-1_34}
\end{bchapter}
\endbibitem

\bibitem[\protect\citeauthoryear{Kingma and Ba}{2015}]{bib25}
\begin{bchapter}
\bauthor{\bsnm{Kingma}, \binits{D.P.}},
\bauthor{\bsnm{Ba}, \binits{J.}}:
\bctitle{{Adam}: A method for stochastic optimization}.
In: \bbtitle{Proceedings of the 3rd International Conference on Learning Representations (ICLR)}
(\byear{2015}).
\burl{https://arxiv.org/abs/1412.6980}
\end{bchapter}
\endbibitem

\bibitem[\protect\citeauthoryear{Desbrun et~al.}{1999}]{bib29}
\begin{bchapter}
\bauthor{\bsnm{Desbrun}, \binits{M.}},
\bauthor{\bsnm{Meyer}, \binits{M.}},
\bauthor{\bsnm{Schr{\"o}der}, \binits{P.}},
\bauthor{\bsnm{Barr}, \binits{A.H.}}:
\bctitle{Implicit fairing of irregular meshes using diffusion and curvature flow}.
In: \bbtitle{Proceedings of the 26th Annual Conference on Computer Graphics and Interactive Techniques (SIGGRAPH)},
pp. \bfpage{317}--\blpage{324}
(\byear{1999}).
\doiurl{10.1145/311535.311576}
\end{bchapter}
\endbibitem

\bibitem[\protect\citeauthoryear{Kazhdan et~al.}{2006}]{bib30}
\begin{bchapter}
\bauthor{\bsnm{Kazhdan}, \binits{M.}},
\bauthor{\bsnm{Bolitho}, \binits{M.}},
\bauthor{\bsnm{Hoppe}, \binits{H.}}:
\bctitle{{Poisson} surface reconstruction}.
In: \bbtitle{Symposium on Geometry Processing},
pp. \bfpage{61}--\blpage{70}
(\byear{2006}).
\doiurl{10.2312/SGP/SGP06/061-070}
\end{bchapter}
\endbibitem

\bibitem[\protect\citeauthoryear{Schroeder et~al.}{2000}]{bib31}
\begin{barticle}
\bauthor{\bsnm{Schroeder}, \binits{W.J.}},
\bauthor{\bsnm{Avila}, \binits{L.S.}},
\bauthor{\bsnm{Hoffman}, \binits{W.}}:
\batitle{Visualizing with {VTK}: A tutorial}.
\bjtitle{IEEE Computer Graphics and Applications}
\bvolume{20}(\bissue{5}),
\bfpage{20}--\blpage{27}
(\byear{2000})
\doiurl{10.1109/38.865875}
\end{barticle}
\endbibitem

\bibitem[\protect\citeauthoryear{Paszke et~al.}{2019}]{bib32}
\begin{bchapter}
\bauthor{\bsnm{Paszke}, \binits{A.}},
\bauthor{\bsnm{Gross}, \binits{S.}},
\bauthor{\bsnm{Massa}, \binits{F.}},
\bauthor{\bsnm{Lerer}, \binits{A.}},
\bauthor{\bsnm{Bradbury}, \binits{J.}},
\bauthor{\bsnm{Chanan}, \binits{G.}},
\bauthor{\bsnm{Killeen}, \binits{T.}},
\bauthor{\bsnm{Lin}, \binits{Z.}},
\bauthor{\bsnm{Gimelshein}, \binits{N.}},
\bauthor{\bsnm{Antiga}, \binits{L.}},
\bauthor{\bsnm{Desmaison}, \binits{A.}},
\bauthor{\bsnm{Kopf}, \binits{A.}},
\bauthor{\bsnm{Yang}, \binits{E.}},
\bauthor{\bsnm{DeVito}, \binits{Z.}},
\bauthor{\bsnm{Raison}, \binits{M.}},
\bauthor{\bsnm{Tejani}, \binits{A.}},
\bauthor{\bsnm{Chilamkurthy}, \binits{S.}},
\bauthor{\bsnm{Steiner}, \binits{B.}},
\bauthor{\bsnm{Fang}, \binits{L.}},
\bauthor{\bsnm{Bai}, \binits{J.}},
\bauthor{\bsnm{Chintala}, \binits{S.}}:
\bctitle{{PyTorch}: An imperative style, high-performance deep learning library}.
In: \bbtitle{Advances in Neural Information Processing Systems},
vol. \bseriesno{32},
pp. \bfpage{8026}--\blpage{8037}
(\byear{2019}).
\burl{https://proceedings.neurips.cc/paper/2019/hash/bdbca288fee7f92f2bfa9f7012727740-Abstract.html}
\end{bchapter}
\endbibitem

\bibitem[\protect\citeauthoryear{Tsang et~al.}{2024}]{bib33}
\begin{botherref}
\oauthor{\bsnm{Tsang}, \binits{C.F.}},
\oauthor{\bsnm{Shugrina}, \binits{M.}},
\oauthor{\bsnm{Lafleche}, \binits{J.-F.}},
\oauthor{\bsnm{Perel}, \binits{O.}},
\oauthor{\bsnm{Loop}, \binits{C.}},
\oauthor{\bsnm{Takikawa}, \binits{T.}},
\oauthor{\bsnm{Modi}, \binits{V.}},
\oauthor{\bsnm{Zook}, \binits{A.}},
\oauthor{\bsnm{Wang}, \binits{J.}},
\oauthor{\bsnm{Chen}, \binits{W.}},
\oauthor{\bsnm{Shen}, \binits{T.}},
\oauthor{\bsnm{Gao}, \binits{J.}},
\oauthor{\bsnm{Jatavallabhula}, \binits{K.M.}},
\oauthor{\bsnm{Smith}, \binits{E.}},
\oauthor{\bsnm{Rozantsev}, \binits{A.}},
\oauthor{\bsnm{Fidler}, \binits{S.}},
\oauthor{\bsnm{State}, \binits{G.}},
\oauthor{\bsnm{Gorski}, \binits{J.}},
\oauthor{\bsnm{Xiang}, \binits{T.}},
\oauthor{\bsnm{Li}, \binits{J.}},
\oauthor{\bsnm{Li}, \binits{M.}},
\oauthor{\bsnm{Lebaredian}, \binits{R.}}:
{Kaolin}: A {PyTorch} Library for Accelerating {3D} Deep Learning Research.
Software library.
Accessed 16 October 2026
(2024).
\url{https://github.com/NVIDIAGameWorks/kaolin}
\end{botherref}
\endbibitem

\bibitem[\protect\citeauthoryear{M{\"u}ller et~al.}{2021}]{bib28}
\begin{bchapter}
\bauthor{\bsnm{M{\"u}ller}, \binits{L.}},
\bauthor{\bsnm{Osman}, \binits{A.A.A.}},
\bauthor{\bsnm{Tang}, \binits{S.}},
\bauthor{\bsnm{Huang}, \binits{C.-H.P.}},
\bauthor{\bsnm{Black}, \binits{M.J.}}:
\bctitle{On self-contact and human pose}.
In: \bbtitle{Proceedings of the IEEE/CVF Conference on Computer Vision and Pattern Recognition (CVPR)},
pp. \bfpage{9990}--\blpage{9999}
(\byear{2021}).
\doiurl{10.1109/CVPR46437.2021.00986}
\end{bchapter}
\endbibitem

\bibitem[\protect\citeauthoryear{Lowekamp et~al.}{2013}]{bib34}
\begin{barticle}
\bauthor{\bsnm{Lowekamp}, \binits{B.C.}},
\bauthor{\bsnm{Chen}, \binits{D.T.}},
\bauthor{\bsnm{Ib{\'a}{\~n}ez}, \binits{L.}},
\bauthor{\bsnm{Blezek}, \binits{D.}}:
\batitle{The design of {SimpleITK}}.
\bjtitle{Frontiers in Neuroinformatics}
\bvolume{7},
\bfpage{45}
(\byear{2013})
\doiurl{10.3389/fninf.2013.00045}
\end{barticle}
\endbibitem

\bibitem[\protect\citeauthoryear{Bradski and Kaehler}{2008}]{bib35}
\begin{bbook}
\bauthor{\bsnm{Bradski}, \binits{G.}},
\bauthor{\bsnm{Kaehler}, \binits{A.}}:
\bbtitle{Learning {OpenCV}: Computer Vision with the {OpenCV} Library}.
\bpublisher{O'Reilly Media},
\blocation{Sebastopol, CA}
(\byear{2008}).
\burl{https://www.oreilly.com/library/view/learning-opencv/9780596516130/}
\end{bbook}
\endbibitem

\bibitem[\protect\citeauthoryear{Muntoni and Cignoni}{2021}]{bib36}
\begin{botherref}
\oauthor{\bsnm{Muntoni}, \binits{A.}},
\oauthor{\bsnm{Cignoni}, \binits{P.}}:
{PyMeshLab}.
Zenodo.
Software
(2021).
\doiurl{10.5281/zenodo.4438750}
\end{botherref}
\endbibitem

\bibitem[\protect\citeauthoryear{Attene}{2010}]{bib37}
\begin{barticle}
\bauthor{\bsnm{Attene}, \binits{M.}}:
\batitle{A lightweight approach to repairing digitized polygon meshes}.
\bjtitle{The Visual Computer}
\bvolume{26},
\bfpage{1393}--\blpage{1406}
(\byear{2010})
\doiurl{10.1007/s00371-010-0416-3}
\end{barticle}
\endbibitem

\bibitem[\protect\citeauthoryear{Durnin and Womersley}{1974}]{bib38}
\begin{barticle}
\bauthor{\bsnm{Durnin}, \binits{J.V.G.A.}},
\bauthor{\bsnm{Womersley}, \binits{J.}}:
\batitle{Body fat assessed from total body density and its estimation from skinfold thickness: measurements on 481 men and women aged from 16 to 72 years}.
\bjtitle{British Journal of Nutrition}
\bvolume{32}(\bissue{1}),
\bfpage{77}--\blpage{97}
(\byear{1974})
\doiurl{10.1079/BJN19740060}
\end{barticle}
\endbibitem

\bibitem[\protect\citeauthoryear{Choutas et~al.}{2022}]{bib39}
\begin{bchapter}
\bauthor{\bsnm{Choutas}, \binits{V.}},
\bauthor{\bsnm{M{\"u}ller}, \binits{L.}},
\bauthor{\bsnm{Huang}, \binits{C.-H.P.}},
\bauthor{\bsnm{Tang}, \binits{S.}},
\bauthor{\bsnm{Tzionas}, \binits{D.}},
\bauthor{\bsnm{Black}, \binits{M.J.}}:
\bctitle{Accurate {3D} body shape regression using metric and semantic attributes}.
In: \bbtitle{Proceedings of the IEEE/CVF Conference on Computer Vision and Pattern Recognition (CVPR)},
pp. \bfpage{2718}--\blpage{2728}
(\byear{2022}).
\doiurl{10.1109/CVPR52688.2022.00274}
\end{bchapter}
\endbibitem

\bibitem[\protect\citeauthoryear{Welch}{1947}]{bib48}
\begin{barticle}
\bauthor{\bsnm{Welch}, \binits{B.L.}}:
\batitle{The generalization of `{Student}'s' problem when several different population variances are involved}.
\bjtitle{Biometrika}
\bvolume{34}(\bissue{1-2}),
\bfpage{28}--\blpage{35}
(\byear{1947})
\doiurl{10.1093/biomet/34.1-2.28}
\end{barticle}
\endbibitem

\bibitem[\protect\citeauthoryear{Redlarski et~al.}{2016}]{bib40}
\begin{barticle}
\bauthor{\bsnm{Redlarski}, \binits{G.}},
\bauthor{\bsnm{Palkowski}, \binits{A.}},
\bauthor{\bsnm{Krawczuk}, \binits{M.}}:
\batitle{Body surface area formulae: an alarming ambiguity}.
\bjtitle{Scientific Reports}
\bvolume{6},
\bfpage{27966}
(\byear{2016})
\doiurl{10.1038/srep27966}
\end{barticle}
\endbibitem

\bibitem[\protect\citeauthoryear{Verweij et~al.}{2013}]{bib41}
\begin{barticle}
\bauthor{\bsnm{Verweij}, \binits{L.M.}},
\bauthor{\bsnm{Terwee}, \binits{C.B.}},
\bauthor{\bsnm{Proper}, \binits{K.I.}},
\bauthor{\bsnm{Hulshof}, \binits{C.T.J.}},
\bauthor{\bsnm{Mechelen}, \binits{W.}}:
\batitle{Measurement error of waist circumference: gaps in knowledge}.
\bjtitle{Public Health Nutrition}
\bvolume{16}(\bissue{2}),
\bfpage{281}--\blpage{288}
(\byear{2013})
\doiurl{10.1017/S1368980012002741}
\end{barticle}
\endbibitem

\bibitem[\protect\citeauthoryear{Ostchega et~al.}{2019}]{bib42}
\begin{botherref}
\oauthor{\bsnm{Ostchega}, \binits{Y.}},
\oauthor{\bsnm{Seu}, \binits{R.}},
\oauthor{\bsnm{Sarafrazi~Isfahani}, \binits{N.}},
\oauthor{\bsnm{Zhang}, \binits{G.}},
\oauthor{\bsnm{Hughes}, \binits{J.P.}},
\oauthor{\bsnm{Miller}, \binits{I.}}:
Waist circumference measurement methodology study: {National Health and Nutrition Examination Survey}, 2016.
Vital and Health Statistics, Series 2 182,
National Center for Health Statistics,
Hyattsville, MD
(jan 2019).
DHHS Publication No. 2019-1382.
\url{https://stacks.cdc.gov/view/cdc/61728}
\end{botherref}
\endbibitem

\bibitem[\protect\citeauthoryear{{NVIDIA Corporation}}{2025}]{nvidia2025blackwell}
\begin{botherref}
\oauthor{\bsnm{{NVIDIA Corporation}}}:
{NVIDIA RTX Blackwell GPU Architecture}.
White paper,
NVIDIA Corporation
(2025).
Version 1.1.
\url{https://images.nvidia.com/aem-dam/Solutions/geforce/blackwell/nvidia-rtx-blackwell-gpu-architecture.pdf}
\end{botherref}
\endbibitem

\bibitem[\protect\citeauthoryear{{NVIDIA Corporation}}{}]{nvidiatitanv}
\begin{botherref}
\oauthor{\bsnm{{NVIDIA Corporation}}}:
{NVIDIA TITAN V}.
Product specifications.
Accessed 19 October 2026.
\url{https://www.nvidia.com/cs-cz/titan/titan-v/}
\end{botherref}
\endbibitem

\bibitem[\protect\citeauthoryear{Zheng et~al.}{2022}]{bib43}
\begin{barticle}
\bauthor{\bsnm{Zheng}, \binits{Z.}},
\bauthor{\bsnm{Yu}, \binits{T.}},
\bauthor{\bsnm{Liu}, \binits{Y.}},
\bauthor{\bsnm{Dai}, \binits{Q.}}:
\batitle{{PaMIR}: Parametric model-conditioned implicit representation for image-based human reconstruction}.
\bjtitle{IEEE Transactions on Pattern Analysis and Machine Intelligence}
\bvolume{44}(\bissue{6}),
\bfpage{3170}--\blpage{3184}
(\byear{2022})
\doiurl{10.1109/TPAMI.2021.3050505}
\end{barticle}
\endbibitem

\end{thebibliography}

\end{document}